\pdfoutput=1

\documentclass{article}
\PassOptionsToPackage{numbers, sort&compress}{natbib}

\usepackage[main, final]{neurips_2026}

\usepackage[utf8]{inputenc} % allow utf-8 input
\usepackage[T1]{fontenc}    % use 8-bit T1 fonts
\usepackage{hyperref}       % hyperlinks
\usepackage{url}            % simple URL typesetting
\usepackage{booktabs}       % professional-quality tables
\usepackage{amsfonts}       % blackboard math symbols
\usepackage{nicefrac}       % compact symbols for 1/2, etc.
\usepackage{microtype}      % microtypography
\usepackage{xcolor}         % colors
\usepackage{enumitem}
\usepackage{pifont}

\usepackage{graphicx}
\usepackage{wrapfig}
\usepackage{subcaption}
\usepackage{multirow}
\usepackage{tcolorbox}
\tcbuselibrary{breakable, skins}
\newtcolorbox{longprompt}{
    breakable,
    enhanced standard,
    colback=white,
    colframe=black,
    boxrule=0.5pt,
    arc=0pt,
    sharp corners,
    left=4pt, right=4pt, top=4pt, bottom=4pt,
    fontupper=\small,
}

\usepackage[table]{xcolor}
\definecolor{pcmlightblue}{RGB}{230,244,255}

\usepackage{amsmath}
\usepackage{amssymb}
\usepackage{amsfonts}
\usepackage{mathtools}
\usepackage{amsthm}
\usepackage[capitalize,noabbrev]{cleveref}
\usepackage{algorithm}
\usepackage{algpseudocode}

\theoremstyle{plain}

\theoremstyle{definition}

\theoremstyle{remark}

\title{PreCoMem: Predictive Cognitive Memory for Self-Evolving Long-Term Dialogue Agents}

\author{%
  \textbf{Jian Zhong}$^{1}$\thanks{Equal contribution.} \hspace{0.45em}
  \textbf{Zeyu Liu}$^{1*}$ \hspace{0.45em}
  \textbf{Pingchuan Cao}$^{1}$ \hspace{0.45em}
  \textbf{Rongduo Han}$^{1}$ \hspace{0.45em}
  \textbf{Shunye Tang}$^{1}$ \\
  \textbf{Guohuan Xie}$^{3}$ \hspace{0.45em}
  \textbf{Mingyuan Qin}$^{4}$ \hspace{0.45em}
  \textbf{Ran Ji}$^{5}$ \hspace{0.45em}
  \textbf{Xiao Liang}$^{2}$ \hspace{0.45em}
  \textbf{Wei Wang}$^{1}$\thanks{Corresponding authors.} \hspace{0.45em}
  \textbf{Haining Zhang}$^{1\dagger}$ \\[4pt]
  $^{1}$Nankai University \quad
  $^{2}$vivo AI Lab \quad
  $^{3}$Tsinghua University \\
  $^{4}$Huazhong University of Science and Technology \\
  $^{5}$University of the Chinese Academy of Sciences \\
  \texttt{zhongjian@mail.nankai.edu.cn}
}

\begin{document}
% Hide main-text sections from the appendix table of contents (re-enabled in 6_appendix.tex).
\addtocontents{toc}{\protect\setcounter{tocdepth}{-10}}

\maketitle

\begin{abstract}

Long-horizon dialogue agents need to keep user memories stable enough for personalization while adapting when the user genuinely changes. Existing memory systems often append every observation or overwrite old entries by heuristic rules, which makes transient noise hard to distinguish from real belief drift. We introduce \textbf{PreCoMem}, a memory consolidation framework inspired by predictive coding that represents memory as confidence-weighted beliefs about the user's state. Its core signal, \emph{Effective Surprisal}, combines semantic deviation and contradiction evidence, then downweights this signal when retrieval is diffuse and unreliable. According to effective surprisal, a two-threshold gate maps each turn to one of three deterministic updates: \texttt{MAINTAIN} reinforces supported beliefs, \texttt{PROFILE} stores ambiguous cues as low-confidence hypotheses, and \texttt{CORRECT} softly decays reliably contradicted facts. The \texttt{PROFILE} stage acts as an explicit waiting room, preventing premature commitment to noisy evidence. Experiments on LoCoMo, LongMemEval\mbox{-}S, and PersonaMem\mbox{-}v2 show consistent state-of-the-art accuracy; our new DiMoBench confirms that PreCoMem adapts to non-stationary belief drift without over-reacting to noise. The code is available at \url{https://github.com/ZhongWishing/PreCoMem}.

\end{abstract}

\vspace{-4mm}
\section{Introduction}
\label{sec:intro}

\begin{figure}[h]
  \centering
  \vspace{-12pt}
  \includegraphics[width=0.63\textwidth]{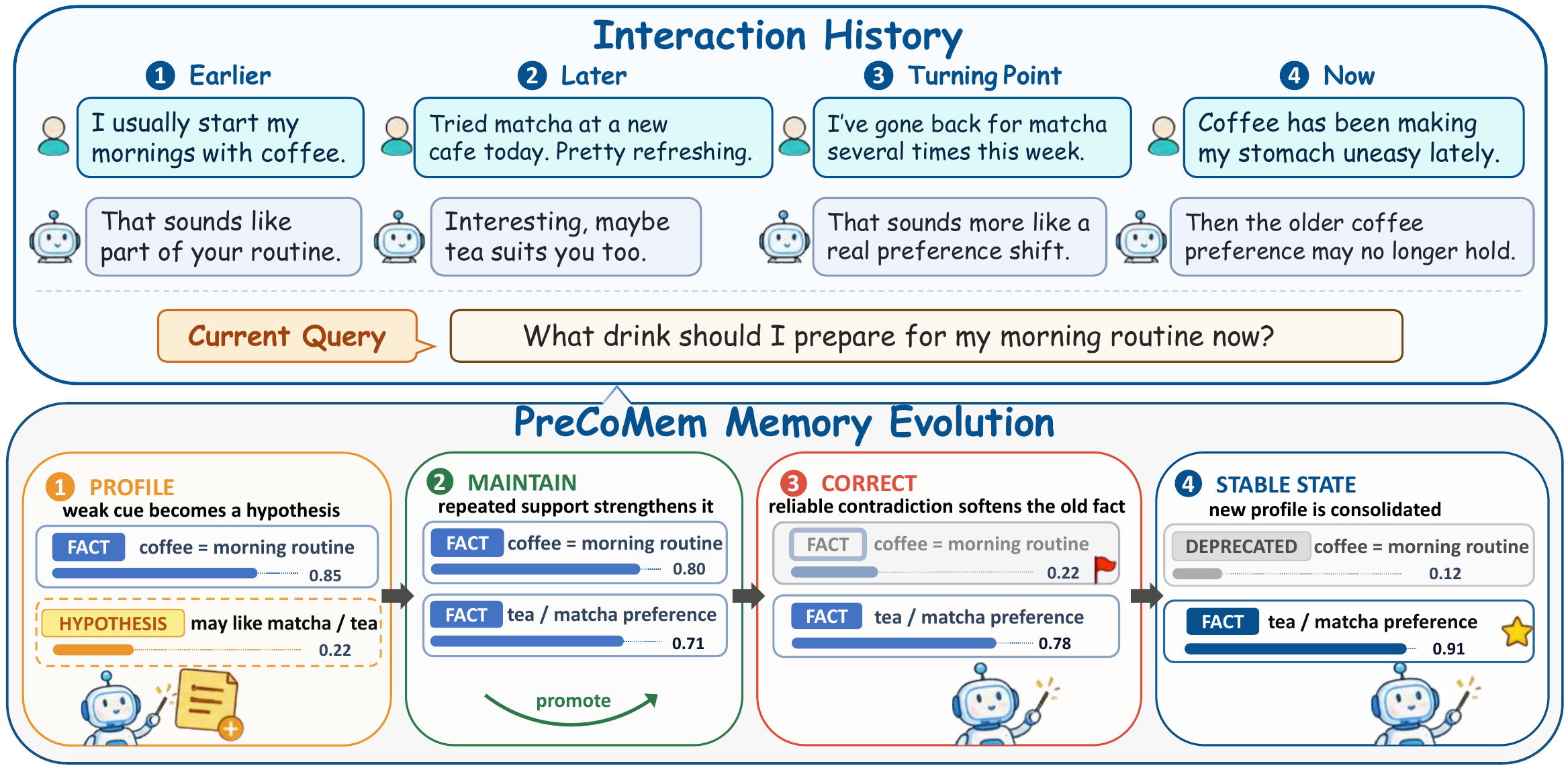}
  \caption{\textbf{Memory evolution case study (coffee~$\rightarrow$~tea).} PreCoMem autonomously manages the lifecycle of facts and hypotheses, achieving a stable transition between conflicting belief states without manual intervention.}
  \label{fig:pcm-case-study}
\end{figure}

% Large Language Models (LLMs) are moving from stateless responders toward long-term partners that interact with the user over hundreds or thousands of turns~\citep{park2023generative,wang2024survey,xi2025rise}. Such partners need more than stored transcripts; they need a working model of the user~\citep{zhong2024memorybank, yu2025memagent}. 
% We use \emph{belief} to mean a compact, actionable claim about the user's preferences, goals, or life context, paired with a confidence weight that can be reinforced or revised over time. 
% Because this model drives personalization and self-correction, long-term memory is more than context extension~\citep{packer2023memgpt,wang2023augmenting}: it is how the agent stays consistent across sessions.
Large Language Models (LLMs) are moving from stateless responders toward long-term partners that interact with the user over hundreds or thousands of turns~\citep{park2023generative,wang2024survey,xi2025rise}. Such partners need more than stored transcripts; they require a user model to sustain personalized interaction~\citep{zhong2024memorybank, jiang2025know}.
To formalize the core units of this user model, we define a \emph{belief} as a compact, actionable claim about the user's preferences, goals, or life context, paired with a confidence weight that can be reinforced or revised over time.
This user model underpins both agent personalization and behavioral self-correction, thereby redefining the role of long-term memory: rather than merely serving as simple context extension~\citep{packer2023memgpt,wang2023augmenting}, long-term memory enables the agent to maintain consistent behavior across sessions.

The harder question is not \emph{whether} information can be stored, but \emph{how} it should be consolidated and revised. A practical memory consolidation mechanism must decide which observations to retain, how strongly to trust them, and when to maintain, revise, or correct stored items without explicit supervision. We refer to this as the \textbf{memory stability--plasticity problem}: the agent should preserve stable user traits while remaining plastic enough to adapt when those traits change~\citep{dohare2024loss}. Long-horizon interaction makes the problem hard for three coupled reasons. First, most turns carry little signal, while a small fraction may contain cues indicative of more persistent user states. Second, user preferences and life facts evolve over time~\citep{dhingra2022time}, so the system must adapt under \emph{non-stationarity and belief drift}~\citep{yousuf2025can,dongre2025drift}. Third, corrective feedback is rarely explicit~\citep{jannach2018recommending}, and noisy retrieval over long contexts further blurs the picture~\citep{yoran2023making,liu2024lost,cuconasu2024power,wu2024longmemeval}. Therefore, the brittle memory update mechanism causes two opposite failures: premature correction from noise, or persistent over-reliance on outdated memories, which we call \emph{old-memory pull}, a phenomenon related to loss of plasticity and retrieval interference in continual adaptation systems~\citep{dohare2024loss,hu2026continual}.

Recent memory systems have made impressive progress on retrieval and storage organization. Hierarchical and OS-style stores~\citep{packer2023memgpt,li2025memos,hu2026evermemos}, structured indices with timestamps or topic shards~\citep{rasmussen2025zep,wang2025mem,li2026timem,tao2026membox}, and agentic retrievers~\citep{xu2025mem,yan2025general} all sharpen \emph{what} the system can recall. However, the question of \emph{when} to update is still often handled by recency overwriting, fixed-threshold importance, or one-shot append rules~\citep{zhong2024memorybank, chhikara2025mem0}. These triggers cannot reliably separate persistent drift from transient noise~\citep{hu2026continual}, so they collapse the stability--plasticity trade-off precisely where long-horizon dialogue stresses it most. A fuller literature discussion is provided in Appendix~\cref{sec:related}.

We draw inspiration from \emph{predictive coding}, which posits that learning is driven by mismatch between expectation and reality, i.e., prediction error~\citep{friston2010free,clark2013whatever}. In information theory the same quantity is called \emph{surprisal}~\citep{levy2008expectation,meister2020if}. We use this view as a design principle for the update path rather than as a claim that dialogue memory is a fully specified probabilistic model. The resulting criterion is simple: events consistent with current memory should be reinforced, while surprising events may require learning or correction~\citep{ranganath2003neural,sinclair2019prediction}. Crucially, surprisal must be paired with a measure of \emph{retrieval reliability}, so that update decisions stay conservative when recall itself is uncertain or noisy~\citep{kuhn2023semantic,yoran2023making}.

Building on this view, we propose \textbf{PreCoMem (Predictive Cognitive Memory)}, a self-evolving long-term memory framework for conversational LLM agents. PreCoMem operates in two active phases over a shared memory store. In \emph{memory leveraging}, it retrieves confidence-weighted beliefs to condition response generation. In \emph{memory evolution}, it compares new observations with recalled evidence and computes \emph{Effective Surprisal} $S_{\mathrm{eff}}$, a reliability-adjusted mismatch score. A two-threshold gate then routes each update to \texttt{MAINTAIN}, \texttt{PROFILE}, or \texttt{CORRECT}, corresponding to consistent, ambiguous, and reliably contradictory evidence. Figure~\ref{fig:pcm-case-study} illustrates this loop on a coffee-to-tea preference shift, where a low-confidence hypothesis is gradually promoted to a fact while the contradicted old fact is softly decayed.

In summary, we (i) propose \textbf{PreCoMem}, a memory consolidation framework that casts long-term memory as a dynamic prediction of the user's state; (ii) introduce a \emph{reliability-weighted surprisal signal} that gates each update through \texttt{MAINTAIN}, \texttt{PROFILE}, or \texttt{CORRECT}; and (iii) evaluate on LoCoMo, LongMemEval\mbox{-}S, PersonaMem\mbox{-}v2, and DiMoBench/DiMoBench\mbox{-}XL, controlled write-path stress tests with gradual drift and decoy retrieval. PreCoMem outperforms the strongest contemporary baselines while keeping effective memory stable across 1050 turns.

\vspace{-3mm}
\section{PreCoMem}
\label{sec:method}

\vspace{-3mm}
\subsection{Framework overview}
\label{sec:pcm_framework}

\begin{figure*}[t]
  \centering
  \includegraphics[width=0.96\textwidth]{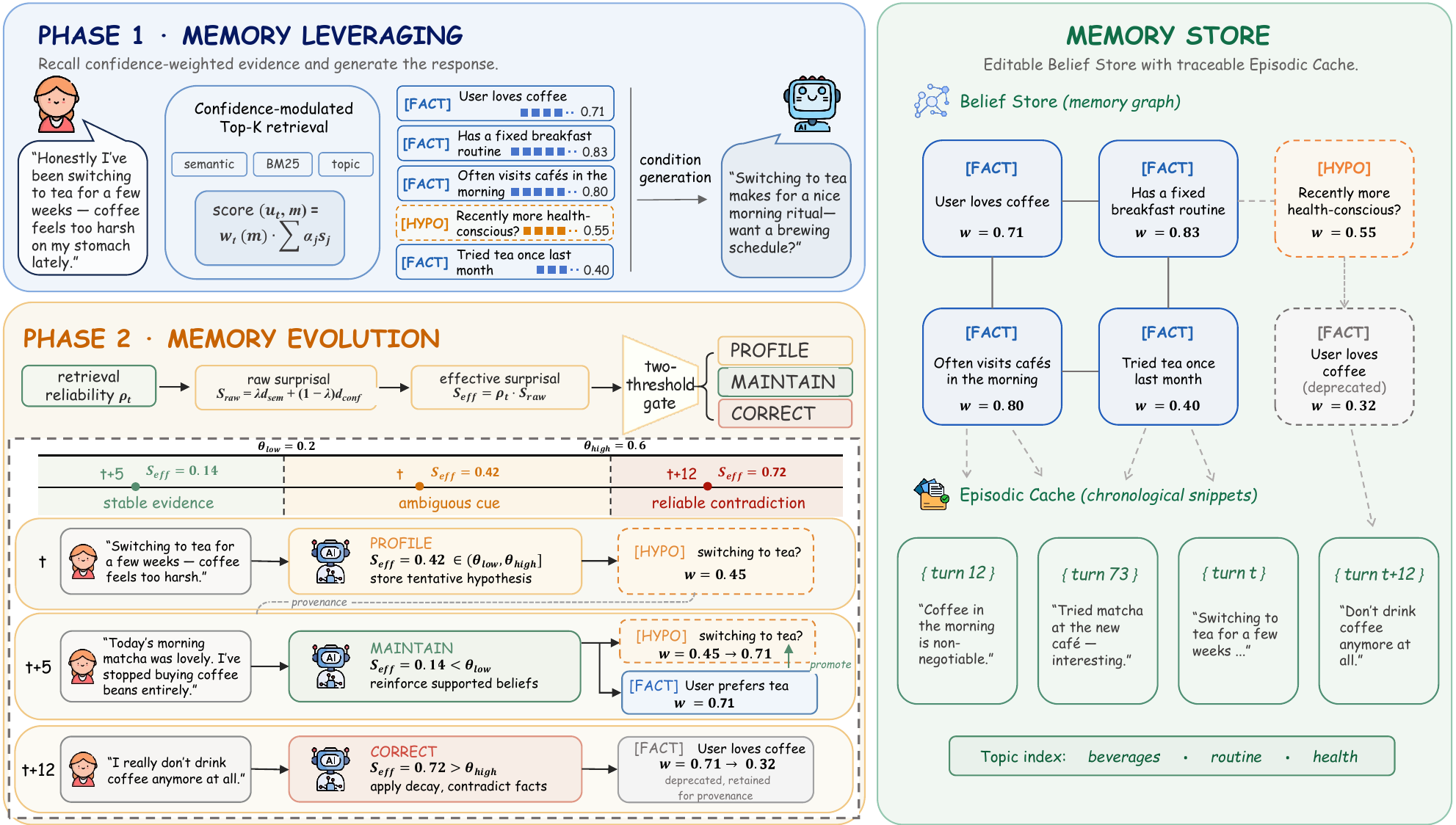}
  \caption{\textbf{The PreCoMem framework.} Two active phases operate over a shared memory store. \textbf{Phase 1, Memory Leveraging}: a confidence-modulated retrieval score combines semantic, lexical, and topic signals to select Top-$K$ beliefs $\mathcal{C}_t$ for response generation. \textbf{Phase 2, Memory Evolution}: the effective surprisal $S_{\mathrm{eff}}$ and a two-threshold gate route each turn to \texttt{MAINTAIN}, \texttt{PROFILE}, or \texttt{CORRECT}, illustrated by representative low/medium/high-$S_{\mathrm{eff}}$ turns. \textbf{Memory Store}: an editable Belief Store $\mathcal{B}$ of \texttt{FACT}/\texttt{HYPO} nodes with confidence weights, linked by provenance to an immutable Episodic Cache $\mathcal{E}$ under a shared topic index; belief items are updated independently, and the drawn links are provenance pointers and topic groupings (Appendix~\cref{app:memory_store_example}).}
  \label{fig:framework}
\end{figure*}

As illustrated in \Cref{fig:framework}, PreCoMem operates as a closed loop of two active phases around a shared memory store. \textbf{Memory Leveraging} retrieves confidence-weighted beliefs and uses them to condition response generation. \textbf{Memory Evolution} then revises the memory state in light of the new interaction: it computes \emph{Effective Surprisal} $S_{\mathrm{eff}}$ against recalled evidence, and routes the update to one of three deterministic operators---\texttt{MAINTAIN}, \texttt{PROFILE}, or \texttt{CORRECT}. This read--write loop differs from prior memory systems in two ways. First, every belief item carries a continuous confidence weight $w_t(m)\!\in\![0,1]$ that is reinforced or softly decayed by the operators, instead of relying on discrete commit decisions such as \textsc{Add}/\textsc{Ignore}/\textsc{Update} in O-Mem~\citep{wang2025mem} or binary expiry intervals in EverMemOS~\citep{hu2026evermemos}. Second, \texttt{PROFILE} buffers ambiguous observations as low-confidence hypotheses, letting weak evidence accumulate across turns before it is committed or discarded.

In implementation, each operator pairs a closed-form update rule with a fixed-prompt LLM sub-agent. The sub-agent only identifies which items are relevant and emits one bounded scalar, such as \textsc{evidence\_strength}, \textsc{severity}, or \textsc{initial\_confidence}. The deterministic layer then plugs these scalars into \cref{eq:pcm_maintain,eq:pcm_profile_init,eq:pcm_correct}. Thus the LLM proposes local magnitudes, while the gate, confidence bounds, and closed-form $w$ updates control when and how memory changes accumulate. Algorithm~\ref{alg:precomem} (Appendix~\cref{app:algorithm}) gives the per-turn pipeline.

\subsection{Memory primitives}
\label{sec:memory_primitives}

\textbf{Dialogue and memory state.}\quad
At turn $t=1,2,\ldots$, the user input is $u_t$, the agent response is $y_t$, and $Q_{t-1}$ is a rolling working-memory window. Long-term memory is $\mathcal{M}_t=(\mathcal{B}_t,\mathcal{E}_t,\mathcal{T}_t)$. The belief store $\mathcal{B}_t$ contains editable facts and hypotheses with confidence weights. The episodic cache $\mathcal{E}_t$ stores immutable raw dialogue traces with source pointers. The optional topic table $\mathcal{T}_t$ is only a retrieval prior. This keeps beliefs editable and evidence traceable.

\textbf{Items, confidence, and hypothesis lifecycle.}\quad A memory item $m$ has content $c(m)$, type $\tau(m)\in\{\texttt{FACT},\texttt{HYPOTHESIS}\}$, and retrieval metadata such as embeddings, topic labels, entities, timestamps, and source pointers into $\mathcal{E}_t$. For belief items, $w_t(m)\in[0,1]$ is an operational support score. Episodic traces are fixed at $w_t(e)\equiv 1$ and used only as provenance, not in confidence-weighted Top-$K$ ranking. A \texttt{HYPOTHESIS} is a provisional belief: it is reinforced under supporting evidence, decayed when evidence stays absent, and promoted to \texttt{FACT} when confidence crosses $\tau_{\mathrm{promote}}$. Type transitions are one-way. Contradicted facts are softly decayed by \texttt{CORRECT} (\cref{sec:pcm_evolve}) rather than demoted, yielding low-confidence ``deprecated'' facts that confidence-weighted retrieval naturally deprioritizes. Appendix~\cref{app:fact_pruning} confirms this retention is accuracy-neutral.

\subsection{Memory leveraging: evidence recall and response generation}
\label{sec:pcm_leverage}

In \textbf{Memory Leveraging}, PreCoMem recalls a compact evidence set from long-term memory and leverages it to generate the current response.

\textbf{Evidence recall.}\quad
Given $u_t$ and short-term context $Q_{t-1}$, PreCoMem ranks candidates in $\mathcal{B}_t$ with a confidence-modulated score. The score combines three retrieval signals: semantic similarity, lexical overlap (BM25), and topic relevance. Omitting $Q_{t-1}$ in the notation,
\begin{equation}
\label{eq:pcm_retrieval_score}
s(u_t,m)=w_t(m)\sum_{j=1}^{3}\alpha_j s_j(u_t,m),\quad m\in\mathcal{B}_t,
\end{equation}
where $(s_1,s_2,s_3)=(s_{\mathrm{sem}},s_{\mathrm{fact}},s_{\mathrm{topic}})$ and $\boldsymbol{\alpha}=(\alpha_1,\alpha_2,\alpha_3)^{\top}$ is a convex weight vector. The multiplier $w_t(m)$ biases retrieval toward high-confidence items. The Top-$K$ belief items form $\mathcal{C}_t=\{m_i\}_{i=1}^{K}$ with retrieval scores $\mathbf{s}_t$. Linked episodic traces from $\mathcal{E}_t$ are appended as provenance context, but they do not participate in Top-$K$ ranking. The same scores $\mathbf{s}_t$ are retained for the reliability estimate in \cref{sec:pcm_evolve}.

\textbf{Evidence-conditioned response.}\quad
The recalled evidence texts $\{c(m)\mid m\in\mathcal{C}_t\}$, together with $u_t$ and the short-term window $Q_{t-1}$, condition the response model:
\begin{equation}
\label{eq:pcm_response}
y_t=\texttt{Respond}(u_t,Q_{t-1},\mathcal{C}_t).
\end{equation}
We reuse the same recall outputs $(\mathcal{C}_t,\mathbf{s}_t)$ to perform memory evolution in \cref{sec:pcm_evolve}.

\subsection{Memory evolution: regime-gated belief updating}
\label{sec:pcm_evolve}
Given the recall outputs $(\mathcal{C}_t,\mathbf{s}_t)$ and the new interaction $(u_t,y_t)$,
PreCoMem computes a prediction-error signal adjusted by retrieval reliability,
and applies a regime-gated update to the belief store.
It also writes the new turn back into the episodic cache for future provenance and recall.

\textbf{Retrieval reliability.}\quad
Surprisal computed against recalled evidence can be misleading when recall is uncertain, so we derive a reliability coefficient from the entropy of retrieval scores. Let $\mathbf{s}_t=\{s_i\}_{i=1}^{|\mathcal{C}_t|}$ denote the Top-$K$ retrieval scores returned with $\mathcal{C}_t$, and let $\hat{s}_i = (s_i+\epsilon)/\sum_j(s_j+\epsilon)$ be the smoothed normalization. We define the normalized entropy $H_t$ and the reliability coefficient $\rho_t$ as
\begin{equation}
\label{eq:pcm_recall_reliability}
H_t = \frac{-\sum_i \hat{s}_i\log \hat{s}_i}{\log\max\{|\mathcal{C}_t|,2\}}\in[0,1],\qquad \rho_t = 1-H_t.
\end{equation}
Because $H_t$ is minimized when one item dominates and maximized under near-uniform scores, $\rho_t$ measures score concentration as a proxy for evidence precision. Diffuse recall (small $\rho_t$) downweights deviation signals to avoid over-updating; concentrated recall (large $\rho_t$) preserves them for correction. Thus the gate becomes aggressive only when recall is concentrated and the raw contradiction/deviation signal is large. Since $\rho_t$ enters only multiplicatively, it can defer an update but never trigger one on its own; whether \texttt{CORRECT} fires still depends on the contradiction term below, and Appendix~\cref{app:rho_calibration} verifies empirically that $\rho_t$ tracks evidence precision. The constant $\epsilon>0$ ensures numerical stability.

\textbf{Raw surprisal proxy.}\quad We define:
\begin{equation}
\label{eq:pcm_raw_surprisal}
S_{\mathrm{raw}}(u_t)=\lambda\, d_{\mathrm{sem}}(u_t,\mathcal{C}_t)+(1-\lambda)\, d_{\mathrm{conf}}(u_t,\mathcal{C}_t),
\end{equation}
where $\lambda\in(0,1)$, $d_{\mathrm{sem}}$ is the normalized semantic deviation between $u_t$ and score-weighted recalled evidence, and $d_{\mathrm{conf}}$ is a verifier-based contradiction score quantifying how strongly $u_t$ conflicts with recalled beliefs (see Appendix~\cref{app:surprisal_components}).

\textbf{Effective surprisal.}\quad We define:
\begin{equation}
\label{eq:pcm_effective_surprisal}
S_{\mathrm{eff}}(u_t)=\rho_t\, S_{\mathrm{raw}}(u_t),
\end{equation}
which corresponds to a precision-weighted prediction error that becomes conservative when recall is unreliable.
We use $x_t:=S_{\mathrm{eff}}(u_t)$ as the scalar gate score below. Appendix~\cref{app:online_surrogate} gives an optimization view that interprets $x_t$ as a Monte Carlo surrogate of a long-horizon predictive objective.

\textbf{Regime-gated evolution.}\quad
Given two ordered thresholds $\theta_{\mathrm{low}}<\theta_{\mathrm{high}}$, PreCoMem maps the gate score $x_t$ into three regimes:
\begin{equation}
\label{eq:pcm_gate}
a_t=
\begin{cases}
\texttt{MAINTAIN}, & x_t\le \theta_{\mathrm{low}}\\
\texttt{PROFILE}, & \theta_{\mathrm{low}}<x_t\le \theta_{\mathrm{high}}\\
\texttt{CORRECT}, & x_t>\theta_{\mathrm{high}}.
\end{cases}
\end{equation}

\textbf{Two-threshold gate.}\quad Low, medium, and high $x_t$ correspond to consistent, ambiguous, and reliably contradictory evidence. \texttt{PROFILE} handles the middle regime that a binary maintain/correct rule must either ignore or overwrite. We fix $(\theta_{\mathrm{low}},\theta_{\mathrm{high}})=(0.2,0.6)$ throughout the main experiments to keep behavior controllable and the comparison reproducible across benchmarks; the cost-sensitive break-even derivation, operator ablations, threshold sensitivity, and dynamic-threshold variant are in Appendix~\cref{app:gate_derivation,app:dimobench_operator_ablation,app:theta_sensitivity,app:dynamic_threshold}.

\paragraph{\texttt{MAINTAIN} (reinforcement).} When surprisal is low, PreCoMem strengthens supported beliefs in $\mathcal{C}_t$ via:
\begin{equation}
\label{eq:pcm_maintain}
w_{t+1}(m)=w_t(m)+\eta_t(m)\,(1-w_t(m)),\quad m\in\mathcal{C}_t\cap\mathcal{B}_t,
\end{equation}
where $\eta_t(m)\in(0,1]$ is the per-item \textsc{evidence\_strength} returned by the \texttt{Maintain-Agent}; weakly supported items barely move, while strongly supported ones are pulled toward $1$. The optimization view in Appendix~\cref{app:maintain_derivation} treats $\eta_t(m)$ as a scalar step size.

\paragraph{\texttt{PROFILE} (hypothesis writing).} When surprisal is medium, the observation is novel but not decisively contradictory. PreCoMem generates a low-confidence hypothesis $h$ and writes it as a \texttt{HYPOTHESIS} belief:
\begin{equation}
\label{eq:pcm_profile_init}
\begin{aligned}
h&=\texttt{GenerateHypothesis}(u_t,\mathcal{C}_t),\qquad \tau(h)=\texttt{HYPOTHESIS},\\
b_t^{(c)}&=w_{\min}^{(c)}+\sigma(x_t)\bigl(w_{\max}^{(c)}-w_{\min}^{(c)}\bigr),\qquad
w_0(h)=\tfrac12\,b_t^{(c)}+\tfrac12\,\mathrm{clip}\bigl(q_t,[w_{\min}^{(c)},w_{\max}^{(c)}]\bigr),
\end{aligned}
\end{equation}
where \texttt{GenerateHypothesis} is the \texttt{Profile-Agent} under a fixed prompt, $\sigma(x)=1/(1+e^{-x})$, $c\!\in\!\{\textsc{explicit},\textsc{latent}\}$ is the class label assigned by the agent, and $q_t$ is the agent's \textsc{initial\_confidence}, clipped to the class band $[w_{\min}^{(c)},w_{\max}^{(c)}]$. The gate score thus supplies a reproducible prior $b_t^{(c)}$ inside the band, and the agent scalar can shift $w_0$ by at most half the band width. \emph{Explicit-event} hypotheses paraphrase stated facts or actions and use the band $[0.60,0.70]$. \emph{Latent-inference} hypotheses capture deduced traits, moods, or hidden goals and use $[0.50,0.59]$. These initial values are low relative to stable \texttt{FACT}s, which sit at $w\!\ge\!0.80$ and meet the promotion threshold $\tau_{\mathrm{promote}}\!=\!0.80$. A fresh hypothesis therefore ranks below committed facts in confidence-weighted retrieval (\cref{eq:pcm_retrieval_score}) and still needs repeated support to graduate. In later turns, hypotheses are reinforced under \texttt{MAINTAIN} when supported and decayed when evidence remains absent:
\begin{equation}
\label{eq:pcm_hypothesis_decay}
w_t(h)\leftarrow w_t(h)\exp\bigl(-\gamma\, a_t(h)\bigr),\qquad a_t(h)=t-\ell(h),
\end{equation}
where $\ell(h)$ is the turn at which $h$ was created or last received supporting evidence and $\gamma>0$ is a decay rate. The rule is applied once per turn to unsupported hypotheses: any supporting update resets $\ell(h)$ and skips the decay in that turn, and hypotheses flagged as \textsc{decay\_protection} by the agent (explicit events) are exempt, so the compounding decay over consecutive unsupported turns (cumulative factor $\exp(-\gamma\,n(n+1)/2)$ after $n$ turns; Appendix~\cref{app:decay_horizon}) prunes only guesses that are never revalidated. Hypotheses are promoted to facts once confidence crosses $\tau_{\mathrm{promote}}$, with $0\!\le\!w_{\min}^{(c)}\!<\!w_{\max}^{(c)}\!<\!\tau_{\mathrm{promote}}\!\le\!1$ enforced for every class so that promotion always requires repeated support.

\paragraph{\texttt{CORRECT} (soft correction).}
When surprisal is high and recall is reliable, the new observation likely contradicts existing high-confidence beliefs. PreCoMem records the new observation as a fresh belief node (a \texttt{FACT} with initial weight $0.85$) and separately decays the retrieved beliefs it contradicts. The decay side is:
\begin{equation}
\label{eq:pcm_correct}
w_{t+1}(m)=w_t(m)\exp\!\bigl(-\beta\,x_t\,\zeta_t(m)\bigr),\quad m\in\mathcal{C}_t\cap\mathcal{B}_t,
\end{equation}
where $\beta>0$ controls the correction strength and $\zeta_t(m)\!\in\![0,1]$ is the per-item \emph{contradiction severity} returned by the \texttt{Correct-Agent}; non-contradictory items receive $\zeta_t(m)\!\approx\!0$ and are essentially unchanged, enabling selective correction without extra thresholds. We deliberately separate this context-aware decay magnitude from the fixed NLI contradiction score that feeds $d_{\mathrm{conf}}$ (\cref{eq:pcm_dconf} in Appendix~\cref{app:surprisal_components}): the NLI score drives the gate via $S_{\mathrm{raw}}$, whereas $\zeta_t$ localizes how strongly each retrieved belief should decay. A derivation is provided in Appendix~\cref{app:correct_derivation}.

\paragraph{Trace and topic write-back.}
After generating $y_t$, PreCoMem appends the turn trace $(u_t,y_t)$ to the episodic cache $\mathcal{E}_{t+1}$ with a fixed-capacity eviction policy; optionally, new topics are added to $\mathcal{T}_{t+1}$ when detected, yielding $\mathcal{M}_{t+1}=(\mathcal{B}_{t+1},\mathcal{E}_{t+1},\mathcal{T}_{t+1})$.
When \texttt{PROFILE}/\texttt{MAINTAIN}/\texttt{CORRECT} creates or updates belief items, we record source pointers to supporting traces so beliefs remain editable while evidence remains traceable.

% % ----------------------------
% % Merged main results: LoCoMo (top) + LongMemEval-S (bottom)
% % ----------------------------
% ----------------------------
% Merged main results: LoCoMo (top) + LongMemEval-S (bottom)
% ----------------------------

\begin{table}[!t]
  \caption{Main evaluation on LoCoMo and LongMemEval\mbox{-}S. Accuracies are reported as mean $\pm$ std over three seeds; within each backbone block, \textbf{bold} marks the best score per column and \underline{underline} marks the second-best. \textit{Avg.\ Tokens} reports the average response-model input token usage per query. LongMemEval\mbox{-}S columns are question types: SSU/SSA/SSP = single-session user/assistant/preference, MS = multi-session, KU = knowledge update, and TR = temporal reasoning.}
  \label{tab:main_results}
  \centering
  \begin{small}
    \renewcommand{\arraystretch}{1.06}
    %---- LoCoMo block (resized to \linewidth so widths match) ----
    \resizebox{\linewidth}{!}{%
    \setlength{\tabcolsep}{4.6pt}%
    \begin{tabular}{@{}l c c c c c c@{}}
      \toprule
      \multicolumn{7}{c}{\textbf{LoCoMo} (1{,}540 questions; four splits)} \\
      \midrule
      \textbf{Method} & \textbf{Avg.\ Tokens} &
      \textbf{Single-Hop} & \textbf{Multi-Hop} &
      \textbf{Temporal} & \textbf{Open-Domain} & \textbf{Overall} \\
      \midrule
      \rowcolor{black!8}\multicolumn{7}{c}{\textit{\textsc{Backbone:}~~gpt-4o-mini}} \\
      \cmidrule(lr){1-7}
      MemoryOS  & 5.5k & 66.68\,$\pm$\,0.85 & 54.83\,$\pm$\,0.87 & 52.90\,$\pm$\,0.59 & 32.16\,$\pm$\,0.11 & 59.58\,$\pm$\,0.06 \\
      Mem0      & \textbf{1.6k} & 62.17\,$\pm$\,0.77 & 52.91\,$\pm$\,0.70 & 59.44\,$\pm$\,0.37 & 33.89\,$\pm$\,0.96 & 58.36\,$\pm$\,0.31 \\
      A-Mem     & 2.7k & 45.15\,$\pm$\,0.53 & 23.99\,$\pm$\,0.46 & 53.67\,$\pm$\,0.23 & 30.71\,$\pm$\,0.84 & 42.18\,$\pm$\,0.65 \\
      GAM       & 3.5k & 68.24\,$\pm$\,0.76 & 55.47\,$\pm$\,0.43 & 58.19\,$\pm$\,0.54 & 33.50\,$\pm$\,0.09 & 61.62\,$\pm$\,0.22 \\
      MemBox    & 3.3k & 70.89\,$\pm$\,0.68 & 53.31\,$\pm$\,0.15 & 62.70\,$\pm$\,0.57 & 35.18\,$\pm$\,0.77 & 63.30\,$\pm$\,0.48 \\
      MemOS     & \underline{2.5k} & 77.05\,$\pm$\,0.93 & 68.34\,$\pm$\,0.79 & 61.27\,$\pm$\,0.01 & \underline{39.26\,$\pm$\,0.07} & 70.20\,$\pm$\,0.66 \\
      EverMemOS & 2.6k & \textbf{83.69\,$\pm$\,0.30} & 59.97\,$\pm$\,0.72 & 66.00\,$\pm$\,0.39 & 36.95\,$\pm$\,0.56 & 70.13\,$\pm$\,0.55 \\
      O-Mem     & 2.9k & 78.59\,$\pm$\,0.04 & \textbf{71.08\,$\pm$\,0.34} & 60.20\,$\pm$\,0.91 & 37.41\,$\pm$\,0.87 & 70.88\,$\pm$\,0.22 \\
      TiMem     & 2.8k & 76.64\,$\pm$\,0.60 & 60.43\,$\pm$\,0.80 & \textbf{76.74\,$\pm$\,0.48} & 38.86\,$\pm$\,0.94 & \underline{71.88\,$\pm$\,0.61} \\
      \rowcolor{pcmlightblue}
      \textbf{PreCoMem} & 3.3k & \underline{81.41\,$\pm$\,0.51} & \underline{68.49\,$\pm$\,0.98} & \underline{66.21\,$\pm$\,0.12} & \textbf{39.63\,$\pm$\,0.25} & \textbf{73.57\,$\pm$\,0.28} \\
      \midrule
      \rowcolor{black!8}\multicolumn{7}{c}{\textit{\textsc{Backbone:}~~gpt-4.1}} \\
      \cmidrule(lr){1-7}
      MemoryOS  & 5.2k & 72.21\,$\pm$\,0.38 & 63.66\,$\pm$\,0.18 & 63.10\,$\pm$\,0.83 & 43.33\,$\pm$\,0.88 & 66.02\,$\pm$\,0.25 \\
      Mem0      & \textbf{1.8k} & 72.59\,$\pm$\,0.73 & 62.37\,$\pm$\,0.49 & 67.85\,$\pm$\,0.20 & 42.98\,$\pm$\,0.89 & 67.17\,$\pm$\,0.51 \\
      A-Mem     & 2.9k & 50.63\,$\pm$\,0.09 & 35.96\,$\pm$\,0.39 & 59.50\,$\pm$\,0.24 & 34.30\,$\pm$\,0.54 & 48.90\,$\pm$\,0.22 \\
      GAM       & 3.4k & 70.35\,$\pm$\,0.38 & 67.45\,$\pm$\,0.02 & 68.47\,$\pm$\,0.64 & 39.03\,$\pm$\,0.05 & 68.81\,$\pm$\,0.11 \\
      MemBox    & 3.4k & 76.76\,$\pm$\,0.10 & 67.47\,$\pm$\,0.33 & 68.40\,$\pm$\,0.52 & 41.78\,$\pm$\,0.82 & 71.29\,$\pm$\,0.92 \\
      MemOS     & \underline{2.4k} & 84.57\,$\pm$\,0.70 & 80.14\,$\pm$\,0.56 & 70.79\,$\pm$\,0.43 & \underline{46.93\,$\pm$\,0.87} & 78.62\,$\pm$\,0.76 \\
      O-Mem     & 2.7k & 86.81\,$\pm$\,0.73 & \textbf{84.32\,$\pm$\,0.42} & 68.04\,$\pm$\,0.03 & 42.24\,$\pm$\,0.75 & 80.08\,$\pm$\,0.14 \\
      EverMemOS & 2.5k & \textbf{92.13\,$\pm$\,0.94} & 77.75\,$\pm$\,0.52 & 74.72\,$\pm$\,0.91 & 43.32\,$\pm$\,0.31 & 82.27\,$\pm$\,0.86 \\
      TiMem     & 2.7k & 87.00\,$\pm$\,0.58 & \underline{83.95\,$\pm$\,0.15} & \textbf{85.92\,$\pm$\,0.71} & 45.82\,$\pm$\,0.60 & \underline{84.04\,$\pm$\,0.84} \\
      \rowcolor{pcmlightblue}
      \textbf{PreCoMem} & 3.1k & \underline{90.41\,$\pm$\,0.35} & 83.48\,$\pm$\,0.29 & \underline{84.74\,$\pm$\,0.80} & \textbf{48.23\,$\pm$\,0.65} & \textbf{85.06\,$\pm$\,0.07} \\
      \bottomrule
    \end{tabular}}%

    \vspace{4pt}

    %---- LongMemEval-S block (resized to \linewidth so widths match) ----
    \resizebox{\linewidth}{!}{%
    \setlength{\tabcolsep}{4.0pt}%
    \begin{tabular}{@{}l c c c c c c c c@{}}
      \toprule
      \multicolumn{9}{c}{\textbf{LongMemEval\mbox{-}S} (500 questions; six splits)} \\
      \midrule
      \textbf{Method} & \textbf{Avg.\ Tokens} & \textbf{SSU} & \textbf{SSA} & \textbf{SSP} & \textbf{MS} & \textbf{KU} & \textbf{TR} & \textbf{Overall} \\
      \midrule
      \rowcolor{black!8}\multicolumn{9}{c}{\textit{\textsc{Backbone:}~~gpt-4.1 \quad (gpt-4o-mini results in Appendix~\cref{app:longmemeval_4omini})}} \\
      \cmidrule(lr){1-9}
      MemoryOS  & 6.8k & 62.77\,$\pm$\,0.01 & 63.53\,$\pm$\,0.19 & 48.71\,$\pm$\,0.00 & 59.27\,$\pm$\,0.57 & 58.94\,$\pm$\,0.67 & 54.95\,$\pm$\,0.14 & 58.49\,$\pm$\,0.04 \\
      MemOS     & \textbf{2.6k} & 80.88\,$\pm$\,0.48 & 73.29\,$\pm$\,0.20 & 58.43\,$\pm$\,0.15 & 70.10\,$\pm$\,0.65 & 75.05\,$\pm$\,0.55 & 68.32\,$\pm$\,0.21 & 72.74\,$\pm$\,0.40 \\
      TiMem     & 3.5k & 85.06\,$\pm$\,0.03 & 79.18\,$\pm$\,0.54 & 71.22\,$\pm$\,0.62 & 76.58\,$\pm$\,0.84 & 81.96\,$\pm$\,0.50 & \textbf{89.86\,$\pm$\,0.41} & 79.63\,$\pm$\,0.79 \\
      GAM       & 4.2k & 80.90\,$\pm$\,0.70 & 83.73\,$\pm$\,0.26 & 70.82\,$\pm$\,0.02 & 80.81\,$\pm$\,0.46 & 78.30\,$\pm$\,0.51 & 79.28\,$\pm$\,0.23 & 80.91\,$\pm$\,0.34 \\
      MemBox    & 4.0k & 82.07\,$\pm$\,0.42 & 85.13\,$\pm$\,0.64 & 73.68\,$\pm$\,0.61 & 80.56\,$\pm$\,0.38 & 79.59\,$\pm$\,0.78 & 81.52\,$\pm$\,0.89 & 81.24\,$\pm$\,0.60 \\
      O-Mem     & 3.4k & 91.25\,$\pm$\,0.12 & \underline{86.76\,$\pm$\,0.74} & \underline{74.69\,$\pm$\,0.07} & \underline{83.89\,$\pm$\,0.64} & \underline{87.05\,$\pm$\,0.29} & 82.77\,$\pm$\,0.88 & 85.36\,$\pm$\,0.08 \\
      EverMemOS & \underline{3.1k} & \textbf{96.31\,$\pm$\,0.92} & 83.33\,$\pm$\,0.97 & 72.45\,$\pm$\,0.47 & 80.66\,$\pm$\,0.37 & 86.25\,$\pm$\,0.11 & 79.17\,$\pm$\,0.75 & \underline{86.16\,$\pm$\,0.69} \\
      \rowcolor{pcmlightblue}
      \textbf{PreCoMem}  & 3.8k & \underline{94.09\,$\pm$\,0.98} & \textbf{90.72\,$\pm$\,0.12} & \textbf{83.93\,$\pm$\,0.85} & \textbf{87.44\,$\pm$\,0.99} & \textbf{91.80\,$\pm$\,0.87} & \underline{84.35\,$\pm$\,0.83} & \textbf{89.39\,$\pm$\,0.76} \\
      \bottomrule
    \end{tabular}}%
  \end{small}
\end{table}

% ----------------------------
% Evaluation overview, split for compact main-text placement:
%   Figure 3: (a)--(d) backbone generalization.
%   Wrapfigure: Figure 5(a)--(b) efficiency frontier and Top-K sensitivity.
% ----------------------------
\begin{figure*}[t]
  \centering
  \includegraphics[width=0.99\linewidth]{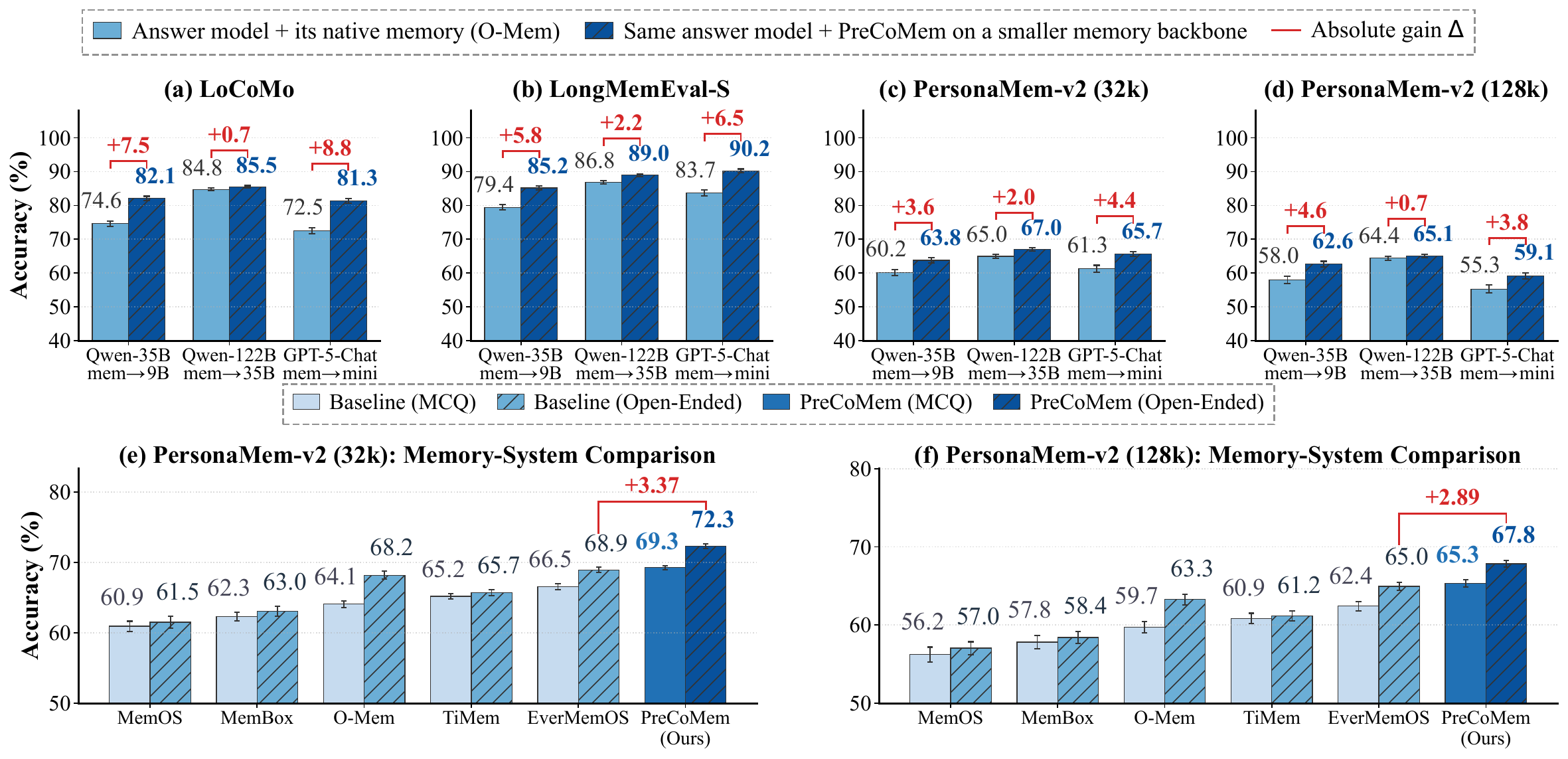}
  \caption{\textbf{Backbone generalization and PersonaMem\mbox{-}v2 system comparison.}
  \textbf{(a)--(d)} Scale-vs-mechanism comparison on LoCoMo, LongMemEval\mbox{-}S, PersonaMem\mbox{-}v2~(32k), and PersonaMem\mbox{-}v2~(128k). Within each group the answer model is fixed (Qwen3.5-35B, Qwen3.5-122B, or GPT-5-Chat): the left bar runs it with its native O-Mem memory, while the right bar keeps the same answer model and delegates memory construction to PreCoMem on a one-tier-smaller backbone (Qwen3.5-9B, Qwen3.5-35B, GPT-5-mini); the protocol is detailed in Appendix~\cref{app:backbone_protocol}.
  \textbf{(e)--(f)} Head-to-head with the five strongest baselines on PersonaMem\mbox{-}v2 under \textit{gpt-4.1}, in both MCQ and Open-Ended formats. Error bars are $\pm$std over three seeds.}
  \label{fig:locomo_eval_overview}
\end{figure*}

\section{Experiments}
\subsection{Experimental setup}
\label{sec:setup}

\textbf{Benchmarks.}\quad
We evaluate on LoCoMo~\citep{maharana2024evaluating} (1,500+ questions in long contexts), LongMemEval\mbox{-}S \citep{wu2024longmemeval} (500 questions covering six question types over five long-term memory abilities), and PersonaMem\mbox{-}v2~\citep{jiang2025personamemv2} ($1{,}000$ personas and $20{,}000+$ implicit preferences, evaluated under both 32k and 128k contexts; \cref{fig:locomo_eval_overview}(c)--(f)). To stress-test belief drift under sparse signals we additionally introduce DiMoBench, which adds controlled non-stationarity and decoy retrieval (\cref{sec:dimobench}). Two third-party benchmarks, AMemGym~\citep{cheng2026amemgym} and LoCoMo\mbox{-}Plus~\citep{li2026locomo}, are evaluated under their official protocols in Appendix~\cref{app:external_benchmarks}.

\textbf{Baselines.}\quad
We compare PreCoMem against O-Mem~\citep{wang2025mem}, MemoryOS~\citep{kang2025memory}, MemOS~\citep{li2025memos}, Mem0~\citep{chhikara2025mem0}, A-Mem~\citep{xu2025mem}, GAM~\citep{yan2025general}, MemBox~\citep{tao2026membox}, TiMem~\citep{li2026timem}, and EverMemOS~\citep{hu2026evermemos}, each under its officially recommended configuration; we standardize the backbone LLM and embedding model across methods whenever applicable.

\textbf{Evaluation protocol.}\quad
We adopt the LLM-as-a-judge protocol from MemOS~\citep{li2025memos}: each answer is scored by a primary judge (\textit{gpt-4.1-mini}) and two auxiliary \textit{gemini-2.5-flash} judges (run with distinct system prompts to reduce judge-correlation) in a blinded setup, with accuracy reported as the average of the three binary judgments~\citep{li2024llms, gu2024survey, huang2025empirical, shi2025judging, ye2025learning}. Appendix~\cref{app:llm_judge_validation} reports human--LLM agreement on LoCoMo, LongMemEval\mbox{-}S, and DiMoBench plus a cross-method spot-check on baseline outputs.

\textbf{Implementation details.}\quad
PreCoMem uses \textit{gpt-4.1} (or \textit{gpt-4o-mini}) for all LLM-based reasoning and memory operations and \texttt{Qwen3-Embedding-0.6B} for dense embeddings across methods. Unless otherwise specified, $k{=}10$, $(\theta_{\mathrm{low}},\theta_{\mathrm{high}})=(0.2,0.6)$, $\lambda{=}0.5$, $\beta{=}1.0$, $\boldsymbol{\alpha}{=}(0.4,0.4,0.2)$, and $\gamma{=}0.05$; sensitivity studies and the adaptive variant are in Appendix~\cref{app:topk_sensitivity,app:theta_sensitivity,app:dynamic_threshold}.

\subsection{Main results}

Main results are reported in \cref{tab:main_results}: the upper block on LoCoMo and the lower block on LongMemEval\mbox{-}S. We make three observations.

(1) \textbf{Overall performance.} PreCoMem achieves the best overall accuracy on both tabulated benchmarks. On LoCoMo it leads the strongest baseline (TiMem) by $+1.69$ with \textit{gpt-4o-mini} ($73.57$ vs.\ $71.88$) and $+1.02$ with \textit{gpt-4.1} ($85.06$ vs.\ $84.04$); both gaps exceed the cross-seed standard deviation of either system reported in \cref{tab:main_results}. On LongMemEval\mbox{-}S it reaches $89.39$ under \textit{gpt-4.1} ($+3.23$ over EverMemOS), and the \textit{gpt-4o-mini} appendix block follows the same pattern ($84.28$, $+1.96$; Appendix~\cref{app:longmemeval_4omini}). On PersonaMem\mbox{-}v2 (\cref{fig:locomo_eval_overview}(e)--(f)) under \textit{gpt-4.1}, PreCoMem reaches $72.31$ Open-Ended at 32k and $67.84$ at 128k, $+3.37$ and $+2.89$ over EverMemOS. The advantage also holds on two third-party benchmarks we did not design: $0.458$ Overall / $0.329$ normalized Memory on AMemGym against $0.430$ / $0.296$ for the strongest system reported there, and $22.4$ on LoCoMo\mbox{-}Plus against $21.05$ for full-context GPT-4o (Appendix~\cref{app:external_benchmarks}). \Cref{fig:locomo_eval_overview}(a)--(d) further shows that, with the answer model held fixed, delegating memory construction to PreCoMem on a one-tier-smaller backbone still beats the larger backbone's native memory in all twelve comparisons.

(2) \textbf{Where gains come from.}
On LoCoMo, the gain concentrates on the Open-Domain split (PreCoMem $48.23$ under \textit{gpt-4.1}, $+1.30$ over MemOS; consistently best under \textit{gpt-4o-mini}), whose questions typically ask for a latent state that is not stated in a single turn and must be inferred from cues spread across sessions (Appendix~\cref{app:attribution}), while EverMemOS, O-Mem, and TiMem retain narrow leads on Single-Hop / Multi-Hop / Temporal under \textit{gpt-4.1} from their dedicated read-path designs.
On LongMemEval\mbox{-}S (\textit{gpt-4.1}), the largest gains occur on the question types whose answers depend on updating or aggregating user-specific memory: \textbf{SSP} (single-session preference, $+9.24$ over O-Mem), \textbf{KU} (knowledge update, $+4.75$ over O-Mem; every KU question carries an old value superseded by a newer one), and \textbf{MS} (multi-session aggregation, $+3.55$). The \textbf{SSA} gain (assistant-side single-session information, $+3.96$) reflects selective memory use: the single target session lies among $48.9$ sessions on average, and confidence-weighted retrieval acts as an implicit reranker. PersonaMem\mbox{-}v2 reproduces this pattern: PreCoMem's lead is largest on anti-stereotypical, dynamic, and ownership-disambiguation categories that most stress write-path quality, and shrinks to under two points on stereotypical or static cues where population priors suffice. PreCoMem's advantage is therefore strongest when answers require selective memory use, belief updating, or cross-session aggregation; a structural analysis attributes the gains to two design properties absent from contemporary baselines---continuous per-belief confidence that buffers ambiguous cues, and a confidence-weighted retrieval score that acts as an implicit reranker (Appendix~\cref{app:structural_comparison}).

(3) \textbf{Accuracy--efficiency trade-off.} \Cref{fig:hyperparam_scalability}(a) shows the LoCoMo performance--cost frontier as we vary $k$: PreCoMem peaks at $k\!=\!10$ with moderate token usage, while larger $k$ yields diminishing returns. Supporting diagnostics on hypothesis lifecycle, Top-$k$ sensitivity, and memory footprint are reported in Appendix~\cref{app:hypothesis_lifecycle,app:topk_sensitivity,app:cross_system_footprint}.

\subsection{Ablation studies and belief-drift adaptation}
\label{sec:dimobench}

\begin{figure*}[t]
  \centering
  \includegraphics[width=\linewidth]{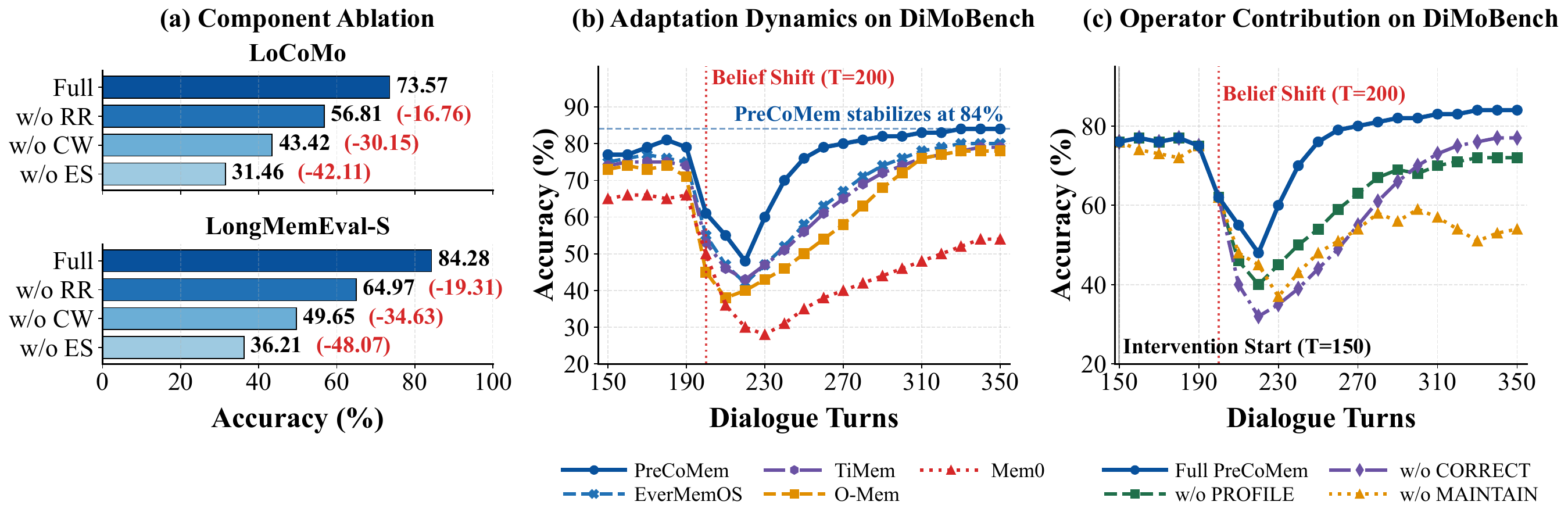}
  \caption{\textbf{Component ablation and DiMoBench dynamics.}
  \textbf{(a)} Component ablation on LoCoMo and LongMemEval\mbox{-}S.
  \textbf{(b)} Moving-average accuracy under implicit drift on DiMoBench.
  \textbf{(c)} Operator-level ablation on DiMoBench.}
  \label{fig:ablation_row}
\end{figure*}

\textbf{Component ablation.}\quad
We ablate three core components while keeping the backbone, prompts, and hyperparameters fixed. ES removes the effective-surprisal gate by setting the gate score to $0$; RR removes retrieval-reliability adjustment via $\rho_t\!\equiv\!1$; CW removes confidence weighting globally, in retrieval and in every lifecycle test, by setting all belief weights to $1$. \Cref{fig:ablation_row}(a) reports LLM-judge accuracy on LoCoMo and LongMemEval\mbox{-}S. Removing ES causes the largest drop on LoCoMo ($73.57\!\to\!31.46$; similar on LongMemEval\mbox{-}S): with the gate score zeroed every turn is routed to \texttt{MAINTAIN}, so this reinforce-only variant marks the lower bound of \emph{surprisal-gated evolution}, the primary mechanism under non-stationarity, while route-matched controls that isolate the gate's marginal value are reported in Appendix~\cref{app:route_controls}. Removing CW ($\!\to\!43.42$) and RR ($\!\to\!56.81$) also substantially reduces accuracy, with the same trend on LongMemEval\mbox{-}S. Letting the same LLM emit the entire write decision end-to-end instead of bounded scalars lowers accuracy by $2.75$ / $3.15$ points on LoCoMo / LongMemEval\mbox{-}S and raises the cross-seed standard deviation $20\times$ on LoCoMo (under \textit{gpt-4.1}; Appendix~\cref{app:direct_write}). Appendix~\cref{app:raw_surprisal_ablation} shows the NLI contradiction term drives most gains.

\textbf{DiMoBench setup.}\quad
\textbf{DiMoBench} (Dialogue Implicit Memory Drift Benchmark) is a controlled diagnostic setting for gradual attitude and preference drift. A user's latent stance reverses around a change point, while keyword-matched chatter and entity confounders pollute the stream so naive retrieval and overwrite-style updates become brittle. The released suite has \emph{five independent} $350$-turn dialogues, each driven by a distinct persona, with $200$ QA pairs per dialogue ($1{,}000$ total; results averaged across personas); DiMoBench\mbox{-}XL extends this to $1050$ turns and three sequential drifts, with 420 QA pairs each, 2,100 in total (Appendix~\cref{app:dimobench,app:dimobench_xl}).

\textbf{Update dynamics.}\quad \Cref{fig:ablation_row}(b) plots LLM-judge accuracy across dialogue turns. Before the change point all methods are stable; afterwards PreCoMem experiences a transient drop but recovers fastest and stabilizes at $84\%$, while EverMemOS, TiMem, and O-Mem plateau at 78--80\% and Mem0 fails to recover ($54\%$). End-to-end traces are provided in Appendix~\cref{app:end2end_trace}.

\textbf{Operator ablation on DiMoBench.}\quad
Ablating each operator individually (\cref{fig:ablation_row}(c)) yields qualitatively distinct failure modes: removing \texttt{PROFILE} caps the asymptote at $72\%$ (vs.\ $84\%$); removing \texttt{CORRECT} delays recovery (turn $320$ vs.\ $250$) and lowers the plateau to $77\%$; removing \texttt{MAINTAIN} fails to stabilize within the horizon ($54\%$ at the end), confirming that the three operators play distinct stabilizing roles. Appendix~\cref{app:dimobench_operator_ablation} provides the full analysis, and the same three qualitative signatures reproduce on the third-party AMemGym benchmark (Appendix~\cref{app:amemgym}).

\subsection{Hyperparameter analysis}
\label{sec:hyperparam_k}

We sweep the retrieval budget $k\!\in\!\{1,\ldots,20\}$ on LoCoMo (\textit{gpt-4.1}). Accuracy rises sharply through $k\!\approx\!6$, peaks at $k\!=\!10$ (Overall $85.06\%$), then drops mildly. Larger $k$ flattens the Top-$k$ score distribution, lowering $\rho_t$ and dampening the effective surprisal $x_t$ that drives \texttt{CORRECT}. It also increases the chance that one retrieved item spuriously appears contradictory under the max-based contradiction component $d_{\mathrm{conf}}$. We therefore fix $k\!=\!10$; the full sweep and retrieval-latency profile are reported in Appendix~\cref{app:topk_sensitivity,app:latency_analysis}. Because the write path acts only on retrieved beliefs, Appendix~\cref{app:retrieval_dependence} further measures how often the belief due for an update actually enters the Top-$k$ and what a miss costs.

\subsection{Memory scalability}
\label{sec:memory_scalability}

\begin{figure}[t]
  \centering
  \includegraphics[width=\linewidth]{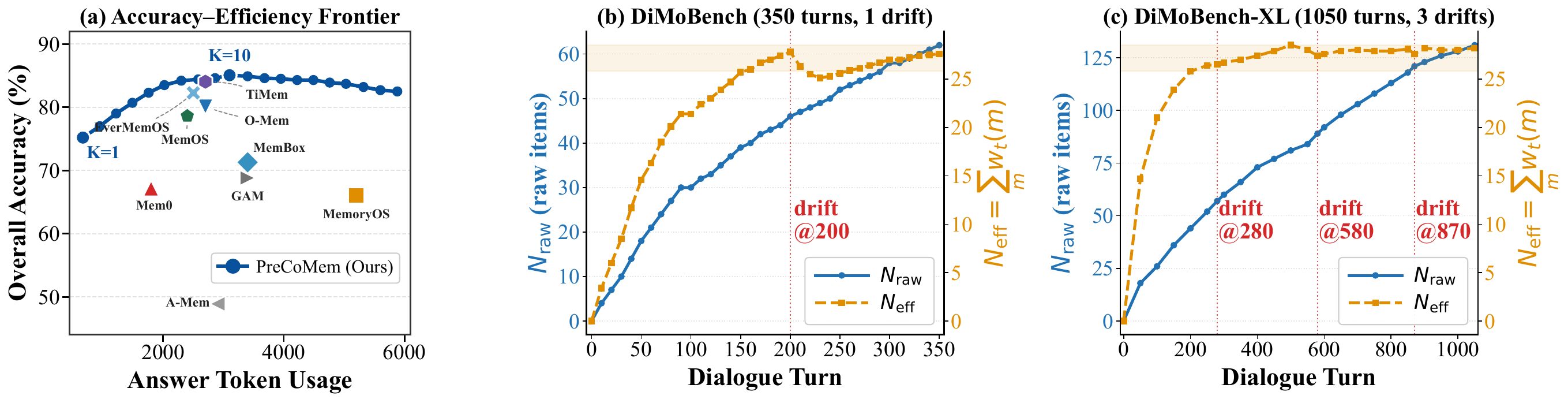}
  \caption{\textbf{Hyperparameter and memory-scalability analysis.}
  \textbf{(a)} LoCoMo accuracy--efficiency frontier as the retrieval budget $k$ varies; overall accuracy peaks at $k\!=\!10$, beyond which additional context yields diminishing returns and dilutes retrieval reliability $\rho_t$.
  \textbf{(b, c)} Memory scalability under long-horizon belief drift: raw memory grows linearly, while effective memory quickly plateaus on both DiMoBench and DiMoBench\mbox{-}XL (panel (c) shows one representative persona).}
  \label{fig:hyperparam_scalability}
\end{figure}

A natural concern for any long-term memory system is whether storage and retrieval cost scale gracefully with the conversation length. We examine this on three settings spanning natural and stress-test horizons: the LoCoMo benchmark (ten multi-session dialogues, mean $\bar{N}\!=\!588$ turns), DiMoBench (350 turns, 1 drift), and DiMoBench\mbox{-}XL (1050 turns, 3 drifts; five personas, of which a representative one is plotted, the others staying within $\pm1.4$ of it; Appendix~\cref{app:dimobench_xl}). Two quantities are tracked every 10 turns on DiMoBench and every 50 turns on DiMoBench\mbox{-}XL:
\begin{itemize}[leftmargin=1.3em,topsep=0pt,itemsep=0pt]
\item \emph{Raw count} $N_{\mathrm{raw}}(t)=|\mathcal{B}_t|$: the number of belief items physically stored, including \emph{deprecated facts}. Deprecated facts are previously committed \texttt{FACT}s whose confidence has been softly decayed below $0.1$ by repeated \texttt{CORRECT} updates (\cref{eq:pcm_correct}); we keep them in $\mathcal{B}_t$ as provenance hooks rather than deleting them.
\item \emph{Effective count} $N_{\mathrm{eff}}(t)=\sum_{m\in\mathcal{B}_t} w_t(m)$: the confidence-weighted total mass. Intuitively, this is the equivalent number of fully-supported items the agent ``acts on'' through the confidence-weighted retrieval score in \cref{eq:pcm_retrieval_score}. A deprecated fact at $w\!=\!0.05$ contributes only $0.05$ to $N_{\mathrm{eff}}$, which matches its negligible influence on Top-$K$ ranking.
\end{itemize}
\Cref{fig:hyperparam_scalability}(b, c) shows $N_{\mathrm{raw}}$ grows roughly linearly with the dialogue, but $N_{\mathrm{eff}}$ saturates by turn 200 (at $25.8$--$28.5$ on the representative persona, with the other personas within $\pm1.4$ of it) and, apart from a transient dip at each drift, stays in that band. Moreover, deprecated facts occupy only $1.0\%$ of Top-10 slots on average even after 1050 turns and three drifts (Appendix~\cref{tab:memory_scalability}). Confidence weighting therefore turns linearly growing storage into a roughly constant retrieval surface.

\begin{table}[t]
\centering
\small
\renewcommand{\arraystretch}{1.10}
\caption{\textbf{Memory footprint vs.\ accuracy on the ten LoCoMo dialogues} (mean turns $\bar{N}\!=\!588$). All numbers are measured under the same backbone (\textit{gpt-4o-mini}) and embedding model. ``Per-item conf.'' indicates whether the system maintains a continuous confidence weight per stored item; under discrete commit decisions, the effective count $N_{\mathrm{eff}}$ and the average Top-10 weight are not defined. PreCoMem (A1) is the optional pruning configuration that drops items with $w_t\!<\!0.05$.}
\label{tab:memory_footprint}
\resizebox{\linewidth}{!}{%
\setlength{\tabcolsep}{4.5pt}%
\begin{tabular}{@{}l c c c c c@{}}
\toprule
\textbf{Method} & Per-item conf.\ & $N_{\mathrm{raw}}^{\mathrm{end}}$ & $N_{\mathrm{eff}}^{\mathrm{end}}$ & Top-10 avg.\ $w$ & LoCoMo Overall \\
\midrule
O-Mem~\citep{wang2025mem}                  & \ding{55} (\textsc{Add}/\textsc{Ignore}/\textsc{Update}) & 232 & --   & --   & 70.88 \\
EverMemOS~\citep{hu2026evermemos}          & \ding{55} ($[t_s,t_e]$ expiry)& 264 & --   & --   & 70.13 \\
\rowcolor{pcmlightblue}
\textbf{PreCoMem (A0, default)}           & \ding{51} ($w_t\!\in\![0,1]$) & 287 & 95.7 & 0.81 & \textbf{73.57} \\
\rowcolor{pcmlightblue!55}
\textbf{PreCoMem (A1, $w<0.05$ prune)}    & \ding{51} ($w_t\!\in\![0,1]$) & \textbf{168} & 95.6 & 0.83 & 73.51 \\
\bottomrule
\end{tabular}}%
\end{table}

\Cref{tab:memory_footprint} extends this picture to physical storage on LoCoMo, where we measure the final $\mathcal{B}_t$ size of each system under matched backbone and embedding protocols. By default (A0) PreCoMem keeps deprecated facts as provenance hooks for traits that may resurface later, since confidence-weighted retrieval already filters them out of the Top-$K$; the optional A1 configuration ($w\!<\!0.05$ prune) trades this provenance for tighter storage and matches A0 within seed variance (Appendix~\cref{app:fact_pruning}). Both O-Mem and EverMemOS commit each interaction through a discrete decision (\textsc{Add}/\textsc{Ignore}/\textsc{Update} in O-Mem; episodic-trace formation with a binary $[t_{\mathrm{start}},t_{\mathrm{end}}]$ validity in EverMemOS), so neither maintains a continuous per-item confidence and $N_{\mathrm{eff}}$ is undefined. Even on raw count alone, A1 stores fewer items than both baselines while still leading on accuracy by $+2.6$ and $+3.4$ overall points, so storage remains bounded in both the soft sense ($N_{\mathrm{eff}}$ plateau) and the hard sense. This suggests the memory gain is not purchased by unchecked accumulation.

\section{Conclusion}
\label{sec:conclusion}

PreCoMem treats long-horizon conversational memory not as a transcript to be stored, but as confidence-weighted beliefs about the user's state that must be continually checked. A reliability-adjusted surprisal signal routes each turn into \texttt{MAINTAIN} (reinforce supported beliefs), \texttt{PROFILE} (buffer ambiguous cues as low-confidence hypotheses), or \texttt{CORRECT} (softly decay contradicted facts). The \texttt{PROFILE} waiting room is the central design choice: it lets evidence accumulate before commitment, so the agent stays consistent under noise without freezing under genuine drift. Across LoCoMo, LongMemEval\mbox{-}S, PersonaMem\mbox{-}v2, and DiMoBench, this discipline yields the best overall accuracy while keeping memory bounded over a thousand turns; qualitative case studies (Appendix~\cref{app:case_study}) show calibrated recommendations beyond retrieval-only baselines. We see continuous confidence and explicit hypothesis lifecycles as a useful baseline for long-term memory systems, where \textbf{what to commit matters as much as what to recall}. Our evaluation covers English dialogue benchmarks; other languages and modalities are left for future work.

\clearpage
% Camera-ready requirement (Main Track Handbook): funding statement and
% competing-interests disclosure, both confirmed by the authors on 2026-10-06.
\begin{ack}
We thank the anonymous reviewers and the area chair for their constructive feedback.

\paragraph{Funding.}
This research is supported by the National Natural Science Foundation of China under Grant 62406163, the Tianjin Science and Technology Plan Project under Grant 25ZXSFSN00150, and the Fundamental Research Funds for the Central Universities under Grant 63261039.

\paragraph{Competing interests.}
The authors declare no competing interests.
\end{ack}

\bibliography{paper}
\bibliographystyle{unsrtnat}

% APPENDIX
\newpage
\appendix
%\onecolumn
% Appendix table of contents. Main-text sections are suppressed via the tocdepth
% switch written to the .toc file at \begin{document} (see main.tex); the switch
% below re-enables entries down to subsection level for the appendix only.
\addtocontents{toc}{\protect\setcounter{tocdepth}{2}}
\renewcommand{\contentsname}{Appendix contents}
\tableofcontents

\section{Related work}
\label{sec:related}

\paragraph{Long-term memory for conversational LLMs.}
Early work attacks the fixed context window with externalized hierarchical or virtual memory~\citep{zhong2024memorybank,packer2023memgpt}. More recent systems either organize the store more aggressively or learn richer write rules. \emph{Structured organizations} include interlinked note stores with associative links~\citep{xu2025mem}, time-aware knowledge graphs~\citep{rasmussen2025zep}, OS-style hierarchical scheduling~\citep{li2025memos,hu2026evermemos}, and dedicated memory-box layouts that segment a user's history into typed shards~\citep{tao2026membox}. \emph{Learning-based update rules} optimize memory operations against downstream rewards~\citep{wang2025memalpha,yu2025memagent,yan2025general}, while \emph{interpretable triggers} target a single failure mode such as preference change~\citep{sun2025preference}, redundancy~\citep{liu2026simplemem,li2026timem}, controllable reliance~\citep{tian2026controllable}, or affective state~\citep{lu2025dynamic}. A systematic gap shared by these systems is the \emph{stability--plasticity} regime: append-only memories~\citep{xu2025mem} treat every observation as equally valid and dilute retrieval; overwrite memories compress aggressively but lose information that later turns out to be relevant~\citep{chhikara2025mem0}; OS-style memories~\citep{li2025memos,hu2026evermemos} schedule access well but still fall back on importance heuristics for writes; and reinforcement-trained memories~\citep{wang2025memalpha,yu2025memagent,yan2025general} face sparse credit assignment, since most reward signal is several turns downstream of the responsible read/write decision. PreCoMem is complementary to all of these: it treats the write path as a single \emph{regime-gated} decision driven by a calibrated prediction-error signal, instead of relying on recency heuristics or task-specific reward shaping.

\paragraph{Predictive coding and surprisal as write signals.}
Predictive coding theory holds that perceptual systems update internal models when observations deviate from prediction~\citep{friston2010free,clark2013whatever}, and information theory gives this deviation a precise scalar form: surprisal~\citep{levy2008expectation,meister2020if}. The same logic has long shaped accounts of episodic memory: novel events are encoded preferentially, and reactivation drives reconsolidation of related traces~\citep{ranganath2003neural,sinclair2019prediction}. Translating this into long-term conversational memory requires two extra ingredients absent from prior memory systems. First, surprisal must be \emph{calibrated against retrieval reliability}, otherwise diffuse recall produces noisy update signals~\citep{kuhn2023semantic,yoran2023making}. Second, surprisal must be \emph{regime-gated}, since consistent, ambiguous, and reliably contradictory evidence call for qualitatively different memory edits. PreCoMem operationalizes both ingredients: $S_{\mathrm{eff}}$ multiplies raw surprisal by an entropy-based reliability coefficient, and a two-threshold gate maps the result to \texttt{MAINTAIN}/\texttt{PROFILE}/\texttt{CORRECT}.

\paragraph{Benchmarking long-term memory.}
LoCoMo~\citep{maharana2024evaluating} provides 1{,}500$+$ questions over multi-session dialogues with explicit single-hop, multi-hop, temporal, and open-domain splits. LongMemEval\mbox{-}S~\citep{wu2024longmemeval} tests five core abilities (information extraction, multi-session reasoning, knowledge updates, temporal reasoning, abstention) with 500 questions over chat histories of roughly $115$k tokens each. PersonaMem\mbox{-}v2~\citep{jiang2025personamemv2} stresses \emph{implicit} personalization across $1{,}000$ personas and $20{,}000+$ preferences, where most signal is buried in everyday task-oriented turns. Two more recent benchmarks move closer to the write path: AMemGym~\citep{cheng2026amemgym} evaluates memory on-policy against a simulated user whose latent task state changes across periods, and LoCoMo\mbox{-}Plus~\citep{li2026locomo} probes implicit cognitive constraints (causal, state, goal, and value) over long dialogues; we report both in \cref{app:external_benchmarks}. EvalMem~\citep{liu2026evalmem} complements these end-task benchmarks with an operation-level diagnostic framework that scores individual memory operations such as adding, updating, and deleting items. These benchmarks probe the read path or the end result of memory updates; none of them, however, isolates the \emph{write-path} decision under continuous, gradual drift with adversarial retrieval interference. Our companion benchmark \textbf{DiMoBench} fills this gap with five independent $350$-turn dialogues, one per persona, whose latent attitudes reverse gradually across change points; an extended variant, \textbf{DiMoBench\mbox{-}XL}, contains five independent $1050$-turn dialogues, each with three sequential drifts.

\section{Additional method details}
\label{app:method_details}

\subsection{From expected predictive loss to a per-turn on-policy loss}
\label{app:online_surrogate}

\paragraph{Conceptual predictive loss.}
Let $h_t=(u_{\le t},y_{<t})$ denote the dialogue history up to turn $t$. The conceptual predictive loss
\begin{equation}
\label{eq:pcm_pred_loss}
\mathcal{L}_{\mathrm{pred}}(t)=\mathbb{E}_{u\sim p(u\mid h_t)}\!\left[S_{\mathrm{eff}}(u)\right]
\end{equation}
captures the expected effective surprisal under the (unknown) interaction distribution $p(u\mid h_t)$, and the long-horizon goal is to update the memory state to reduce future expected losses,
\begin{equation}
\label{eq:pcm_evolution_objective}
\min_{\mathcal{M}_{t+1}}\;\mathbb{E}\!\left[\mathcal{L}_{\mathrm{pred}}(t+k)\mid \mathcal{M}_{t+1}\right],\qquad k\ge 1.
\end{equation}

\paragraph{Monte Carlo surrogate.} Because $p(u\mid h_t)$ is generally unknown, under a fixed response policy we treat the observed turn $u_t$ as a Monte~Carlo sample and use the estimator
\begin{equation}
\label{eq:app_mc_pred_loss}
\widehat{\mathcal{L}}_{\mathrm{pred}}(t):=S_{\mathrm{eff}}(u_t).
\end{equation}
Defining the per-turn on-policy loss $x_t:=\widehat{\mathcal{L}}_{\mathrm{pred}}(t)$ yields an unbiased estimate
\begin{equation}
\label{eq:app_unbiased}
\mathbb{E}\!\left[x_t\mid h_t\right]=\mathcal{L}_{\mathrm{pred}}(t).
\end{equation}
In the main text we write $S_{\mathrm{eff}}(u_t)$ without explicitly conditioning on the memory state; implicitly, $S_{\mathrm{eff}}$ depends on the retrieved evidence produced by $\texttt{Retrieve}(\mathcal{M}_t,u_t)$.

\paragraph{Greedy regularized surrogate for memory updates.} The long-horizon objective in \cref{eq:pcm_evolution_objective} can be approximated on-policy by folding ``future benefit'' into a myopic, regularized surrogate:
\begin{equation}
\label{eq:app_step_surrogate}
\mathcal{M}_{t+1}\approx \arg\min_{\mathcal{M}}\;x_t(\mathcal{M}) + D(\mathcal{M},\mathcal{M}_t) + \Omega(\mathcal{M}),
\end{equation}
where $x_t(\mathcal{M})$ denotes the effective-surprisal loss one would obtain if evaluated with memory state $\mathcal{M}$, $D$ is an edit/stability cost (e.g., penalizing large changes or changes to high-confidence beliefs), and $\Omega$ is a capacity/complexity penalty that discourages unbounded growth of provisional beliefs.
Concretely, we instantiate $\Omega$ with a lightweight pruning rule for the hypothesis pool: belief items with $\tau(m)=\texttt{HYPOTHESIS}$ are removed once their confidence falls below a small threshold (we use $w_t(m)<0.1$), which controls memory size while retaining only actively supported hypotheses.
In contrast, belief items with $\tau(m)=\texttt{FACT}$ are not directly deleted; due to the one-way promotion scheme in \Cref{sec:memory_primitives}, outdated facts are instead handled by soft-decay correction that reduces their confidence and thus deprioritizes them under confidence-weighted retrieval.
The Memory Evolution phase instantiates a discrete, interpretable approximation of \cref{eq:app_step_surrogate}: action selection corresponds to cost-sensitive choice (\cref{app:gate_derivation}), while operator updates admit simple optimization views (\cref{app:maintain_derivation,app:correct_derivation}).

\paragraph{Why low-confidence facts are retained rather than pruned.}
We deliberately apply the $w_t(m)<0.1$ pruning rule only to hypotheses, not to facts. Two design considerations motivate this asymmetry. First, because retrieval is confidence-weighted, an outdated fact with $w\!\approx\!0.02$ contributes roughly 50$\times$ less to the retrieval score than a currently supported fact with $w\!\approx\!1$; such items are naturally pushed out of the Top-$K$ evidence set by currently supported beliefs and become \emph{functionally inactive} under confidence-weighted retrieval even though they remain in storage. Second, long-horizon dialogue often revisits past user traits after long gaps (e.g., a user stops drinking coffee for months and then mentions it again); retaining the deprecated fact at low confidence preserves the reasoning chain and episodic-cache links needed to recover such context, whereas hard deletion would force the system to rebuild that trait from scratch. The resulting \emph{effective memory}, i.e., the subset of items that can still influence Top-$K$ retrieval, therefore remains controlled even if raw storage accumulates history. Under truly extreme horizons, raw storage can be further bounded by a timestamped time-to-live on long-deprecated facts without modifying the core operators; we leave this as an engineering safeguard rather than a core mechanism.

\subsection{Retrieval scoring and surprisal component definitions}
\label{app:retrieval_scoring}

\paragraph{Retrieval scoring.}
Let $\mathbf{e}(\cdot)$ be a text embedding function and let $\mathcal{P}_t$ be the merged candidate pool recalled from $\mathcal{B}_t$ and $\mathcal{E}_t$. We use three reproducible relevance signals:
\begin{align}
s_{\mathrm{sem}}(u_t,m) &= \tfrac12\left(1+\cos(\mathbf{e}(u_t),\mathbf{e}(c(m)))\right)\in[0,1],\label{eq:pcm_sem_score}\\
s_{\mathrm{fact}}(u_t,m) &= \frac{\mathrm{BM25}(u_t,c(m))}{\max_{m'\in\mathcal{P}_t}\mathrm{BM25}(u_t,c(m'))+\epsilon}\in[0,1],\label{eq:pcm_fact_score}
\end{align}
where $\epsilon>0$ is a stability constant. For topical relevance, we maintain topic prototypes $\{\mathbf{t}_z\}_{z=1}^{Z_t}$ (e.g., $\mathbf{t}_z=\mathbf{e}(\ell_z)$ for topic description $\ell_z$) and define
\begin{equation}
\label{eq:pcm_topic_score}
\begin{aligned}
s_{\mathrm{topic}}(u_t,m) &= p(z(m)\mid u_t),\\
p(z\mid x) &= \operatorname{softmax}_z\!\left(\cos(\mathbf{e}(x),\mathbf{t}_z)\right).
\end{aligned}
\end{equation}
We assign each item a topic label at write time as $z(m)=\arg\max_{z}p(z\mid c(m))$, and $Z_t$ denotes the number of topics at turn $t$.

\paragraph{Surprisal components.}\label{app:surprisal_components}
Let $\hat{s}_i$ be the normalized retrieval weights from \cref{eq:pcm_recall_reliability}. We compute a score-weighted evidence embedding
\begin{equation}
\label{eq:pcm_evidence_embed}
\mathbf{e}(\mathcal{C}_t)=\sum_{i=1}^{K}\hat{s}_i\,\mathbf{e}(c(m_i)),
\end{equation}
and define semantic deviation as a cosine distance:
\begin{equation}
\label{eq:pcm_dsem}
d_{\mathrm{sem}}(u_t,\mathcal{C}_t)=\tfrac12\left(1-\cos(\mathbf{e}(u_t),\mathbf{e}(\mathcal{C}_t))\right)\in[0,1].
\end{equation}
For contradiction, let
\begin{equation}
\label{eq:pcm_pi}
\pi_t(m):=\mathrm{Verif}_{\mathrm{conf}}(u_t,c(m))\in[0,1]
\end{equation}
be the \emph{calibrated} contradiction probability predicted by a fixed NLI verifier.
In all experiments, $\mathrm{Verif}_{\mathrm{conf}}$ is instantiated as the \texttt{cross-encoder/nli-deberta-v3-base} cross-encoder:
given the premise--hypothesis pair $(c(m),u_t)$, let $\mathbf{z}(u_t,c(m))\in\mathbb{R}^3$ denote its logits over $\{\text{entail},\text{neutral},\text{contr}\}$.
We fit a scalar temperature $T>0$ on the MultiNLI (MNLI) matched development split and define
$\pi_t(m)=\operatorname{softmax}(\mathbf{z}(u_t,c(m))/T)_{\text{contr}}$, and set
\begin{equation}
\label{eq:pcm_dconf}
d_{\mathrm{conf}}(u_t,\mathcal{C}_t)=\max_{m_i\in\mathcal{C}_t}\pi_t(m_i)\in[0,1].
\end{equation}

\paragraph{Memory item fields.}\label{app:memory_item_fields}
A memory item $m$ is a typed belief node (\texttt{FACT} or \texttt{HYPOTHESIS}) with content $c(m)$, confidence weight $w_t(m)\in[0,1]$, and the retrieval metadata listed in \cref{sec:memory_primitives}; the three metadata fields are constructed as follows. \emph{Entities and keywords} are proposed by the write-path sub-agent at write time (the \texttt{Profile-Agent}'s \textsc{evidence\_keywords}); no separate entity-recognition model is involved, and they are stored as a structured field that is used only by the lexical term $s_{\mathrm{fact}}$ of \cref{eq:pcm_retrieval_score}. \emph{Topic labels} are assigned at write time as $z(m)=\arg\max_z p(z\mid c(m))$ with the prototypes of \cref{eq:pcm_topic_score}; the topic table $\mathcal{T}_t$ keeps one running-mean centroid per topic and acts only as the topic prior $s_{\mathrm{topic}}$. \emph{Dense embeddings} are computed solely from the belief text, $\mathbf{e}(c(m))$ with \texttt{Qwen3-Embedding-0.6B} (normalized); entity and topic information never enter the embedding and contribute only through their own terms. Episodic traces are embedded from their own dialogue text and serve as provenance only. \Cref{app:memory_store_example} shows a serialized snapshot.

\paragraph{Cold start.}\label{app:cold_start}
When the belief store is still empty, retrieval returns no evidence and the surprisal terms above are undefined. By convention the first observations of a user are then handled by \texttt{PROFILE}: they enter the store as provisional hypotheses that later turns can reinforce, correct, or let decay, exactly as for a populated store. \Cref{app:cold_start_stats} reports how reliably first mentions are written under this convention.

\begin{algorithm}[t]
\caption{\textbf{PreCoMem per-turn pipeline (hybrid execution).} Deterministic steps (retrieval, $\rho_t$, $S_{\mathrm{raw}}$, gate, $w$ updates, lifecycle) are interleaved with three regime-specific LLM sub-agents that supply per-item targets and bounded scalar scores. Equation references point to the main text and, for $d_{\mathrm{sem}}$ and $d_{\mathrm{conf}}$, to \cref{app:retrieval_scoring}.}
\label{alg:precomem}
\small
\begin{algorithmic}[1]
\Require $(u_t,\, Q_{t-1},\, M_t)$ with $M_t=(B_t, E_t, T_t)$
\Ensure $(y_t,\, M_{t+1})$
\Statex \textit{// Phase A: Memory Leveraging (deterministic)}
\State $(C_t, s_t) \gets \textsc{Retrieve}(M_t, u_t)$ \Comment{\cref{eq:pcm_retrieval_score}: $w_t$-modulated tri-signal}
\State $y_t \gets \textsc{Respond}(u_t, Q_{t-1}, C_t)$ \Comment{\cref{eq:pcm_response}}
\Statex \textit{// Phase B: Reliability + Surprisal (deterministic)}
\State $\rho_t \gets 1 - \textsc{NormalizedEntropy}(s_t)$ \Comment{\cref{eq:pcm_recall_reliability}}
\State $d_{\mathrm{sem}} \gets 0.5\,(1-\cos(e(u_t), e(C_t)))$ \Comment{\cref{eq:pcm_dsem}}
\State $d_{\mathrm{conf}} \gets \max_{m \in C_t} \textsc{NLI\_contradict}(u_t, c(m))$ \Comment{\cref{eq:pcm_dconf}}
\State $S_{\mathrm{raw}} \gets \lambda\, d_{\mathrm{sem}} + (1-\lambda)\, d_{\mathrm{conf}}$ \Comment{\cref{eq:pcm_raw_surprisal}}
\State $x_t \gets \rho_t \cdot S_{\mathrm{raw}}$ \Comment{\cref{eq:pcm_effective_surprisal}}
\Statex \textit{// Phase C: Regime-gated update (hybrid: agent + formula)}
\If{$x_t \le \theta_{\mathrm{low}}$} \Comment{\texttt{MAINTAIN} regime}
  \State $\{(m_i, \mathrm{evidence\_strength}_i, \mathrm{refinement}_i)\} \gets \textsc{MaintainAgent}(C_t, u_t, x_t)$
  \For{each $(m_i, \eta_i, \mathrm{ref}_i)$}
    \State $\eta_t(m_i) \gets \mathrm{evidence\_strength}_i$
    \State $w_{t+1}(m_i) \gets w_t(m_i) + \eta_t(m_i)\,(1 - w_t(m_i))$ \Comment{\cref{eq:pcm_maintain}}
    \If{$\mathrm{ref}_i$} \State \textsc{AttachRefinement}$(m_i, \mathrm{ref}_i)$ \EndIf
  \EndFor
\ElsIf{$x_t \le \theta_{\mathrm{high}}$} \Comment{\texttt{PROFILE} regime}
  \State $(\mathrm{class}_c, \mathrm{content}, q_t, \mathrm{protect}) \gets \textsc{ProfileAgent}(u_t, C_t, x_t)$ \Comment{band $[w_{\min}^c, w_{\max}^c]$ from $\mathrm{class}_c$}
  \State $h \gets \textsc{NewHypothesis}(\mathrm{content},\ \mathrm{protect})$
  \State $b_t \gets w_{\min}^c + \sigma(x_t)\,(w_{\max}^c - w_{\min}^c)$;\quad $w_0(h) \gets \tfrac12\, b_t + \tfrac12\,\mathrm{clip}(q_t, [w_{\min}^c, w_{\max}^c])$ \Comment{\cref{eq:pcm_profile_init}}
  \State $B_{t+1} \gets B_t \cup \{h\}$
\Else \Comment{\texttt{CORRECT} regime}
  \State $\{(m_i, \mathrm{severity}_i, \mathrm{op}_i)\} \gets \textsc{CorrectAgent}(C_t, u_t, x_t)$
  \For{each $(m_i, \mathrm{severity}_i)$}
    \State $\zeta_t(m_i) \gets \mathrm{severity}_i$
    \State $w_{t+1}(m_i) \gets w_t(m_i)\,\exp(-\beta\, x_t\, \zeta_t(m_i))$ \Comment{\cref{eq:pcm_correct}}
  \EndFor
  \State Record the new observation as a fresh belief node.
\EndIf
\Statex \textit{// Phase D: Hypothesis lifecycle + write-back (deterministic)}
\For{each $h \in B_{t+1}$ with $\tau(h) = \textsc{Hypothesis}$}
  \If{$h$ not protected and not supported this turn}
    \State $w_t(h) \gets w_t(h)\,\exp(-\gamma\, a_t(h))$,\ $a_t(h) = t - \ell(h)$ \Comment{\cref{eq:pcm_hypothesis_decay}}
  \Else
    \State $\ell(h) \gets t$ if supported this turn \Comment{reset anchor, skip decay}
  \EndIf
  \If{$w_t(h) \ge \tau_{\mathrm{promote}}$} \State $\tau(h) \gets \textsc{Fact}$ \EndIf
  \If{$w_t(h) < 0.1$} \State Prune $h$ \EndIf
\EndFor
\State $E_{t+1} \gets E_t \cup \{(u_t, y_t)\}$ \Comment{episodic write-back}
\end{algorithmic}
\end{algorithm}

\subsection{Operator and gate derivations}
\label{app:maintain_derivation}

\paragraph{\texttt{MAINTAIN} update.}
The \texttt{MAINTAIN} update in \cref{eq:pcm_maintain} is a single gradient-descent step on the convex consolidation loss
$\ell^{+}(w)=\tfrac12(1-w)^2$. For a scalar weight $w$,
\[
w_{t+1}=w_t-\eta\nabla \ell^{+}(w_t)=w_t+\eta(1-w_t),
\]
which matches \cref{eq:pcm_maintain}. If $w_t\in[0,1]$ and $\eta\in(0,1]$, then $w_{t+1}\in[0,1]$ and the residual shrinks as $1-w_{t+1}=(1-\eta)(1-w_t)$.

\paragraph{\texttt{CORRECT} update.}\label{app:correct_derivation}
For a retrieved belief item $m$, let $x_t=S_{\mathrm{eff}}(u_t)$ denote the on-policy loss and let $\zeta_t(m)\in[0,1]$ be the per-item severity returned by the \texttt{Correct-Agent} (\cref{sec:appendix_agents}); operationally $\zeta_t(m)$ plays the role of a context-aware contradiction weight, and reduces to a calibrated NLI score $\pi_t(m)$ when the agent is replaced by the fixed verifier in \cref{eq:pcm_pi}. Consider the KL-proximal mirror-descent update:
\begin{equation}
\label{eq:app_correct_md}
w_{t+1}=\arg\min_{w\ge 0}\;x_t\,\zeta_t(m)\,w+\tfrac{1}{\beta}\mathrm{KL}(w\Vert w_t),
\end{equation}
where for scalar weights,
\begin{equation}
\label{eq:app_scalar_kl}
\mathrm{KL}(w\Vert w_t)=w\log\frac{w}{w_t}-w+w_t.
\end{equation}
Taking the first-order optimality condition of \cref{eq:app_correct_md} gives
$x_t\,\zeta_t(m)+\tfrac{1}{\beta}\log\frac{w_{t+1}}{w_t}=0$, hence
\begin{equation}
\label{eq:app_correct_closed_form}
w_{t+1}=w_t\exp\!\left(-\beta x_t\,\zeta_t(m)\right),
\end{equation}
which matches \cref{eq:pcm_correct}.

\paragraph{Two-threshold gate.}\label{app:gate_derivation}
We interpret the gate in \cref{eq:pcm_gate} as selecting an action that minimizes an on-policy risk plus an action cost:
\begin{equation}
\label{eq:pcm_cost_sensitive}
a_t=\arg\min_{a\in\{\mathrm{M,P,C}\}}\widehat{J}_t(a),\qquad \widehat{J}_t(a)=\widehat{L}_t(a)+c(a).
\end{equation}
Using the proxy $\widehat{L}_t(a)=x_t-\Delta_a(x_t)$ with a linear improvement model $\Delta_a(x)\approx g_a x$ yields
\[
\widehat{J}_t(a)\approx (1-g_a)x_t+c(a).
\]
The action boundaries are given by pairwise equality of these lines: $\widehat{J}_t(\mathrm{P})=\widehat{J}_t(\mathrm{M})$ gives
\[
x_t=\frac{c(\mathrm{P})-c(\mathrm{M})}{g_{\mathrm{P}}-g_{\mathrm{M}}}=: \theta_{\mathrm{low}},
\]
and $\widehat{J}_t(\mathrm{C})=\widehat{J}_t(\mathrm{P})$ gives
\[
x_t=\frac{c(\mathrm{C})-c(\mathrm{P})}{g_{\mathrm{C}}-g_{\mathrm{P}}}=: \theta_{\mathrm{high}},
\]
which gives the two thresholds:
\begin{equation}
\label{eq:pcm_thresholds}
\theta_{\mathrm{low}}=\frac{c(\mathrm{P})-c(\mathrm{M})}{g_{\mathrm{P}}-g_{\mathrm{M}}},
\qquad
\theta_{\mathrm{high}}=\frac{c(\mathrm{C})-c(\mathrm{P})}{g_{\mathrm{C}}-g_{\mathrm{P}}}.
\end{equation}
We treat $(c(a),g_a)$ as interpretable design abstractions rather than quantities to be estimated end-to-end: they motivate a \emph{two-threshold} rule and explain why the boundaries should be ordered, but the operating point $(\theta_{\mathrm{low}},\theta_{\mathrm{high}})=(0.2,0.6)$ is calibrated empirically via the sensitivity sweep in \cref{app:theta_sensitivity}, and a personalized data-driven instantiation is given in \cref{app:dynamic_threshold}.

\paragraph{\texttt{PROFILE} initialization.}\label{app:profile_init}
\Cref{eq:pcm_profile_init} fuses a gate-level prior with the agent's scalar. The class label $c$ returned by the \texttt{Profile-Agent} selects the band $I_c=[w_{\min}^{(c)},w_{\max}^{(c)}]$; the deterministic layer computes the prior $b_t^{(c)}=w_{\min}^{(c)}+\sigma(x_t)\bigl(w_{\max}^{(c)}-w_{\min}^{(c)}\bigr)$, clips the agent's \textsc{initial\_confidence} $q_t$ to $I_c$, and averages the two. For example, with $I_c=[0.60,0.70]$, $x_t=0.4$ and $q_t=0.65$, one obtains $b_t^{(c)}=0.6599$ and $w_0(h)=0.6549$. Because $q_t$ is clipped before it is halved, two agent scalars that differ by the full band width move $w_0$ by at most $0.05$, so the number of \texttt{MAINTAIN} updates needed for promotion under \cref{eq:pcm_maintain} changes by at most one, and $w_{\max}^{(c)}<\tau_{\mathrm{promote}}$ guarantees that no single agent output can commit a hypothesis as a \texttt{FACT}. The narrow range of $\sigma(x_t)$ inside the band is therefore intentional: the class band carries the semantics, and the within-band value only fine-tunes the starting point. The agent additionally returns the \textsc{decay\_protection} flag that governs \cref{eq:pcm_hypothesis_decay}.

\paragraph{Hypothesis decay horizon.}\label{app:decay_horizon}
\Cref{eq:pcm_hypothesis_decay} is applied once per turn to every unprotected hypothesis that received no supporting evidence in that turn, with the exponent $a_t(h)$ measured from the last evidence-bearing update $\ell(h)$. For a hypothesis that is never reinforced the factors compound, so after $n$ consecutive unsupported turns
\begin{equation}
\label{eq:app_decay_cumulative}
w_{\ell+n}(h)=w_{\ell}(h)\exp\!\Bigl(-\gamma\,\frac{n(n+1)}{2}\Bigr),
\end{equation}
which is faster than a fixed per-turn factor. This is intended: it instantiates the capacity penalty $\Omega$ of \cref{eq:app_step_surrogate} (hypotheses are pruned below $w=0.1$) and is why effective memory plateaus over the dialogue (\cref{sec:memory_scalability}). With the decay rate $\gamma=0.05$ used in our experiments, an unsupported latent hypothesis started inside its band is pruned after eight consecutive unsupported turns. The horizon counts only consecutive unsupported turns: a single reinforcement resets $\ell(h)$ and restarts the window, and explicit-event hypotheses are decay-protected, so only guesses that are never revalidated are culled.

\subsection{Per-turn pipeline (hybrid execution)}
\label{app:algorithm}

Algorithm~\ref{alg:precomem} makes the hybrid execution discussed in \cref{sec:pcm_evolve} explicit. Each turn invokes (i) a deterministic \texttt{Retrieve} step, (ii) a deterministic computation of $\rho_t,S_{\mathrm{raw}},x_t$, (iii) a regime-specific LLM sub-agent (one of \texttt{Maintain-Agent}, \texttt{Profile-Agent}, \texttt{Correct-Agent}; prompts in \cref{sec:appendix_agents}) that emits per-item scalar scores, and (iv) a deterministic application of the closed-form updates in \cref{eq:pcm_maintain,eq:pcm_profile_init,eq:pcm_correct}. The sub-agents only inject \emph{which} items to touch and \emph{how strongly}; gating, soft-decay, reinforcement, and hypothesis-promotion mechanics are entirely symbolic.

\section{Additional experimental details}
\label{app:experimental_details}

\subsection{Third-party benchmarks: AMemGym and LoCoMo-Plus}
\label{app:external_benchmarks}

To test whether the write-path advantage transfers to data we did not design, we evaluate PreCoMem on two public benchmarks built by other groups: AMemGym~\citep{cheng2026amemgym}, an on-policy benchmark with gold state transitions in which a simulated user's latent task state (for example a project's preparation stage or budget band) changes across periods, and LoCoMo\mbox{-}Plus~\citep{li2026locomo}, a static benchmark of 401 implicit cognitive-constraint questions. Both are run under their official protocols with the same backbone for all compared systems.

\paragraph{AMemGym.}\label{app:amemgym}
We use the official on-policy \texttt{v1.base} release (code snapshot \texttt{ffcd188}, data \texttt{4b8f64f}) at the official 20-profile / 11-period setting (ten gold state transitions per profile; 943 sessions and 2,200 questions per run) with three paired seeds $(13, 42, 1234)$. The user simulator is \textit{gpt-4.1} ($T{=}0.2$); PreCoMem and all of its variants use \textit{gpt-4.1-mini} ($T{=}0$) with \texttt{text-embedding-3-small}, Top-$K{=}10$, $(\theta_{\mathrm{low}},\theta_{\mathrm{high}})=(0.2,0.6)$ and $\lambda=0.5$. Each AMemGym session is a single user turn, so one gate event is logged per session. We report the benchmark's two headline scores, the Overall accuracy over the 2,200 questions and the normalized Memory score defined by AMemGym (Overall rescaled between the benchmark's random-answer and upper-bound references), together with its native write / read / utilization failure rates. \Cref{tab:amemgym_main} compares PreCoMem with the strongest systems reported in the AMemGym paper under the same \textit{gpt-4.1-mini} backbone: PreCoMem leads the strongest of them (AWE) by $+0.028$ Overall and $+0.033$ Memory, so the advantage reproduces on data we did not design.

\begin{table}[t]
\centering
\small
\setlength{\tabcolsep}{8pt}
\renewcommand{\arraystretch}{1.05}
\caption{\textbf{AMemGym} (official \texttt{v1.base}, 20 profiles $\times$ 11 periods, \textit{gpt-4.1-mini} for all memory systems). Overall accuracy and normalized Memory score; PreCoMem is mean $\pm$ std over three seeds. Rows marked $\dagger$ are the values reported in the AMemGym paper under the same backbone; AWE was additionally run locally only for the probe metrics quoted in the text.}
\label{tab:amemgym_main}
\begin{tabular}{@{}l c c@{}}
\toprule
\textbf{System} & \textbf{Overall} $\uparrow$ & \textbf{Memory} $\uparrow$ \\
\midrule
No memory (gpt-4.1-mini)$^{\dagger}$ & 0.368 & 0.203 \\
A-Mem$^{\dagger}$ & 0.378 & 0.220 \\
Nemori$^{\dagger}$ & 0.385 & 0.231 \\
Mem0-Graph$^{\dagger}$ & 0.424 & 0.284 \\
AWE$^{\dagger}$ & 0.430 & 0.296 \\
\rowcolor{pcmlightblue}
\textbf{PreCoMem} & \textbf{0.458}\,$\pm$\,0.006 & \textbf{0.329}\,$\pm$\,0.009 \\
\bottomrule
\end{tabular}
\end{table}

To see which operator carries the gain, we remove each operator in turn on the same runs (\cref{tab:amemgym_ablation}). Besides the native scores we track, per seed, 203 probes that query an attribute in the first period after its transition (changed-state probes) and 397 probes on unchanged attributes (stable-state probes). \emph{Changed-state} and \emph{stable-state accuracy} are the accuracies on these two probe sets; \emph{first-write} is the share of changed-state probes for which the belief that dominates retrieval already encodes the new value; \emph{stale-state error} is the share of changed-state probes answered with the superseded value; \emph{recovery} is the mean number of periods until a changed-state probe that was not immediately correct is answered correctly; and \emph{old-state suppression} is the share of changed-state probes for which the superseded belief is absent from the answering Top-10 or ranked below the new one. The three operators leave three distinct signatures, matching \cref{fig:ablation_row}(c): removing \texttt{PROFILE} mainly costs first-write success and changed-state accuracy; removing \texttt{CORRECT} raises the stale-state error, stretches recovery from one period to three, and degrades old-state suppression the most; removing \texttt{MAINTAIN} degrades stable-state accuracy together with the highest read and utilization failure. Paired 95\% confidence intervals of Full minus each variant (three seeds, $t$-distribution) exclude zero on every metric; for the Memory score they are $0.020\,[0.011,0.029]$, $0.026\,[0.012,0.040]$ and $0.056\,[0.040,0.072]$ for the three variants. Under the same probes, our local run of AWE gives a stale-state error of $15.6\%$ and a recovery of $2.0$ periods, against $10.8\%$ and $1.0$ for PreCoMem. In w/o \texttt{PROFILE}, medium-surprisal observations bypass the waiting room and are committed directly as beliefs; this is the same variant as in \cref{fig:ablation_row}(c) and \cref{tab:locomo_plus}.

Operator ablations of this kind need a gold transition timeline that splits accuracy into stable, change-point and recovery phases. Static question-answering benchmarks such as LoCoMo and LongMemEval\mbox{-}S yield a single end number per run and cannot attribute an effect to one operator, which is why we report operator ablations on AMemGym and DiMoBench and complement them with the route-matched controls of \cref{app:route_controls}.

\begin{table}[t]
\centering
\footnotesize
\setlength{\tabcolsep}{5pt}
\renewcommand{\arraystretch}{1.08}
\caption{\textbf{Operator ablation on AMemGym} (mean over three seeds). Top: the benchmark's native scores and failure rates (the three failure rates sum to $1-$Overall). Bottom: the state-query probes defined in the text (203 changed-state and 397 stable-state probes per seed). Paired 95\% CIs of Full minus each variant exclude zero on every metric.}
\label{tab:amemgym_ablation}
\begin{tabular}{@{}l ccccc@{}}
\toprule
\textbf{Variant} & Overall$\uparrow$ & Memory$\uparrow$ & Write fail.$\downarrow$ & Read fail.$\downarrow$ & Util.\ fail.$\downarrow$ \\
\midrule
\rowcolor{pcmlightblue}
\textbf{Full PreCoMem} & \textbf{0.458} & \textbf{0.329} & \textbf{.300} & .170 & .072 \\
w/o \texttt{PROFILE}  & 0.444 & 0.309 & .320 & \textbf{.165} & \textbf{.071} \\
w/o \texttt{CORRECT}  & 0.439 & 0.303 & .312 & .178 & \textbf{.071} \\
w/o \texttt{MAINTAIN} & 0.418 & 0.273 & .310 & .190 & .082 \\
\bottomrule
\end{tabular}

\vspace{5pt}

\setlength{\tabcolsep}{3.5pt}
\begin{tabular}{@{}l cccccc@{}}
\toprule
\textbf{Variant} & Changed-state$\uparrow$ & First-write$\uparrow$ & Stale-state$\downarrow$ & Recovery (periods)$\downarrow$ & Stable-state$\uparrow$ & Old-state suppr.$\uparrow$ \\
\midrule
\rowcolor{pcmlightblue}
\textbf{Full PreCoMem} & \textbf{0.704} & \textbf{76.4\%} & \textbf{10.8\%} & \textbf{1.0} & \textbf{0.762} & \textbf{83.1\%} \\
w/o \texttt{PROFILE}  & 0.654 & 67.8\% & 14.4\% & 1.8 & 0.746 & 74.5\% \\
w/o \texttt{CORRECT}  & 0.631 & 70.1\% & 19.2\% & 3.0 & 0.745 & 64.2\% \\
w/o \texttt{MAINTAIN} & 0.586 & 65.5\% & 21.5\% & 3.2 & 0.704 & 76.5\% \\
\bottomrule
\end{tabular}
\end{table}

\paragraph{LoCoMo-Plus.}\label{app:locomo_plus}
LoCoMo\mbox{-}Plus contains 401 questions that probe implicit cognitive constraints of four types (causal 101, state 100, goal 100, value 100) and scores each answer by constraint-consistency accuracy (\%), averaged over repeated judgments so that scores are not integer counts. All systems answer with GPT-4o. PreCoMem reaches $22.4\pm0.8$ overall, above A-Mem ($17.2$), Mem0 ($15.8$), SeCom~\citep{pan2025secom} ($14.9$) and even full-context GPT-4o ($21.05$), as reported in the LoCoMo\mbox{-}Plus paper. \Cref{tab:locomo_plus} breaks the score down by type against the w/o \texttt{PROFILE} control: the gains are largest on the non-preference types goal ($+5.0$) and state ($+4.8$), smaller on value ($+3.0$), and within seed noise on causal ($+1.1\pm0.5$), whose questions depend more on backbone reasoning than on the write path. Together with the task-state targets tracked by AMemGym, this shows the mechanism operating beyond preference modeling: nothing in the effective surprisal, the gate or the closed-form updates is preference-specific, since a belief is any compact claim carrying a confidence weight.

\begin{table}[t]
\centering
\small
\setlength{\tabcolsep}{7pt}
\renewcommand{\arraystretch}{1.05}
\caption{\textbf{LoCoMo\mbox{-}Plus by constraint type} (constraint-consistency accuracy, \%, GPT-4o for all systems; mean $\pm$ std over three seeds; Overall is the question-weighted aggregate over the 401 questions). Baseline overall scores are those reported in the LoCoMo\mbox{-}Plus paper.}
\label{tab:locomo_plus}
\begin{tabular}{@{}l c c c c@{}}
\toprule
\textbf{Type} & $n$ & \textbf{PreCoMem} & \textbf{w/o \texttt{PROFILE}} & $\Delta$ \\
\midrule
goal   & 100 & 22.3\,$\pm$\,1.1 & 17.3\,$\pm$\,2.0 & $+5.0$ \\
state  & 100 & 22.6\,$\pm$\,0.9 & 17.8\,$\pm$\,1.2 & $+4.8$ \\
value  & 100 & 21.6\,$\pm$\,0.8 & 18.6\,$\pm$\,0.9 & $+3.0$ \\
causal & 101 & 23.0\,$\pm$\,0.6 & 21.9\,$\pm$\,1.0 & $+1.1$ \\
\midrule
\rowcolor{pcmlightblue}
Overall & 401 & \textbf{22.4}\,$\pm$\,0.8 & 18.9\,$\pm$\,1.2 & $+3.5$ \\
\midrule
\multicolumn{5}{@{}l}{Reported overall baselines: full-context GPT-4o 21.05; A-Mem 17.2; Mem0 15.8; SeCom 14.9} \\
\bottomrule
\end{tabular}
\end{table}

\subsection{Backbone-generalization protocol and token usage}
\label{app:backbone_protocol}

\paragraph{Scale-vs-mechanism protocol behind \cref{fig:locomo_eval_overview}(a)--(d).}
Each x-axis group of \cref{fig:locomo_eval_overview}(a)--(d) fixes the model that answers the questions: Qwen3.5-35B, Qwen3.5-122B or GPT-5-Chat. The two bars of a group differ only in how memory is built and maintained. The left bar runs the answer model with its own native memory, instantiated by O-Mem at the same scale, so every such bar already includes a full memory system. The right bar keeps the answer model unchanged and delegates memory construction (the write path and retrieval) to PreCoMem running on a one-tier-smaller backbone: Qwen3.5-9B for the Qwen3.5-35B group, Qwen3.5-35B for the Qwen3.5-122B group, and GPT-5-mini for the GPT-5-Chat group. The right bar exceeds the left one in every group and on every dataset (all twelve $\Delta>0$): memory construction can be delegated to a much cheaper model and still beat the larger model's native memory, i.e., with the answer model held equal the mechanism substitutes for roughly one tier of model scale.

\paragraph{Token usage across benchmarks.}\label{app:token_usage}
\Cref{tab:token_usage} reports the per-query token usage of the response model on all three benchmarks under the same accounting as the \textit{Avg.\ Tokens} column of \cref{tab:main_results}: the mean number of input tokens sent to the response model per query (retrieved memory, short-term window, question and instructions), excluding memory-construction calls. PreCoMem's prompt is bounded by construction (Top-$K{=}10$ beliefs, a six-turn window and at most eight episodic excerpts), so its budget grows only from 3.1k on LoCoMo to 3.8k on LongMemEval\mbox{-}S under \textit{gpt-4.1}, and by under 9\% when the PersonaMem\mbox{-}v2 context quadruples from 32k to 128k. On every benchmark no baseline is both cheaper and more accurate than PreCoMem, while MemOS and EverMemOS use 0.6--1.4k fewer tokens per query at lower accuracy.

\begin{table}[t]
\centering
\small
\setlength{\tabcolsep}{5pt}
\renewcommand{\arraystretch}{1.05}
\caption{\textbf{Response-model input tokens per query (thousands)} across benchmarks and backbones; PersonaMem\mbox{-}v2 under \textit{gpt-4.1}, averaged over MCQ and Open-Ended queries. Token usage is reported for all nine baselines, including those whose accuracy is not tabulated on a given benchmark in \cref{tab:main_results} or \cref{fig:locomo_eval_overview}. The cross-seed standard deviation is below $0.15$k for every entry.}
\label{tab:token_usage}
\begin{tabular}{@{}l cc cc cc@{}}
\toprule
& \multicolumn{2}{c}{\textbf{LoCoMo}} & \multicolumn{2}{c}{\textbf{LongMemEval\mbox{-}S}} & \multicolumn{2}{c}{\textbf{PersonaMem\mbox{-}v2}} \\
\cmidrule(lr){2-3}\cmidrule(lr){4-5}\cmidrule(lr){6-7}
\textbf{Method} & gpt-4o-mini & gpt-4.1 & gpt-4o-mini & gpt-4.1 & 32k & 128k \\
\midrule
MemoryOS  & 5.5 & 5.2 & 7.2 & 6.8 & 5.8 & 7.2 \\
Mem0      & 1.6 & 1.8 & 1.8 & 2.0 & 1.9 & 2.1 \\
A-Mem     & 2.7 & 2.9 & 3.3 & 3.6 & 3.2 & 3.8 \\
GAM       & 3.5 & 3.4 & 4.3 & 4.2 & 3.7 & 4.4 \\
MemBox    & 3.3 & 3.4 & 3.9 & 4.0 & 3.6 & 4.3 \\
MemOS     & 2.5 & 2.4 & 2.7 & 2.6 & 2.5 & 2.7 \\
O-Mem     & 2.9 & 2.7 & 3.7 & 3.4 & 3.0 & 3.6 \\
TiMem     & 2.8 & 2.7 & 3.6 & 3.5 & 2.9 & 3.6 \\
EverMemOS & 2.6 & 2.5 & 3.2 & 3.1 & 2.7 & 3.2 \\
\rowcolor{pcmlightblue}
\textbf{PreCoMem} & 3.3 & 3.1 & 4.1 & 3.8 & 3.5 & 3.8 \\
\bottomrule
\end{tabular}
\end{table}

\subsection{Extended results and judge validation}
\label{app:extended_results}

\paragraph{LongMemEval\mbox{-}S results under \textit{gpt-4o-mini}.}
\label{app:longmemeval_4omini}

For completeness, \cref{tab:longmemeval_4omini} reports the LongMemEval\mbox{-}S results under the \textit{gpt-4o-mini} backbone (omitted from the main Table~\ref{tab:main_results} to save space). The pattern mirrors \textit{gpt-4.1}: PreCoMem leads on overall accuracy ($84.28\!\pm\!0.05$, $+1.96$ over the strongest baseline EverMemOS) and on SSA, SSP, MS, and KU. These columns cover assistant-side single-session recall (SSA), preference-conditioned response generation (SSP), cross-session aggregation (MS), and knowledge updates (KU), while EverMemOS retains an SSU lead from its read-path optimization and TiMem leads on temporal reasoning (TR).

\begin{table}[t]
  \centering
  \small
  \renewcommand{\arraystretch}{1.06}
  \caption{\textbf{LongMemEval\mbox{-}S under \textit{gpt-4o-mini}.} Mean $\pm$ std over three seeds; \textbf{bold} best, \underline{underline} second-best. \textit{Avg.\ Tokens} is the average response-model input token usage per query, as in \cref{tab:main_results}. Columns are question types: SSU/SSA/SSP = single-session user/assistant/preference, MS = multi-session, KU = knowledge update, and TR = temporal reasoning.}
  \label{tab:longmemeval_4omini}
  \resizebox{\linewidth}{!}{%
  \setlength{\tabcolsep}{4.0pt}%
  \begin{tabular}{@{}l c c c c c c c c@{}}
    \toprule
    \textbf{Method} & \textbf{Avg.\ Tokens} & \textbf{SSU} & \textbf{SSA} & \textbf{SSP} & \textbf{MS} & \textbf{KU} & \textbf{TR} & \textbf{Overall} \\
    \midrule
    MemoryOS  & 7.2k & 58.21\,$\pm$\,0.62 & 59.60\,$\pm$\,0.92 & 46.03\,$\pm$\,0.33 & 53.68\,$\pm$\,0.35 & 54.71\,$\pm$\,0.23 & 51.45\,$\pm$\,0.06 & 54.85\,$\pm$\,0.80 \\
    MemOS     & \textbf{2.7k} & 70.15\,$\pm$\,0.57 & 68.95\,$\pm$\,0.67 & 56.02\,$\pm$\,0.96 & 62.53\,$\pm$\,0.17 & 66.54\,$\pm$\,0.26 & 64.70\,$\pm$\,0.41 & 64.66\,$\pm$\,0.78 \\
    TiMem     & 3.6k & 81.79\,$\pm$\,0.93 & 75.01\,$\pm$\,0.36 & 69.99\,$\pm$\,0.30 & 72.97\,$\pm$\,0.91 & 77.63\,$\pm$\,0.46 & \textbf{85.27\,$\pm$\,0.08} & 75.51\,$\pm$\,0.90 \\
    GAM       & 4.3k & 77.61\,$\pm$\,0.94 & 79.52\,$\pm$\,0.72 & 68.18\,$\pm$\,0.31 & 76.16\,$\pm$\,0.43 & 74.09\,$\pm$\,0.65 & 76.50\,$\pm$\,0.48 & 76.19\,$\pm$\,0.83 \\
    MemBox    & 3.9k & 79.40\,$\pm$\,0.67 & \underline{83.26\,$\pm$\,0.77} & \underline{71.19\,$\pm$\,0.46} & 76.16\,$\pm$\,0.34 & 75.44\,$\pm$\,0.53 & 78.55\,$\pm$\,0.11 & 77.37\,$\pm$\,0.39 \\
    O-Mem     & 3.7k & 88.97\,$\pm$\,0.99 & 82.28\,$\pm$\,0.42 & 70.55\,$\pm$\,0.45 & \underline{79.61\,$\pm$\,0.01} & \underline{84.68\,$\pm$\,0.12} & 78.08\,$\pm$\,0.69 & 81.78\,$\pm$\,0.36 \\
    EverMemOS & \underline{3.2k} & \textbf{94.10\,$\pm$\,0.24} & 79.82\,$\pm$\,0.20 & 68.44\,$\pm$\,0.49 & 76.98\,$\pm$\,0.59 & 82.87\,$\pm$\,0.42 & 75.47\,$\pm$\,0.86 & \underline{82.32\,$\pm$\,0.75} \\
    \rowcolor{pcmlightblue}
    \textbf{PreCoMem}  & 4.1k & \underline{91.58\,$\pm$\,0.04} & \textbf{86.38\,$\pm$\,0.19} & \textbf{79.40\,$\pm$\,0.81} & \textbf{84.84\,$\pm$\,0.22} & \textbf{87.13\,$\pm$\,0.56} & \underline{80.34\,$\pm$\,0.73} & \textbf{84.28\,$\pm$\,0.05} \\
    \bottomrule
  \end{tabular}}%
\end{table}

\paragraph{LLM-as-a-judge validation across benchmarks.}
\label{app:llm_judge_validation}

To validate the reliability of our automated evaluation, we conduct a per-category human--LLM agreement study on all three benchmarks. \Cref{tab:judge_validation} pools the results: across LoCoMo (199-QA stratified audit), LongMemEval\mbox{-}S (full 500 QAs), and DiMoBench (200-QA audit covering one canonical persona), the \textit{gpt-4.1-mini} primary judge agrees with expert human re-annotation at $96.9\%$ overall ($93.8$--$98.5\%$ per question type) with Cohen's $\kappa$ between $0.81$ and $0.96$ in every question type ($0.957$, $0.878$ and $0.821$ at the dataset level for LoCoMo, LongMemEval\mbox{-}S and DiMoBench). $\kappa$ is lowest on DiMoBench, which combines the lowest agreement ($95.4\%$) with a high chance-agreement baseline: both raters mark about 85\% of its answers correct, so more of the agreement is expected by chance than on LoCoMo, where about 72\% are marked correct. To rule out the concern that the judge might be calibrated only on PreCoMem outputs, we additionally re-ran the same human-vs-judge study on $50$ outputs each from the two strongest baselines (EverMemOS and TiMem) on LoCoMo and LongMemEval\mbox{-}S; agreement remained within a $\pm 0.7$-point band of the pooled PreCoMem agreement ($96.4$--$97.5\%$), confirming that the protocol is stable across systems. This supports the use of LLM-as-a-judge in our experiments.

\begin{table}[t]
  \caption{\textbf{Per-category LLM-judge validation on LoCoMo, LongMemEval\mbox{-}S, and DiMoBench.} For each question type we report the LLM judge's verdict on PreCoMem outputs (\textit{gpt-4o-mini} backbone), the expert human re-annotation accuracy on the same outputs, the agreement rate, and Cohen's $\kappa$, the chance-corrected agreement between the LLM judge and the human rater. For LongMemEval\mbox{-}S, SSU/SSA/SSP denote single-session user/assistant/preference, MS denotes multi-session, KU denotes knowledge update, and TR denotes temporal reasoning. Following the human-agreement check of \citet{wu2024longmemeval}, we find the \textit{gpt-4.1-mini} primary judge highly consistent with human raters across all three benchmarks; an additional cross-method spot-check on EverMemOS and TiMem outputs (described in the text) confirms the protocol is not calibrated to PreCoMem outputs alone, so the LLM-judged numbers in \cref{tab:main_results} faithfully reflect human assessments.}
  \label{tab:judge_validation}
  \centering
  \small
  \renewcommand{\arraystretch}{1.05}
  \resizebox{\linewidth}{!}{%
  \setlength{\tabcolsep}{4pt}%
  \begin{tabular}{llccccc}
    \toprule
    Dataset & Question Type & \#Q & LLM-judge (\%) & Human (\%) & Agreement (\%) & Cohen's $\kappa$ \\
    \midrule
    \multirow{5}{*}{LoCoMo}
      & Single-Hop  & 109 & 81.41 & 82.73 & 98.5 & 0.949 \\
      & Multi-Hop   & 30  & 68.49 & 70.40 & 98.0 & 0.953 \\
      & Temporal    & 40  & 66.21 & 67.65 & 98.2 & 0.959 \\
      & Open-Domain & 20  & 39.63 & 41.55 & 97.5 & 0.948 \\
      \cmidrule(lr){2-7}
      & Overall     & 199 & 71.96 & 73.30 & 98.3 & 0.957 \\
    \midrule
    \multirow{7}{*}{LongMemEval\mbox{-}S}
      & SSU & 70  & 91.58 & 92.86 & 97.4 & 0.819 \\
      & SSA & 56  & 86.38 & 87.50 & 97.8 & 0.903 \\
      & SSP & 30  & 79.40 & 82.50 & 95.6 & 0.858 \\
      & MS  & 133 & 84.84 & 86.41 & 96.9 & 0.874 \\
      & KU  & 78  & 87.13 & 88.46 & 97.2 & 0.869 \\
      & TR  & 133 & 80.34 & 82.65 & 95.8 & 0.861 \\
      \cmidrule(lr){2-7}
      & Overall & 500 & 84.28 & 85.78 & 96.9 & 0.878 \\
    \midrule
    \multirow{5}{*}{DiMoBench}
      & Current-State Inference   & 50  & 86.18 & 87.83 & 96.2 & 0.832 \\
      & History Recall            & 50  & 87.74 & 89.62 & 96.4 & 0.821 \\
      & Change-Point Detection    & 50  & 78.31 & 80.07 & 93.8 & 0.812 \\
      & Hypothesis vs.\ Fact      & 50  & 83.93 & 85.61 & 95.1 & 0.810 \\
      \cmidrule(lr){2-7}
      & Overall                   & 200 & 84.04 & 85.78 & 95.4 & 0.821 \\
    \bottomrule
  \end{tabular}}%
\end{table}

\begin{table*}[t]
  \centering
  \caption{Hypothesis lifecycle statistics on LoCoMo (averaged over 10 samples).}
  \label{tab:hypothesis_lifecycle}
  \begin{tabular}{lrrrrr}
    \toprule
    Config & Created & Promoted & Expired & Promotion rate & Promotion precision \\
    \midrule
    PreCoMem (full) & 123 & 43 & 81  & 0.34 & 0.78 \\
    w/o C1   & 149 & 42 & 108 & 0.28 & 0.62 \\
    w/o C2   & 138 & 40 & 93  & 0.30 & 0.70 \\
    w/o C3   & 0   & 0  & 0   & 0.00 & -- \\
    \bottomrule
  \end{tabular}
\end{table*}

\paragraph{Hypothesis lifecycle statistics.}
\label{app:hypothesis_lifecycle}

\begin{figure}[t]
  \centering
  \includegraphics[width=\linewidth]{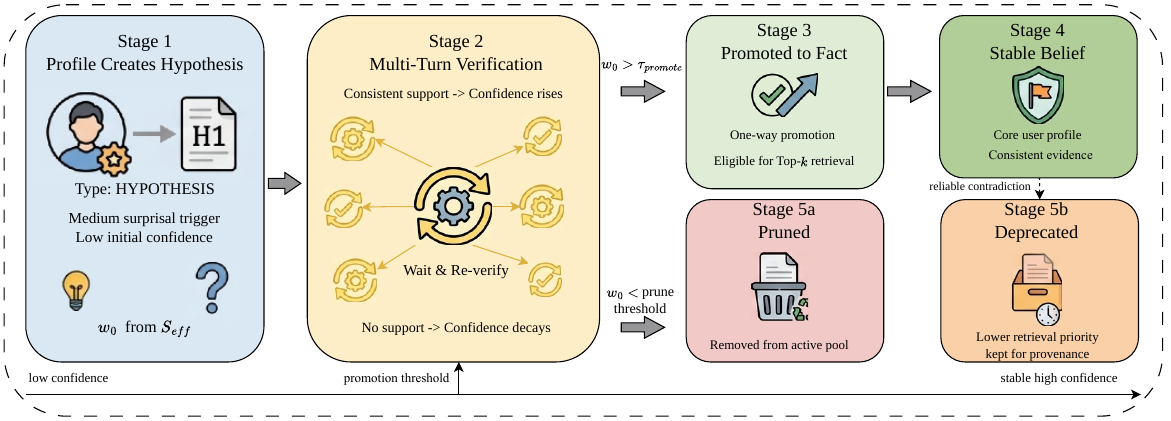}
  \caption{\textbf{Hypothesis lifecycle under PreCoMem.} A hypothesis created by \texttt{PROFILE} (Stage 1) enters multi-turn verification (Stage 2): consistent support raises its confidence toward the promotion threshold $\tau_{\mathrm{promote}}$ (Stage 3 $\to$ Stage 4, becoming a stable core-profile belief that ranks above hypotheses in confidence-weighted retrieval), while no/weak support lets its weight decay; once $w_t$ falls below the prune threshold the hypothesis is removed from the active pool (Stage 5a); a committed \texttt{FACT} that is later reliably contradicted is instead softly decayed and kept for provenance (Stage 5b). The horizontal axis is confidence, from low confidence through the promotion threshold to stable high confidence.}
  \label{fig:hypothesis_lifecycle}
\end{figure}

To make self-evolution \emph{observable}, we treat hypotheses as lifecycle objects that can be created, implicitly verified, promoted, or expired (illustrated in \cref{fig:hypothesis_lifecycle}). \Cref{tab:hypothesis_lifecycle} reports aggregate statistics on LoCoMo (averaged over 10 samples). Removing retrieval-reliability adjustment (w/o C1) increases hypothesis creation but reduces promotion precision, while removing the intermediate \texttt{PROFILE} regime (w/o C2) weakens gradual adaptation. Without the hypothesis lifecycle (w/o C3), these lifecycle metrics become undefined, since medium-surprisal information is written as facts directly.

\paragraph{Category-level annotation supporting the attribution.}\label{app:attribution}
The attribution of the gains in the main text rests on the benchmark annotations. On LoCoMo, Open-Domain is the only split whose questions are inferential in form (20.8\% begin with ``Would/Might/What would\ldots'') and whose gold answers are hedged inferences (14.6\%); every other split scores 0.0\% on both. Its target is typically a state not stated in a single turn: 49.0\% of its questions carry two or more evidence pointers and 32.3\% span two or more sessions (mean gap 4.51 sessions). Session span alone does not explain this: Multi-Hop spans more sessions (10.27 on average) yet has no hedged answers, since it joins co-valid stated facts, whereas Open-Domain must infer and revise a latent state. Temporal resolves from a single evidence pointer in 87.5\% of cases, which is why no gain appears there. On LongMemEval\mbox{-}S, all 78 KU questions have exactly two answer sessions, an old value and the value superseding it, so an un-updated memory is wrong by construction; SSA questions instead have one answer session whose gold evidence lies in assistant turns (51 of 56), a single target among $48.9$ sessions on average, so the SSA gain comes from confidence-weighted retrieval acting as an implicit reranker. We therefore attribute the drift-related gains to KU, SSP, cross-session MS and Open-Domain, and describe SSA as selective memory use.

\subsection{Component, operator, and control ablations}
\label{app:ablations}

\paragraph{Fine-grained ablation of raw-surprisal components.}
\label{app:raw_surprisal_ablation}

The raw-surprisal term $S_{\mathrm{raw}}=\lambda\, d_{\mathrm{sem}}+(1-\lambda)\, d_{\mathrm{conf}}$ mixes a semantic-gap signal and a verifier-based contradiction signal (\cref{eq:pcm_raw_surprisal}, \cref{app:surprisal_components}). To disentangle their individual contributions, we ablate each component by forcing one branch to zero. w/o $d_{\mathrm{sem}}$ sets $\lambda\!=\!0$, keeping only the contradiction verifier; w/o $d_{\mathrm{conf}}$ sets $\lambda\!=\!1$, keeping only the semantic gap. As shown in \cref{tab:ablation_fine}, both components are useful, but $d_{\mathrm{conf}}$ contributes substantially more: removing it costs $-10.08$ points on LoCoMo and $-11.57$ points on LongMemEval\mbox{-}S, roughly double the drop from removing $d_{\mathrm{sem}}$ ($-5.74$ and $-6.63$ points respectively). This matches the design intuition: the semantic gap alone cannot distinguish a genuinely contradictory utterance from one that is merely off-topic relative to current memory, whereas the NLI-based contradiction score directly detects reliable negation and drives \texttt{CORRECT}. Keeping both signals, on the other hand, gives the best trade-off between sensitivity (detecting drift cues via $d_{\mathrm{sem}}$) and precision (avoiding spurious corrections via $d_{\mathrm{conf}}$).

\begin{table}[t]
  \centering
  \small
  \setlength{\tabcolsep}{5pt}
  \renewcommand{\arraystretch}{1.05}
  \caption{\textbf{Fine-grained ablation of the raw-surprisal components.} Accuracy (\%) on LoCoMo and LongMemEval\mbox{-}S (\textit{gpt-4o-mini} backbone). w/o $d_{\mathrm{sem}}$ ($\lambda{=}0$) keeps only the NLI-based contradiction signal; w/o $d_{\mathrm{conf}}$ ($\lambda{=}1$) keeps only the semantic-gap signal. Both signals matter, but $d_{\mathrm{conf}}$ contributes nearly twice as much because it directly detects reliable negation.}
  \label{tab:ablation_fine}
  \begin{tabular}{lcc}
    \toprule
    Variant & LoCoMo & LongMemEval\mbox{-}S \\
    \midrule
    Full PreCoMem                            & \textbf{73.57} & \textbf{84.28} \\
    w/o $d_{\mathrm{sem}}$ ($\lambda{=}0$)   & 67.83 ($-5.74$) & 77.65 ($-6.63$) \\
    w/o $d_{\mathrm{conf}}$ ($\lambda{=}1$)  & 63.49 ($-10.08$) & 72.71 ($-11.57$) \\
    \bottomrule
  \end{tabular}
\end{table}

\paragraph{NLI verifier robustness (sanity check).}
\label{app:nli_robustness}
Because $d_{\mathrm{conf}}$ uses a $\max$ statistic over $\pi_t(m)$, a natural concern is whether the system is over-reliant on the specific NLI model behind $\mathrm{Verif}_{\mathrm{conf}}$. We probe this by replacing the default \texttt{cross-encoder/nli-deberta-v3-base} with two alternative MNLI-trained verifiers (each re-calibrated with a fresh temperature on the MNLI matched dev split, leaving every other PreCoMem component untouched) and one \emph{random}-$\pi_t$ control that draws $\pi_t(m)\!\sim\!\mathrm{Beta}(2,5)$ independently per item. To keep the sweep tractable we use LoCoMo Sample~1 ($400$ turns, $105$ QAs; the same sample as the threshold-sensitivity study in \cref{app:theta_sensitivity}) under the \textit{gpt-4o-mini} backbone, three seeds.

\begin{table}[t]
\centering
\small
\setlength{\tabcolsep}{5.5pt}
\renewcommand{\arraystretch}{1.05}
\caption{\textbf{NLI verifier sanity check on LoCoMo Sample~1.} Sample-1 LLM-judge accuracy (\%) when $\mathrm{Verif}_{\mathrm{conf}}$ is replaced. The two alternative MNLI-trained verifiers stay within $1$ point of the default; a randomized control collapses by $11$ points, confirming that $d_{\mathrm{conf}}$ is load-bearing yet that PreCoMem is not coupled to a specific verifier checkpoint.}
\label{tab:nli_robustness}
\begin{tabular}{lcc}
\toprule
$\mathrm{Verif}_{\mathrm{conf}}$ & Sample-1 Overall & $\Delta$ \\
\midrule
\texttt{cross-encoder/nli-deberta-v3-base} (default)  & $76.19\,\pm\,0.42$ & --       \\
\texttt{facebook/bart-large-mnli}                     & $75.74\,\pm\,0.38$ & $-0.45$  \\
\texttt{roberta-large-mnli}                           & $75.40\,\pm\,0.51$ & $-0.79$  \\
\midrule
random $\pi_t\!\sim\!\mathrm{Beta}(2,5)$ (control)    & $65.08\,\pm\,0.79$ & $-11.11$ \\
\bottomrule
\end{tabular}
\end{table}

\Cref{tab:nli_robustness} shows that both alternative NLI checkpoints stay within a $\pm 1$ point band of the default; a $\mathrm{Beta}(2,5)$-randomized $\pi_t$, in contrast, costs $11$ points, slightly more than removing $d_{\mathrm{conf}}$ altogether ($-11.11$ against $-10.08$ in \cref{tab:ablation_fine}). We read this as a two-sided sanity check: the $\max$-based $d_{\mathrm{conf}}$ is genuinely informative (random replacement is much worse than dropping the signal), but the gate is robust to the precise verifier as long as it is calibrated on a standard NLI distribution. Note that the absolute accuracies in this table are on a single $105$-QA sample and therefore not comparable to the full-LoCoMo numbers in \cref{tab:main_results}; the comparison is a within-row $\Delta$ across verifiers.

\paragraph{DiMoBench evolution-operator ablations.}
\label{app:dimobench_operator_ablation}

We ablate PreCoMem's three evolution operators on DiMoBench: \texttt{PROFILE} (hypothesis writing), \texttt{CORRECT} (surprisal-gated soft correction), and \texttt{MAINTAIN} (reinforcement/smoothing). We report accuracy every 10 turns from turn 150 to 350. As shown in \cref{fig:ablation_row}(c), removing \texttt{PROFILE} lowers the final accuracy ceiling (72\%), removing \texttt{CORRECT} substantially slows post-drift recovery, and removing \texttt{MAINTAIN} leads to high-variance dynamics that do not stabilize within our evaluation horizon (ending at 54\% accuracy).

\paragraph{Operator-level analysis.}
DiMoBench is designed to separate three regimes that stress different update behaviors: a stable pre-change phase (turns $\approx$150--200), a short change-point phase with mixed/ambiguous cues and decoy retrieval (turns $\approx$200--240), and a stable post-change phase where the latent attitude has reversed (turns $\approx$240--350). Under this structure, the three operators contribute in complementary ways.

\textbf{\texttt{PROFILE}: buffered adaptation under implicit drift.}\quad
Many change-point cues are neither clean confirmations nor decisive contradictions: they are novel, partially ambiguous, and often entangled with confounders. \texttt{PROFILE} converts these medium-$S_{\mathrm{eff}}$ signals into low-confidence, testable hypotheses, enabling the system to accumulate multiple weak cues before committing. This ``waiting room'' is especially important on DiMoBench because questions explicitly probe trend/change-point reasoning and hypothesis-vs-fact distinction. When \texttt{PROFILE} is removed, the system loses this intermediate representation and commits weak cues without a waiting room (premature commitment under decoys), which reduces the asymptotic ceiling (72\% vs.\ 84\% for full PreCoMem).

\textbf{\texttt{CORRECT}: targeted suppression of outdated beliefs.}\quad
After the drift, previously correct high-confidence beliefs become systematically misleading, and decoy retrieval makes stale evidence likely to reappear. \texttt{CORRECT} provides a selective revision mechanism by downweighting contradicted items via the $\zeta_t(m)$-modulated soft decay while leaving non-contradictory items largely unchanged. This targeted decay is crucial for reducing old-memory pull and preventing interference from pre-change beliefs. Without \texttt{CORRECT}, outdated facts remain competitive in retrieval, so the agent must rely on accumulating new nodes and indirect effects to ``outvote'' entrenched beliefs, yielding a substantially slower recovery (reaching 75\% only around turn 320 vs.\ turn 250) and a lower final plateau (77\% vs.\ 84\%).

\textbf{\texttt{MAINTAIN}: consolidation for stability.}\quad
\texttt{MAINTAIN} acts as a consolidation step that repeatedly reinforces supported beliefs, which sharpens confidence-weighted retrieval, increases retrieval reliability $\rho_t$ (by making the score mass more concentrated), and stabilizes the routing signal $x_t(=S_{\mathrm{eff}})$. It also underpins the hypothesis lifecycle: hypotheses created by \texttt{PROFILE} require repeated reinforcement to cross the promotion threshold and become stable facts. This stabilization is critical in the post-change phase, where the system should stop oscillating and settle into the post-change regime. When \texttt{MAINTAIN} is ablated, supported post-change beliefs cannot be progressively consolidated; retrieval remains diffuse and volatile, the reliability-adjusted loss becomes noisy, and updates keep ``chasing'' confounded evidence. Empirically, this yields persistent volatility: accuracy continues to fluctuate and ends at 54\% at the end of our evaluation.

\paragraph{Route-matched and factorial controls.}\label{app:route_controls}
Setting the gate score to zero (ES in \cref{fig:ablation_row}(a)) routes every turn to \texttt{MAINTAIN}, so it measures a reinforce-only lower bound; likewise CW sets $w\equiv1$ globally, so its drop is the total effect of confidence weighting across retrieval and the whole lifecycle (every hypothesis then trivially satisfies $w\ge0.80$ and is promoted at its first lifecycle step, pruning never fires, and soft decay is overwritten). \Cref{tab:route_controls} isolates the marginal value of the gate and the two roles of confidence with three controls on LoCoMo and LongMemEval\mbox{-}S (\textit{gpt-4o-mini}, three seeds). (i) A \emph{single-threshold} gate without the \texttt{PROFILE} band, which sends the medium band to \texttt{CORRECT}, reaches 69.92 / 81.16 ($-3.65$ / $-3.12$ against Full); this is the marginal value of the waiting room. (ii) \emph{Frequency-matched random routing} keeps the operator budget at Full's empirical routing frequencies (M/P/C $=56.2/27.4/16.4\%$) and leaves the read path unchanged, but assigns operators at random; it loses $11.74$ / $11.34$ with the update frequency held fixed, so the gain comes from the evidence-to-operator correspondence itself. (iii) A $2\times2$ factorial over read-time weighting ($w$ in \cref{eq:pcm_retrieval_score}) and lifecycle confidence ($w$ in the operator updates and in the promotion and pruning tests) shows that removing read-time weighting alone costs $1.77$ / $0.72$ points, removing lifecycle confidence alone costs $20.69$ / $22.63$, and removing both (the CW variant) costs $30.15$ / $34.63$; in half-coded effect terms the main effects are $+5.62$ / $+6.36$ for read-time weighting and $+24.54$ / $+28.27$ for lifecycle confidence, with an interaction of $-3.85$ / $-5.64$, i.e., each factor is worth more when the other is absent.

\begin{table}[t]
\centering
\small
\setlength{\tabcolsep}{7pt}
\renewcommand{\arraystretch}{1.05}
\caption{\textbf{Route-matched and factorial controls} (LLM-judge accuracy, \textit{gpt-4o-mini}, mean $\pm$ std over three seeds).}
\label{tab:route_controls}
\begin{tabular}{@{}l c c@{}}
\toprule
\textbf{Variant} & \textbf{LoCoMo} & \textbf{LongMemEval\mbox{-}S} \\
\midrule
\rowcolor{pcmlightblue}
Full PreCoMem & \textbf{73.57}\,$\pm$\,0.28 & \textbf{84.28}\,$\pm$\,0.05 \\
\midrule
Single-threshold gate (no \texttt{PROFILE} band; medium band $\to$ \texttt{CORRECT}) & 69.92\,$\pm$\,0.29 & 81.16\,$\pm$\,0.12 \\
Frequency-matched random routing (M/P/C $=56.2/27.4/16.4\%$) & 61.83\,$\pm$\,1.25 & 72.94\,$\pm$\,1.08 \\
\midrule
Read-time weighting off, lifecycle confidence on & 71.80\,$\pm$\,0.43 & 83.56\,$\pm$\,0.23 \\
Lifecycle confidence off, read-time weighting on & 52.88\,$\pm$\,0.69 & 61.65\,$\pm$\,0.68 \\
Both off ($=$ w/o CW) & 43.42\,$\pm$\,0.94 & 49.65\,$\pm$\,0.91 \\
\bottomrule
\end{tabular}
\end{table}

\paragraph{LLM-direct-write baseline.}\label{app:direct_write}
A natural alternative to the hybrid design of \cref{sec:pcm_framework} is to let the same LLM decide the whole write step. The direct-write baseline matches PreCoMem in every other component (retrieval, embedding, Top-$K{=}10$, context, judge, seeds, \textit{gpt-4.1}); the only change is that the LLM emits the entire write decision as one object, \{\texttt{route}, \texttt{target\_ids}, \texttt{operation}, \texttt{new\_weights}, \texttt{new\_belief}\}, whose weights are written verbatim, replacing the gate, the closed-form updates and the $[0,1]$ clamping. \Cref{tab:direct_write} shows that accuracy is lower in every seed ($-2.75$ on LoCoMo, $-3.15$ on LongMemEval\mbox{-}S) while per-query tokens are slightly higher (3.2k / 3.9k against 3.1k / 3.8k), so the loss is not offset by efficiency. The more telling difference is stability: the cross-seed standard deviation is $20\times$ higher on LoCoMo (1.42 against 0.07) and $2.1\times$ higher on LongMemEval\mbox{-}S, because unconstrained numeric decisions make the memory trajectory run-dependent. Replaying the write decisions on a frozen store, direct write emits a missing target id in 3.6\% of decisions, a self-contradictory operation in 2.7\% and invalid JSON in 1.8\% (mutually exclusive categories); such malformed writes corrupt the store silently and compound, whereas the closed-form updates stay in $[0,1]$ by construction. The loss is spread over all LoCoMo splits (Single-Hop $-1.65$, Multi-Hop $-4.80$, Temporal $-2.70$, Open-Domain $-6.50$). The deterministic layer is what keeps long-horizon memory bounded.

\begin{table}[t]
\centering
\small
\setlength{\tabcolsep}{8pt}
\renewcommand{\arraystretch}{1.05}
\caption{\textbf{LLM-direct-write baseline} (Overall LLM-judge accuracy, \textit{gpt-4.1}, mean $\pm$ std over three seeds; the paired difference is the mean over seeds of PreCoMem minus direct write).}
\label{tab:direct_write}
\begin{tabular}{@{}l c c@{}}
\toprule
& \textbf{LoCoMo} & \textbf{LongMemEval\mbox{-}S} \\
\midrule
\rowcolor{pcmlightblue}
PreCoMem (bounded scalars $+$ closed-form updates) & \textbf{85.06}\,$\pm$\,0.07 & \textbf{89.39}\,$\pm$\,0.76 \\
LLM direct write & 82.31\,$\pm$\,1.42 & 86.24\,$\pm$\,1.63 \\
\midrule
Paired difference & $+2.75$ & $+3.15$ \\
\bottomrule
\end{tabular}
\end{table}

\paragraph{Turn-level versus per-item routing.}\label{app:per_item_routing}
PreCoMem computes one gate score per turn and routes the whole retrieved set to a single operator; the per-item magnitudes $\eta_t(m)$ and $\zeta_t(m)$ apply after routing. A turn can therefore reinforce several beliefs, or correct several beliefs, at once, but it cannot reinforce one belief, correct another and profile a third in the same step. To measure how often this matters we annotated gold operators per retrieved item on 9,006 turns (all LoCoMo dialogues plus the LongMemEval\mbox{-}S evidence sessions): 3,475 turns touch at least one retrieved belief, and 95.83\% of them are expressible by a single route, either a single belief or several beliefs sharing one operator. The remaining 145 turns (1.61\% of all turns, 21.4\% of the 678 multi-item turns) are genuinely mixed. A per-item variant, identical in retrieval, backbone and closed-form updates but routing each item independently, raises the whole-turn agreement with the gold operators on these 145 turns by $+5.5$ points and their question accuracy by $+3.5$ points, at $+31\%$ write calls and $+14\%$ write tokens, for a net gain of $+0.3$ points over the full evaluation. Given the small net benefit at a materially higher cost, we keep turn-level routing as the default and regard per-item claim extraction as a direction for future work.

\subsection{Hyperparameter sensitivity and adaptive variants}
\label{app:sensitivity}

\paragraph{Sensitivity to retrieval budget $k$.}
\label{app:topk_sensitivity}

\begin{figure}[t]
  \centering
  \includegraphics[width=0.7\linewidth]{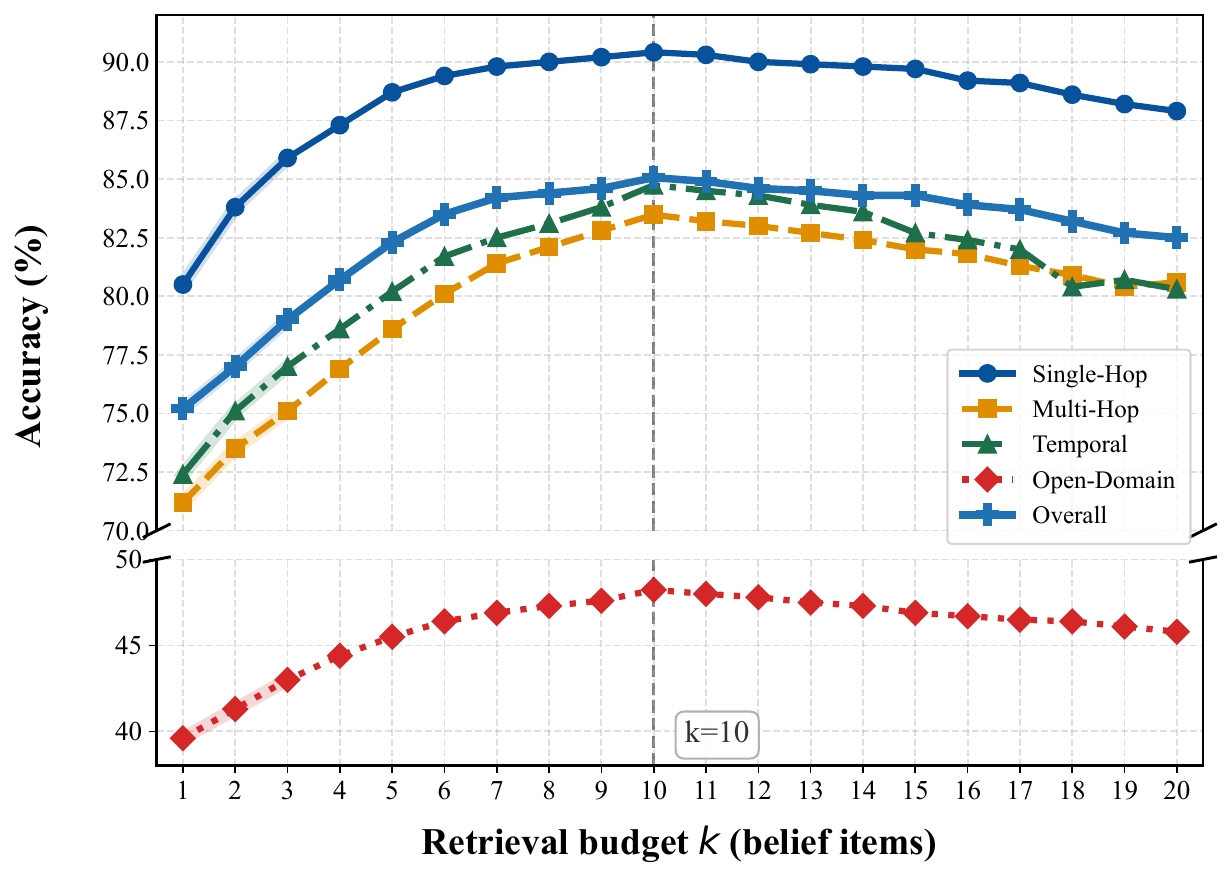}
  \caption{\textbf{Top-$K$ sensitivity on LoCoMo (\textit{gpt-4.1} backbone).} Per-split accuracy as $k$ varies in $\{1,\ldots,20\}$. The dashed vertical line marks our default $k{=}10$.}
  \label{fig:topk_sensitivity_curve}
\end{figure}

We vary the number of retrieved memory items $k$ on LoCoMo. \Cref{tab:topk_analysis} shows a clear ``fast gain, slow gain, mild drop'' trend: Overall accuracy improves substantially from $k{=}1$ to $k{=}5$ (75.2$\rightarrow$82.3), increases marginally from $k{=}5$ to $k{=}10$ (82.3$\rightarrow$85.06), and then declines at larger budgets (85.06$\rightarrow$82.5 at $k{=}20$). Category-wise, Temporal is the most sensitive to oversized budgets (84.7 at $k{=}10$ vs.\ 80.3 at $k{=}20$). \Cref{fig:topk_sensitivity_curve} visualizes the same sweep across all four LoCoMo splits with a broken-$y$ axis to keep the high- and low-accuracy bands legible side-by-side.

\begin{table*}[t]
  \centering
  \caption{Accuracy (\%) at different retrieval sizes $k$ on LoCoMo (\textit{gpt-4.1} backbone). Best results are in \textbf{bold}.}
  \label{tab:topk_analysis}
  \begin{tabular}{lccccc}
    \toprule
    $k$ & Single-Hop & Multi-Hop & Temporal & Open-Domain & Overall \\
    \midrule
    1  & 80.5 & 71.2 & 72.4 & 39.6 & 75.2 \\
    5  & 88.7 & 78.6 & 80.2 & 45.5 & 82.3 \\
    \textbf{10} & \textbf{90.41} & \textbf{83.48} & \textbf{84.74} & \textbf{48.23} & \textbf{85.06} \\
    15 & 89.7 & 82.0 & 82.7 & 46.9 & 84.3 \\
    20 & 87.9 & 80.6 & 80.3 & 45.8 & 82.5 \\
    \bottomrule
  \end{tabular}
\end{table*}

\paragraph{Mechanism-level analysis.}
We provide a more formal explanation of why oversized $k$ can hurt PreCoMem beyond generic retrieval noise. Recall that PreCoMem uses entropy-based retrieval reliability $\rho_t$ and the max-contradiction term $d_{\mathrm{conf}}$:

\begin{equation}
\label{eq:app_k_rho_def}
\begin{aligned}
\hat{s}_i 
&= \frac{s_i+\epsilon}{\sum_{j=1}^{k}(s_j+\epsilon)}, \\[2pt]
H_t 
&= \frac{-\sum_{i=1}^{k}\hat{s}_i \log \hat{s}_i}{\log k}, \\
\rho_t 
&= 1 - H_t .
\end{aligned}
\end{equation}

\begin{equation}
\label{eq:app_k_seff_def}
\begin{aligned}
S_{\mathrm{eff}}(u_t)=x_t
&= \rho_t \Big(
\lambda\, d_{\mathrm{sem}}(u_t,\mathcal{C}_t)
+ (1-\lambda)\, d_{\mathrm{conf}}(u_t,\mathcal{C}_t)
\Big), \\
d_{\mathrm{conf}}
&= \max_{m\in\mathcal{C}_t} \pi_t(m).
\end{aligned}
\end{equation}

\paragraph{(1) $k\uparrow$ flattens scores $\Rightarrow \rho_t\downarrow$ and \texttt{CORRECT} is suppressed.}
The reliability $\rho_t$ decreases when the normalized score distribution $\hat{\mathbf{s}}$ becomes less concentrated.
To make this explicit, consider a simple extension model where the Top-$k$ distribution over the original $k_0$ items is $\mathbf{p}\in\Delta^{k_0-1}$, and increasing $k$ adds $k_1$ additional tail items whose total normalized mass is $\delta\in(0,1)$ and is approximately uniform within the tail:
\begin{equation}
\label{eq:app_k_tail_mixture}
\mathbf{p}'=\big((1-\delta)p_1,\ldots,(1-\delta)p_{k_0}, \underbrace{\tfrac{\delta}{k_1},\ldots,\tfrac{\delta}{k_1}}_{k_1\ \text{items}}\big),  k=k_0+k_1.
\end{equation}
By concavity of Shannon entropy, injecting nonzero tail mass strictly increases entropy unless $\delta=0$. A direct decomposition yields
\begin{equation}
\label{eq:app_k_entropy_increase}
\mathcal{H}(\mathbf{p}')=(1-\delta)\mathcal{H}(\mathbf{p}) + h(\delta)+\delta\log k_1,
\end{equation}
where $\mathcal{H}(\cdot)$ is unnormalized entropy and $h(\delta)=-\delta\log\delta-(1-\delta)\log(1-\delta)$ is the binary entropy.
Thus, for typical retrieval where the tail receives non-negligible probability mass, $H_t$ increases with $k$ and $\rho_t$ decreases.

This interacts with the fixed-threshold gate: \texttt{CORRECT} is triggered when $x_t=S_{\mathrm{eff}}(u_t)>\theta_{\mathrm{high}}$.
Since $x_t=\rho_t S_{\mathrm{raw}}(u_t)$ with $S_{\mathrm{raw}}(u_t)\in[0,1]$, the event can be rewritten as
\begin{equation}
\label{eq:app_k_correct_event}
\rho_t S_{\mathrm{raw}}(u_t)>\theta_{\mathrm{high}}
\quad\Longleftrightarrow\quad
S_{\mathrm{raw}}(u_t)>\frac{\theta_{\mathrm{high}}}{\rho_t}.
\end{equation}
Therefore, as $\rho_t$ decreases, the required raw-surprisal level $\theta_{\mathrm{high}}/\rho_t$ increases, making reliable correction harder to trigger and strengthening old-memory pull. This effect is most visible for Temporal queries, where Top-$k$ often mixes pre- and post-change evidence, yielding a less concentrated score mass (higher $H_t$) even when the underlying contradiction is real.

\paragraph{(2) $k\uparrow$ amplifies spurious contradictions via a max statistic.}
The contradiction term uses an order statistic $d_{\mathrm{conf}}=\max_{m\in\mathcal{C}_t}\pi_t(m)$.
If we view $\pi_t(m)$ for \emph{irrelevant/mismatched} retrieved items as random variables with CDF $F_0$ on $[0,1]$, then for $k$ candidates,
\begin{equation}
\label{eq:app_k_max_tail}
\Pr\!\left(d_{\mathrm{conf}}>\tau\right)
=1-\Pr\!\left(\pi_t(m)\le \tau\right)^{k}
=1-F_0(\tau)^{k},
\end{equation}
which is monotonically increasing in $k$ for any $\tau\in(0,1)$ with $F_0(\tau)\in(0,1)$. When $1-F_0(\tau)$ is small, the increase is approximately linear:
\begin{equation}
\label{eq:app_k_fp_linear}
1-F_0(\tau)^k \approx k\,(1-F_0(\tau)).
\end{equation}
Thus, larger evidence sets increase the probability of observing an artificially high contradiction score from at least one mismatched memory, which can (i) inflate $S_{\mathrm{raw}}$ through the $(1-\lambda)d_{\mathrm{conf}}$ component and (ii) distract response generation by presenting a ``hard contradiction'' that is not actually relevant to the query. Together with (1), this explains why performance can decline for large $k$ despite higher coverage.

\paragraph{Implication.}
The empirical peak at $k{=}10$ reflects a regime where coverage is sufficient to retrieve the right evidence, while $\rho_t$ remains high enough to permit timely \texttt{CORRECT} actions and the max-contradiction term remains less exposed to spurious outliers. This motivates using $k{=}10$ as a robust default under a moderate compute budget.

\paragraph{Sensitivity to gating thresholds.}
\label{app:theta_sensitivity}

Theoretically, one may dynamically compute or adapt the thresholds (e.g., by estimating the action costs/effects in \cref{eq:pcm_thresholds}), which could yield better per-sample or per-stage performance. However, such on-policy adaptation introduces additional computation and confounds controlled evaluation because the observed behavior becomes entangled with an evolving threshold schedule.

\begin{figure}[t]
  \centering
  \includegraphics[width=0.65\linewidth]{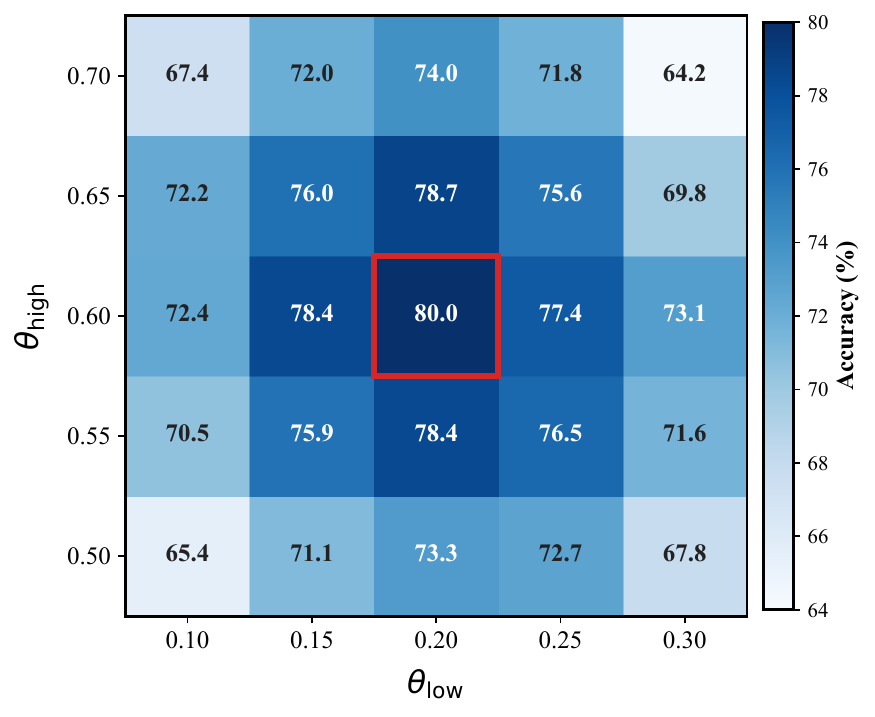}
  \caption{\textbf{Gating threshold sensitivity.} LoCoMo accuracy (Sample 1) over a $5{\times}5$ grid of $(\theta_{\mathrm{low}}, \theta_{\mathrm{high}})$ with step $0.05$ (25 configurations in total). The red box marks the deployed default $(0.20, 0.60)$, which sits at the center of a high-performing plateau; the four immediate neighbors of the peak stay within $1.3$--$2.6$ accuracy points, while accuracy degrades smoothly toward the grid corners.}
  \label{fig:theta_heatmap}
\end{figure}

We therefore adopt a \emph{fixed} threshold pair $(\theta_{\mathrm{low}},\theta_{\mathrm{high}})$ in all experiments and justify a reasonable choice via a sensitivity study. For controllable and cost-bounded ablation, we select a representative LoCoMo sample (Sample 1; 400 dialogue turns and 105 QA pairs) whose question types are relatively balanced and whose scale supports rapid iteration. We sweep $\theta_{\mathrm{low}}\in\{0.10,0.15,0.20,0.25,0.30\}$ and $\theta_{\mathrm{high}}\in\{0.50,0.55,0.60,0.65,0.70\}$ in steps of $0.05$ ($5{\times}5=25$ configurations) and evaluate LLM-judge accuracy on this sample. As shown in \cref{fig:theta_heatmap}, the deployed default $(0.20,0.60)$ (red box) sits at the center of a high-performing plateau: its four direct neighbors $(0.15,0.60)$, $(0.25,0.60)$, $(0.20,0.55)$ and $(0.20,0.65)$ all stay within a narrow $1.3$--$2.6$-point band of the peak. Accuracy decays smoothly toward the corners (down to roughly $64$--$67$\% near $\theta_{\mathrm{low}}{=}0.30,\theta_{\mathrm{high}}{=}0.70$ and the symmetric corner), confirming that the gate is locally robust rather than knife-edged.

Importantly, this fixed choice is a trade-off: we do not claim $(0.2,0.6)$ is universally optimal for every LoCoMo sample. Rather, it provides a stable and interpretable operating point with competitive accuracy under a fixed computational budget, and we keep it constant across experiments for fair comparison and reproducible observations.

\paragraph{Adaptive thresholding for personalization.}
\label{app:dynamic_threshold}

One natural question is whether the fixed gating thresholds $(\theta_{\mathrm{low}},\theta_{\mathrm{high}})=(0.2,0.6)$ are appropriate for all users, or whether different user populations would benefit from different thresholds. Intuitively, users differ in \emph{belief stability}: some keep stable preferences so that short-term deviations are noise, while others change frequently so that the same deviations reflect real drift. A fixed threshold cannot simultaneously satisfy both regimes. We therefore discuss and evaluate a simple adaptive-thresholding extension of PreCoMem.

\paragraph{User stability index.}
We summarize each user's recent dynamics with a scalar \emph{stability index} $\psi_t\in[0,1]$, computed from interaction statistics over a sliding window of length $W$:
\begin{equation}
\label{eq:user_stability}
\psi_t = \omega_1\,\psi_{\mathrm{corr}} + \omega_2\,\psi_{\mathrm{life}} + \omega_3\,\psi_{\mathrm{var}},
\end{equation}
where (i) $\psi_{\mathrm{corr}}=1-N_{\mathrm{correct}}^{(W)}/N_{\mathrm{total}}^{(W)}$ captures how rarely \texttt{CORRECT} fires in the recent window, (ii) $\psi_{\mathrm{life}}=\min\!\left(1,\bar{L}_{\mathrm{belief}}/L_{\mathrm{ref}}\right)$ captures average belief lifespan relative to a reference $L_{\mathrm{ref}}$, and (iii) $\psi_{\mathrm{var}}=\exp\!\left(-\kappa\cdot \mathrm{Var}\!\left[S_{\mathrm{eff}}^{(W)}\right]\right)$ captures the variance of effective surprisal. The coefficients satisfy $\omega_1+\omega_2+\omega_3=1$, and we use $\omega_1{=}0.5,\omega_2{=}0.3,\omega_3{=}0.2$ in the experiments below ($\kappa$ and $\omega_i$ are specific to this variant and unrelated to the $\gamma$ and $\alpha_j$ of the main text).

\paragraph{Threshold adaptation.}
We then map $\psi_t$ to thresholds by linear interpolation between a conservative upper configuration and an aggressive lower configuration:
\begin{equation}
\label{eq:theta_adaptive}
\begin{aligned}
\theta_{\mathrm{low}}(t)  &= \theta_{\mathrm{low}}^{\min} + \psi_t\!\left(\theta_{\mathrm{low}}^{\max}-\theta_{\mathrm{low}}^{\min}\right), \\
\theta_{\mathrm{high}}(t) &= \theta_{\mathrm{high}}^{\min} + \psi_t\!\left(\theta_{\mathrm{high}}^{\max}-\theta_{\mathrm{high}}^{\min}\right),
\end{aligned}
\end{equation}
with $\theta_{\mathrm{low}}\!\in\![0.10,0.35]$ and $\theta_{\mathrm{high}}\!\in\![0.40,0.80]$. Intuitively, a stable user ($\psi_t\!\to\!1$) receives larger thresholds, so the system requires stronger evidence before updating; a dynamic user ($\psi_t\!\to\!0$) receives smaller thresholds, so the system reacts faster. To avoid oscillation, we smooth the thresholds with an exponential moving average with coefficient $\mu{=}0.1$.

\paragraph{Theoretical view.}
From a Bayesian perspective, the thresholds encode a prior on belief persistence: stable users have a stronger prior that beliefs are unchanged, so a higher $\theta_{\mathrm{high}}$ corresponds to requiring stronger likelihood evidence to revise them, and vice versa for dynamic users. From a signal detection perspective, $\theta_{\mathrm{low}}$ and $\theta_{\mathrm{high}}$ trade off false-update rate (triggering \texttt{CORRECT} on noise) against missed-update rate (failing to trigger on genuine drift); since these two error rates have different relative costs for different users, the Bayes-optimal decision boundary is user-dependent. \Cref{eq:theta_adaptive} is a minimal realization of this idea that only reads summary statistics PreCoMem already maintains and introduces no extra trainable parameters.

\paragraph{Empirical results on LoCoMo.}
We evaluate the adaptive variant, denoted \textbf{PreCoMem (DT)}, on the full LoCoMo benchmark with two backbones (\textit{gpt-4o-mini} and \textit{gpt-4.1}), keeping every other setting identical to the main experiments. As shown in \cref{tab:dynamic_threshold_locomo}, relaxing the fixed thresholds to \cref{eq:theta_adaptive} yields consistent gains in overall accuracy ($+0.98$ on \textit{gpt-4o-mini}, $+1.15$ on \textit{gpt-4.1}), with the largest improvement on the \emph{Open-Domain} subset under \textit{gpt-4o-mini} ($+3.92$), which is the category most exposed to long-horizon drift. The same trend holds on LongMemEval\mbox{-}S, where PreCoMem (DT) lifts the overall score from $84.28$ to $85.36$ under \textit{gpt-4o-mini}, with the largest gains on \textbf{KU} (knowledge update, $+1.62$) and \textbf{SSP} (single-session preference, $+1.95$); these are precisely the question types that require per-user calibration of how aggressively to update beliefs.

\begin{table}[t]
  \centering
  \small
  \setlength{\tabcolsep}{3pt}
  \renewcommand{\arraystretch}{1.05}
  \caption{\textbf{Adaptive vs.\ fixed thresholds on LoCoMo} (accuracy \%, \textit{gpt-4o-mini} and \textit{gpt-4.1}). PreCoMem (DT) personalizes $(\theta_{\mathrm{low}},\theta_{\mathrm{high}})$ via the user stability index in \cref{eq:user_stability,eq:theta_adaptive}; PreCoMem (fixed) uses $(0.2,0.6)$. DT improves the overall average on both backbones, with the largest gain on the drift-heavy \emph{Open} category under \textit{gpt-4o-mini}, without any retraining.}
  \label{tab:dynamic_threshold_locomo}
  \begin{tabular}{llccccc}
    \toprule
    Backbone & Method & Single & Multi & Temp. & Open & Avg. \\
    \midrule
    \multirow{2}{*}{gpt-4o-mini}
      & PreCoMem (fixed) & 81.41 & 68.49 & 66.21 & 39.63 & 73.57 \\
      & PreCoMem (DT)    & \textbf{82.27} & \textbf{69.31} & \textbf{67.04} & \textbf{43.55} & \textbf{74.55} \\
    \midrule
    \multirow{2}{*}{gpt-4.1}
      & PreCoMem (fixed) & 90.41 & 83.48 & 84.74 & \textbf{48.23} & 85.06 \\
      & PreCoMem (DT)    & \textbf{91.42} & \textbf{85.16} & \textbf{85.81} & 47.78 & \textbf{86.21} \\
    \bottomrule
  \end{tabular}
\end{table}

These results show that adaptive thresholding is an almost-free extension that further personalizes PreCoMem for users with different stability profiles. We keep the fixed-threshold setting as the default in the main paper for controllable comparison, but we consider adaptive thresholding a natural deployment-time configuration, especially in long-horizon and cross-user settings.

\subsection{DiMoBench: construction and extended-horizon stress test}
\label{app:dimobench_full}

\paragraph{Construction pipeline and characteristics.}
\label{app:dimobench}

DiMoBench (Dialogue Implicit Memory Drift Benchmark) is designed to isolate \emph{belief-drift adaptation} under sparse and implicit cues. The released suite consists of \emph{five} 350-turn two-speaker dialogues, each built around a distinct persona drawn from a different combination of Big-Five extremes and life stages; for clarity the exposition below focuses on one canonical persona, but every quantitative result on DiMoBench (the dynamics in \cref{fig:ablation_row}(b) and the operator ablation in \cref{fig:ablation_row}(c)) is averaged across the five personas to avoid a single-persona artifact; the human--judge agreement audit in \cref{tab:judge_validation} uses the canonical persona, whose per-type accuracies lie within $0.15$ points of the five-persona means. Within each persona the user's latent attitude toward a target topic remains stable in turns 1--200, gradually reverses in turns 201--210 via 3--4 implicit events, and is then reinforced with denser behavioral signals in turns 211--240 before entering a low-frequency maintenance stage in turns 241--350.

To stress long-horizon robustness, DiMoBench injects confounders: every 30 turns we insert irrelevant chit-chat or hypothetical narration that shares surface keywords with the target topic, and we introduce multiple similar entity names to induce attribution ambiguity. Each persona contributes 200 QA pairs distributed across the dialogue (denser sampling near the change point and sparser in the long maintenance tail; $1{,}000$ in total across the five personas), covering (i) current-state inference, (ii) history recall, (iii) change-point detection, and (iv) hypothesis vs.\ fact distinction. Importantly, questions avoid explicit sentiment keywords and instead require reasoning from behavioral cues (plans, actions, reactions), making drift detection and correction rely on implicit evidence chains rather than lexical matches.

\paragraph{Why DiMoBench is needed.}
Existing long-horizon dialogue benchmarks predominantly evaluate the \emph{read path}: whether a system can correctly retrieve and reason over a static conversational record (LoCoMo~\citep{maharana2024evaluating}) or whether it can infer implicit personalization from accumulated preferences (PersonaMem\mbox{-}v2~\citep{jiang2025personamemv2}). None of them explicitly stress-test the \emph{write path}, i.e., whether a memory system can correctly decide \emph{when to maintain, update, or correct} its stored beliefs under gradual, implicit drift. DiMoBench is designed to fill this gap along three axes that are absent from existing benchmarks simultaneously: (i) \emph{continuous non-stationarity}, where the latent attitude reverses gradually over hundreds of turns without any explicit announcement; (ii) \emph{implicit behavioral cues}, where the drift manifests only through activity choices, reactions to invitations, and hedging language rather than sentiment keywords; and (iii) \emph{adversarial retrieval interference}, where five tagged interference types (\texttt{entity\_confusable}, \texttt{keyword\_trap}, \texttt{temporal\_decoy}, \texttt{hypothetical\_vs\_fact}, \texttt{none}) systematically probe two stability--plasticity failure modes: \emph{old-memory pull} and \emph{premature correction}. This write-path orientation is what lets DiMoBench reveal the distinct contributions of \texttt{MAINTAIN}, \texttt{PROFILE}, and \texttt{CORRECT} in \cref{app:dimobench_operator_ablation}, which cannot be isolated on LoCoMo or PersonaMem\mbox{-}v2 alone.

\begin{figure}[t]
  \centering
  \includegraphics[width=\linewidth]{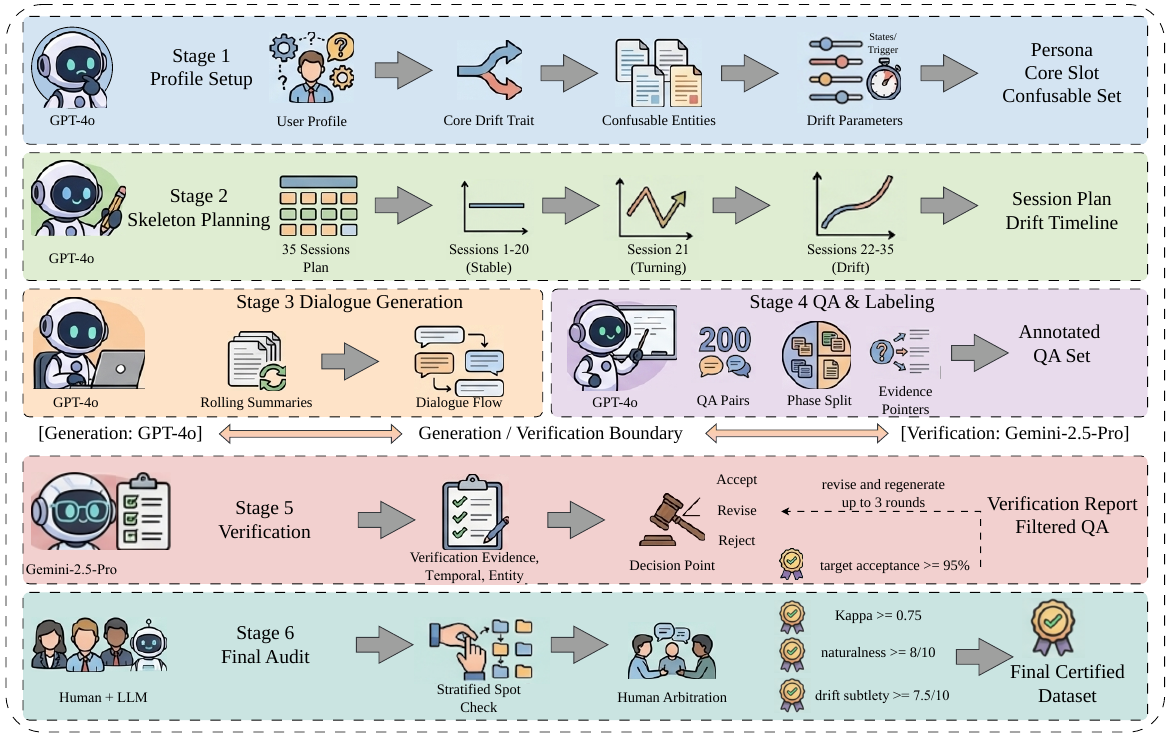}
  \caption{\textbf{DiMoBench six-stage construction pipeline.} A generator (GPT-4o) produces personas, life skeletons, dialogues, and QA pairs; an independent verifier (Gemini-2.5-Pro) cross-validates each QA pair along six criteria, and a 20\% stratified human audit with $\kappa\!\ge\!0.75$ provides final quality control.}
  \label{fig:dimobench_pipeline}
\end{figure}

\paragraph{Six-stage construction pipeline.}
DiMoBench is constructed by a six-stage progressive pipeline with an explicit \emph{generator--verifier separation} (GPT-4o for generation, Gemini-2.5-Pro for verification) to avoid self-reinforcing biases. \Cref{fig:dimobench_pipeline} visualizes the full pipeline as an end-to-end flow, and \cref{tab:dimobench_pipeline} summarizes the stages; details follow below.

\begin{table}[t]
\centering
\small
\setlength{\tabcolsep}{4pt}
\renewcommand{\arraystretch}{1.05}
\caption{\textbf{DiMoBench construction pipeline.} Generator and verifier are deliberately different models; a 20\% stratified human review provides final quality control.}
\label{tab:dimobench_pipeline}
\begin{tabular}{llll}
\toprule
Stage & Name & Model & Output \\
\midrule
1 & Persona construction    & GPT-4o          & User profile + core trait slot \\
2 & Life skeleton \& turn   & GPT-4o          & Event timeline + drift trigger \\
3 & Dialogue generation     & GPT-4o          & 35 sessions $\times$ 10 turns \\
4 & QA generation           & GPT-4o          & 200 QA pairs across 4 categories \\
5 & Consistency verification& Gemini-2.5-Pro  & Cross-validated QA (6 criteria) \\
6 & Human--machine review   & Human + LLM     & 20\% stratified audit, $\kappa\!\ge\!0.75$ \\
\bottomrule
\end{tabular}
\end{table}

\emph{Stage 1 (Persona).} Each persona is characterized by one \emph{core emotional/behavioral slot} (e.g., ``attitude toward social activities'') with a well-defined initial state and target state, a set of Big-Five personality dimensions, a small social circle of 2--3 named characters (some deliberately confusable to enable interference evaluation), routine daily activities, and a psychologically plausible drift trigger.

\emph{Stage 2 (Life skeleton).} The 350-turn trajectory is organized into 35 sessions of 10 turns each. Sessions 1--20 (turns 1--200) establish baseline behavior; session 21 (turns 201--210) introduces the drift trigger via 3--4 implicit events; sessions 22--24 (turns 211--240) reinforce the new behavior with denser cues; sessions 25--35 (turns 241--350) form a low-frequency maintenance stage with monotonically reduced divergence from a steady post-drift baseline.

\emph{Stage 3 (Dialogue generation).} Sessions are generated sequentially under cumulative context management. Generation rules enforce natural small talk and topic shifts, implicit expression through activity choices and hedging language, no meta-commentary (the user never says ``I've changed''), and frequent temporal markers (``last weekend'', ``this month'') for temporal reasoning.

\emph{Stage 4 (QA generation).} 200 QA pairs span all temporal phases (pre-change: 90; change-point: 10; post-change: 100). Each QA carries explicit evidence pointers (\texttt{D\{session\}:\{turn\}}), an interference tag, and a reasoning-type label.

\emph{Stage 5 (Consistency verification).} Each QA is cross-validated by Gemini-2.5-Pro along six dimensions: (i) \emph{answer grounding} (is the answer supported by cited evidence?), (ii) \emph{evidence completeness} (are all relevant turns cited?), (iii) \emph{temporal consistency} (does the answer reflect the correct phase?), (iv) \emph{entity accuracy}, (v) \emph{interference validity} (does the tag match the question?), and (vi) \emph{drift coherence} (is the behavioral change gradual and plausible?). QA pairs marked \texttt{REVISE} are regenerated by GPT-4o using verifier feedback (max 3 rounds); pairs marked \texttt{REJECT} are discarded. We target an $\ge\!95\%$ \texttt{ACCEPT} rate on the auto-verification pass.

\emph{Stage 6 (Human--machine review).} Three graduate-level annotators with NLP research experience underwent a 2-hour calibration session covering implicit emotional expression, temporal reasoning, and entity disambiguation. All annotators independently label the same 20 QA pairs, and independent annotation proceeds only after inter-annotator agreement reaches $\kappa\!\ge\!0.75$ (Cohen's $\kappa$). A 20\% stratified sample (stratified by temporal phase, interference type, and QA category) receives full human review; in addition, one annotator per persona performs a full 350-turn dialogue-coherence assessment. Quality targets are answer correctness $\ge\!95\%$, dialogue naturalness $\ge\!8.0/10$, and drift subtlety $\ge\!7.5/10$.

\paragraph{Representative QA examples.}
To make DiMoBench concrete, we include three canonical QA pairs illustrating each temporal phase and several interference tags:

\begin{tcolorbox}[colback=gray!5,colframe=gray!55,colbacktitle=gray!25,coltitle=black,boxrule=0.4pt,title={\small Pre-change QA (no interference, \texttt{MAINTAIN}-oriented)},fonttitle=\small]
\small
\textbf{Q:} What kind of activities does Mike like? \\
\textbf{A:} He likes group activities like KTV, parties, and escape rooms. \\
\textbf{Evidence:} \texttt{D1:3}\quad \textbf{Phase:} pre-change \quad \textbf{Interference:} none \quad \textbf{Category:} current-state inference
\end{tcolorbox}

\begin{tcolorbox}[colback=gray!5,colframe=gray!55,colbacktitle=gray!25,coltitle=black,boxrule=0.4pt,title={\small Change-point QA (implicit behavior, \texttt{PROFILE}-oriented)},fonttitle=\small]
\small
\textbf{Q:} Is Mike really that busy or is it an excuse? \\
\textbf{A:} Work is indeed busy, but previously he wouldn't decline because of this. \\
\textbf{Evidence:} \texttt{D21:5}, \texttt{D21:7}\quad \textbf{Phase:} change-point \quad \textbf{Interference:} none \quad \textbf{Category:} change-point detection
\end{tcolorbox}

\begin{tcolorbox}[colback=gray!5,colframe=gray!55,colbacktitle=gray!25,coltitle=black,boxrule=0.4pt,title={\small Post-change QA (temporal decoy, \texttt{CORRECT}-oriented)},fonttitle=\small]
\small
\textbf{Q:} Mike previously said he likes lively places; is that still true? \\
\textbf{A:} Not anymore; his preference has changed. \\
\textbf{Evidence:} \texttt{D7:5}, \texttt{D27:9}\quad \textbf{Phase:} post-change \quad \textbf{Interference:} \texttt{temporal\_decoy} \quad \textbf{Category:} current-state inference
\end{tcolorbox}

Additional traces that walk through the operator decisions on DiMoBench, including the retrieved Top-$K$ evidence and the resulting memory edits, are provided in \cref{app:end2end_trace}. All five 350-turn dialogues, the per-persona QA pairs, generation prompts, verification prompts, and human-review scripts are released with the paper at \url{https://github.com/ZhongWishing/PreCoMem}; the generation and verification prompts are also listed verbatim in \cref{app:dimobench_prompts}.

\paragraph{DiMoBench\mbox{-}XL construction.}
\label{app:dimobench_xl}
To stress-test the system beyond DiMoBench's 350-turn horizon and across more personas, we constructed \textbf{DiMoBench\mbox{-}XL}, a suite of \emph{five} 1050-turn dialogues, each with three sequential implicit drifts. Following the DiMoBench pipeline in \cref{app:dimobench}, we instantiate five user personas (each with a distinct demographic and dispositional profile, sampled to span both Big-Five extremes and different life stages); within each persona we keep five independent attributes (drink preference, exercise habit, home city, employment status, social style) and apply three drifts at turns $\{280,580,870\}$, each shifting one or two attributes via $20$--$30$ behavioral cues distributed across the surrounding session. Per persona we sample 20 evaluation QA pairs every 50 turns ($420$ pairs each, $2{,}100$ in total), balanced across four reasoning types and five interference tags. The whole suite is verified by Gemini-2.5-Pro with the same six-criterion protocol as DiMoBench, achieving a $96.4\%$ average \texttt{ACCEPT} rate across personas (range $94.8$--$97.6\%$) after one revision round.

\paragraph{Trajectory data.}
\Cref{tab:memory_scalability_xl_trajectory} reports the full $N_{\mathrm{raw}}$ / $N_{\mathrm{eff}}$ / per-bin counts every $100$ turns and at the three drift points on a representative DiMoBench\mbox{-}XL persona. $N_{\mathrm{eff}}$ enters its $25.8$--$28.5$ plateau by turn $200$, dips by at most $1.1$ points at a drift on this persona (at most $1.6$ across the five personas), and recovers within $\sim$50 turns to the same band. $N_{\mathrm{active}}$ (items with $w_t\!\ge\!0.8$) stabilizes at $17$--$18$ across all three drift epochs, indicating that each drift replaces the previous active set rather than accreting onto it. Across the five personas the per-bin $N_{\mathrm{eff}}$ trajectory stays within $\pm 1.4$ of the values reported here, and $N_{\mathrm{active}}$ never leaves the $16$--$19$ band after turn $200$, confirming the same qualitative behavior without persona-level outliers.

\begin{table}[t]
\centering
\small
\setlength{\tabcolsep}{4.4pt}
\renewcommand{\arraystretch}{1.05}
\caption{\textbf{DiMoBench\mbox{-}XL memory trajectory (representative persona).} Every $100$ turns (and at the three drift points) on one of the five DiMoBench\mbox{-}XL personas; per-persona deviations from this trajectory stay within $\pm 1.4$ on $N_{\mathrm{eff}}$ throughout the run. The four count columns partition $\mathcal{B}_t$ by confidence: $N_{\mathrm{active}}$ ($w_t\!\ge\!0.8$), $N_{\mathrm{stable}}$ ($0.5\!\le\!w_t\!<\!0.8$), $N_{\mathrm{weak}}$ ($0.1\!\le\!w_t\!<\!0.5$) and $N_{\mathrm{depr}}$ ($w_t\!<\!0.1$); the last column is the ratio $N_{\mathrm{eff}}/N_{\mathrm{raw}}$. Drift turns ($t\!\in\!\{280,580,870\}$) are marked with $\dagger$.}
\label{tab:memory_scalability_xl_trajectory}
\begin{tabular}{@{}r r r r r r r r@{}}
\toprule
$t$ & $N_{\mathrm{raw}}$ & $N_{\mathrm{eff}}$ & $N_{\mathrm{active}}$ & $N_{\mathrm{stable}}$ & $N_{\mathrm{weak}}$ & $N_{\mathrm{depr}}$ & $N_{\mathrm{eff}}/N_{\mathrm{raw}}$ \\
\midrule
100  & 26  & 21.0 & 13 & 8  & 3  & 2  & 0.81 \\
200  & 44  & 25.8 & 16 & 11 & 7  & 7  & 0.59 \\
$280^{\dagger}$ & 57 & 26.5 & 17 & 11 & 9  & 20 & 0.46 \\
300  & 60  & 26.7 & 17 & 12 & 9  & 22 & 0.45 \\
400  & 73  & 27.4 & 17 & 12 & 8  & 36 & 0.38 \\
500  & 81  & 28.5 & 18 & 13 & 8  & 42 & 0.35 \\
$580^{\dagger}$ & 89 & 27.4 & 17 & 11 & 10 & 51 & 0.31 \\
600  & 92  & 27.6 & 17 & 12 & 9  & 54 & 0.30 \\
700  & 103 & 28.0 & 18 & 12 & 9  & 64 & 0.27 \\
800  & 113 & 27.9 & 18 & 12 & 8  & 75 & 0.25 \\
$870^{\dagger}$ & 121 & 27.6 & 17 & 11 & 9  & 84 & 0.23 \\
900  & 123 & 28.2 & 18 & 12 & 8  & 85 & 0.23 \\
1000 & 128 & 28.0 & 18 & 12 & 8  & 90 & 0.22 \\
1050 & 131 & 28.2 & 18 & 12 & 8  & 93 & 0.22 \\
\bottomrule
\end{tabular}
\end{table}

\paragraph{QA accuracy stability.}
\Cref{tab:memory_scalability_xl_qa} aggregates QA accuracy across the five DiMoBench\mbox{-}XL personas in five time windows (means across the five personas). Across $1050$ turns and three drifts, Overall accuracy decays only $2.3$ points (from $84.3\%$ to $82.0\%$ on average) and the average Top-10 retrieval confidence stays at or above $0.78$ in every persona, confirming that long-horizon retention is gracefully degraded by confidence weighting rather than catastrophically collapsed; persona-level standard deviations stay below $1.0$ accuracy point in every window.

\begin{table}[t]
\centering
\small
\setlength{\tabcolsep}{6pt}
\renewcommand{\arraystretch}{1.05}
\caption{\textbf{QA accuracy and retrieval quality on DiMoBench\mbox{-}XL across time windows.} Numbers are averaged across the five DiMoBench\mbox{-}XL personas; per-window standard deviations across personas are $\le 1.0$ accuracy point and $\le 0.02$ for the Top-10 weight, and are omitted for readability. Average Top-10 retrieval confidence stays at or above $0.78$ throughout $1050$ turns; Overall accuracy decays by only $2.3$ points.}
\label{tab:memory_scalability_xl_qa}
\begin{tabular}{@{}l r r r r@{}}
\toprule
\textbf{Turn window} & \#QA & Acc.\ (\%) & Avg Top-10 $w$ & Contam.@10 (\%) \\
\midrule
$\phantom{0}50$--$250$    &  80 & 84.3 & 0.82 & 0.4 \\
$250$--$500$              & 100 & 83.6 & 0.80 & 0.6 \\
$500$--$750$              & 100 & 83.1 & 0.79 & 0.9 \\
$750$--$1000$             & 100 & 82.4 & 0.78 & 1.2 \\
$1000$--$1050$            &  40 & 82.0 & 0.78 & 1.3 \\
\bottomrule
\end{tabular}
\end{table}

\subsection{Memory dynamics: footprint, contamination, and retention}
\label{app:memory_dynamics}

\paragraph{Headline metrics and retrieval contamination.}
\label{app:retrieval_contamination}

\Cref{tab:memory_scalability} summarizes the headline scalability numbers used in \cref{sec:memory_scalability}. Below we further examine whether deprecated FACTs leak into Top-$k$ retrieval.

\begin{table}[t]
  \centering
  \small
  \setlength{\tabcolsep}{4pt}
  \renewcommand{\arraystretch}{1.10}
  \caption{\textbf{Memory scalability headline metrics.} $N_{\mathrm{raw}}$ grows linearly with the conversation, but $N_{\mathrm{eff}}$ converges to a stable plateau and the contamination of deprecated FACTs in Top-10 retrieval stays at or below $1.3\%$ in every window even after 1050 turns and three drifts. Both rows average over their five personas; per-persona deviations are $\le 0.5$ on $N_{\mathrm{eff}}^{\mathrm{end}}$ and $\le 0.4$ on Contam.@10.}
  \label{tab:memory_scalability}
  \begin{tabular}{@{}l c c c c@{}}
    \toprule
    \textbf{Setting} & $N_{\mathrm{raw}}^{\mathrm{end}}$ & $N_{\mathrm{eff}}^{\mathrm{end}}$ & Contam.@10 (\%) & QA $\Delta$ (\%) \\
    \midrule
    DiMoBench (350 t.)        & 62  & 27.6 & 0.6 & $-$0.6 \\
    DiMoBench\mbox{-}XL (1050 t.)    & 131 & 28.2 & 1.0 & $-$2.3 \\
    \bottomrule
  \end{tabular}
\end{table}

A second concern is whether deprecated FACTs ($w_t\!<\!0.1$) leak into Top-$k$ retrieval and silently bias the response. We post-analyze \emph{every} evaluation query on DiMoBench and DiMoBench\mbox{-}XL with three different retrieval budgets $k\!\in\!\{5,10,25\}$, recording (i) the score-mass ratio $\sum_{m\in\mathcal{C}_t,\,w_t<0.1} s(u_t,m)/\sum_{m\in\mathcal{C}_t} s(u_t,m)$, (ii) the fraction of Top-$k$ slots occupied by deprecated FACTs, and (iii) the average $w_t$ of retrieved items. Results in \cref{tab:retrieval_contamination} confirm that even at the most aggressive setting (DiMoBench\mbox{-}XL, $k\!=\!25$), deprecated FACTs occupy only $2.6\%$ of slots and contribute $0.94\%$ of total score mass; at our default $k\!=\!10$ the numbers drop to $1.0\%$ and $0.58\%$. The avg retrieved $w$ remains in $[0.71,\,0.84]$ across all settings, showing the score is dominated by high-confidence beliefs.

\begin{table}[t]
\centering
\small
\setlength{\tabcolsep}{6pt}
\renewcommand{\arraystretch}{1.05}
\caption{\textbf{Retrieval contamination by deprecated FACTs ($w_t\!<\!0.1$).} Even at $k\!=\!25$, deprecated items occupy at most $2.6\%$ of Top-$k$ slots and contribute under $1\%$ of the score mass.}
\label{tab:retrieval_contamination}
\begin{tabular}{@{}l r r r r@{}}
\toprule
\textbf{Dataset} & $k$ & Score-mass (\%) & Top-$k$ depr.\ (\%) & Avg retrieved $w$ \\
\midrule
DiMoBench       & 5  & 0.21 & 0.0 & 0.84 \\
DiMoBench       & 10 & 0.34 & 0.6 & 0.81 \\
DiMoBench       & 25 & 0.62 & 1.7 & 0.74 \\
\midrule
DiMoBench\mbox{-}XL    & 5  & 0.36 & 0.2 & 0.81 \\
DiMoBench\mbox{-}XL    & 10 & 0.58 & 1.0 & 0.78 \\
DiMoBench\mbox{-}XL    & 25 & 0.94 & 2.6 & 0.71 \\
\bottomrule
\end{tabular}
\end{table}

\paragraph{Per-persona end states.}\label{app:per_persona}
\Cref{tab:per_persona} lists, for each DiMoBench persona, the end-of-dialogue statistics behind the averages in \cref{tab:memory_scalability}: accuracy varies by at most 1.6 points across personas, and $N_{\mathrm{eff}}^{\mathrm{end}}$ and the Top-10 contamination stay within $0.3$ and $0.2$ points of their means. On DiMoBench\mbox{-}XL the per-persona $N_{\mathrm{eff}}$ trajectories stay within $\pm1.4$ of the representative persona of \cref{tab:memory_scalability_xl_trajectory}, $N_{\mathrm{eff}}^{\mathrm{end}}$ within $0.4$ and Contam.@10 within $0.25$ of the five-persona means.

\begin{table}[t]
\centering
\small
\setlength{\tabcolsep}{8pt}
\renewcommand{\arraystretch}{1.05}
\caption{\textbf{DiMoBench end-of-dialogue statistics per persona} (350 turns; mean over three seeds). The means reproduce the DiMoBench row of \cref{tab:memory_scalability}.}
\label{tab:per_persona}
\begin{tabular}{@{}l c c c c@{}}
\toprule
\textbf{Persona} & Acc.\ (\%) & $N_{\mathrm{raw}}^{\mathrm{end}}$ & $N_{\mathrm{eff}}^{\mathrm{end}}$ & Contam.@10 (\%) \\
\midrule
P1 (canonical) & 84.1 & 62 & 27.6 & 0.60 \\
P2 & 83.2 & 65 & 27.8 & 0.80 \\
P3 & 84.8 & 59 & 27.3 & 0.40 \\
P4 & 84.4 & 61 & 27.5 & 0.55 \\
P5 & 83.6 & 63 & 27.8 & 0.65 \\
\midrule
Mean & 84.0 & 62 & 27.6 & 0.60 \\
\bottomrule
\end{tabular}
\end{table}

\paragraph{FACT pruning ablation.}
\label{app:fact_pruning}

The default PreCoMem (config \textbf{A0}) prunes \texttt{HYPOTHESIS} items at $w_t\!<\!0.1$ but never deletes \texttt{FACT}s; this is intentional because deprecated \texttt{FACT}s serve as immutable provenance hooks for the corresponding episodic-cache traces. To validate that this provenance choice is not a hidden scalability liability, we test two pruning variants: \textbf{A1} also prunes FACTs at $w_t\!<\!0.05$, and \textbf{A2} prunes FACTs at the same $w_t\!<\!0.10$ used for hypotheses. \Cref{tab:fact_pruning_ablation} reports performance and final memory size on LoCoMo (\textit{gpt-4o-mini}, 10-sample mean) and on DiMoBench (350-turn end-state).

\begin{table*}[t]
\centering
\small
\setlength{\tabcolsep}{4pt}
\renewcommand{\arraystretch}{1.18}
\caption{\textbf{FACT pruning ablation.} A1 ($w\!<\!0.05$) shrinks raw memory by $41\%$ on LoCoMo and $34\%$ on DiMoBench while losing $\le\!0.2$ accuracy points, inside seed variance. A2 ($w\!<\!0.10$) starts pruning items that still have residual influence and degrades accuracy by $0.5$--$1.1$ points.}
\label{tab:fact_pruning_ablation}
\begin{tabular}{@{}l c c c c c c @{\hspace{10pt}} c c c@{}}
\toprule
& \multicolumn{6}{c}{\textbf{LoCoMo (gpt-4o-mini)}} & \multicolumn{3}{c}{\textbf{DiMoBench (350 t.\ end)}} \\
\cmidrule(lr){2-7}\cmidrule(lr){8-10}
\textbf{Config} & Single & Multi & Temp. & Open & Overall & $N_{\mathrm{raw}}^{\mathrm{end}}$ & Acc.\ (\%) & $N_{\mathrm{raw}}^{\mathrm{end}}$ & $N_{\mathrm{eff}}^{\mathrm{end}}$ \\
\midrule
\textbf{A0} (default; no FACT prune)        & 81.41 & 68.49 & 66.21 & 39.63 & \textbf{73.57} & 287 & 84.0 & 62 & 27.6 \\
\textbf{A1} (FACT prune $w\!<\!0.05$)       & 81.36 & 68.42 & 66.18 & 39.55 & 73.51 & 168 & 83.8 & 41 & 27.5 \\
\textbf{A2} (FACT prune $w\!<\!0.10$)       & 80.95 & 68.05 & 65.74 & 39.21 & 73.06 & 142 & 82.9 & 33 & 26.7 \\
\bottomrule
\end{tabular}
\end{table*}

A1 cuts $N_{\mathrm{raw}}$ from $287$ to $168$ on LoCoMo while losing only $0.06$ Overall points (well below the seed standard deviation of $0.28$), confirming that the additional FACTs A0 retains were already behaviorally inactive. A2 starts pruning items that still carry residual signal and consequently loses $0.5$ Overall on LoCoMo and $1.1$ points on DiMoBench. We therefore default to A0 for provenance and recommend A1 only for storage-bounded deployments.

\paragraph{Cross-system memory footprint.}
\label{app:cross_system_footprint}

A more detailed cross-system footprint comparison, including the per-item-confidence design dimension, is reported in the main paper as \cref{tab:memory_footprint} (\cref{sec:memory_scalability}). In short, on the ten LoCoMo dialogues ($\bar{N}\!=\!588$ turns), PreCoMem-A1 stores $168$ items, lower than both O-Mem ($232$) and EverMemOS ($264$), while delivering $+2.6$ and $+3.4$ Overall points respectively; O-Mem and EverMemOS commit each interaction through a discrete decision, so their stores admit no per-item confidence and the effective count $N_{\mathrm{eff}}$ is undefined.

\subsection{Retrieval dependence and reliability calibration}
\label{app:retrieval_dependence}

The write path acts only on the retrieved set $\mathcal{C}_t$: if the belief that should be revised is not retrieved, the contradiction cannot be detected and \texttt{CORRECT} cannot fire in that turn. This section quantifies that dependence and then examines what the reliability coefficient $\rho_t$ actually tracks.

\paragraph{Belief-update recall and failure decomposition.}\label{app:belief_update_recall}
We call an \emph{update-due event} a point at which a stored belief must be revised for a later question to be answered correctly (one per evaluation question on DiMoBench and DiMoBench\mbox{-}XL, one per state transition with a pre-existing belief on AMemGym, and one per knowledge-update turn on LongMemEval\mbox{-}S KU), and we measure the Recall@10 of the belief due for update, i.e., how often it enters the write-path Top-10 at that event. \Cref{tab:belief_update_recall} shows that recall covers most events (89--95\%), leaving an explicit 5--11\% blind spot, and that misses matter: question accuracy on events whose target was retrieved exceeds that on missed events by 16--22 points. Retrieval is favorable here because a contradictory utterance shares the attribute of the belief it contradicts with opposite polarity, so the target scores high on the semantic and lexical terms of \cref{eq:pcm_retrieval_score}. Decomposing all update-due events pooled over the four benchmarks, 89.3\% succeed; the residual 10.7\% splits into unrecovered retrieval misses (5.4\%; a further 1.4\% of events are missed at retrieval but corrected later), gate misses (3.1\%; the target was retrieved but the turn was not routed to \texttt{CORRECT}) and localization misses (2.2\%; the turn was routed to \texttt{CORRECT} but the target was not decayed). Retrieval is thus the largest single source, with half of the failures (5.3 of the 10.7 points) sitting in gating and localization; on LongMemEval\mbox{-}S KU, where each update is stated once, retrieval misses dominate. A miss does not stall adaptation: the observation still enters the episodic cache and can form a hypothesis, and 20\% of missed events (147 of 718) are corrected later by subsequent cues or hypothesis promotion.

To separate association from causation we force the target belief into the Top-10 at every missed event (displacing the lowest-ranked item) and re-run the write path and the question. This closes only 26--32\% of the hit--miss gap (last column of \cref{tab:belief_update_recall}): forcing the hit raises accuracy on the affected questions by 4.5--6.5 points, while the rest of the gap reflects that missed events are intrinsically harder and that gating and localization can still fail once the target is retrieved.

\begin{table}[t]
\centering
\small
\setlength{\tabcolsep}{6pt}
\renewcommand{\arraystretch}{1.05}
\caption{\textbf{Belief-update recall and the cost of a miss.} Recall@10 of the belief due for update (mean $\pm$ std over three seeds); question accuracy (\%) on events whose target was retrieved (hit) or not (miss), and on missed events after forcing the target into the Top-10 (forced hit); the last column is the share of the hit--miss gap recovered by the forced hit. AMemGym accuracies are changed-state probe accuracies.}
\label{tab:belief_update_recall}
\begin{tabular}{@{}l c c c c c@{}}
\toprule
\textbf{Benchmark} & \textbf{Recall@10} & \textbf{QA (hit)} & \textbf{QA (miss)} & \textbf{QA (forced hit)} & \textbf{Gap recovered} \\
\midrule
DiMoBench        & 95.0\,$\pm$\,0.4 & 84.8 & 64.2 & 70.7 & 32\% \\
DiMoBench\mbox{-}XL     & 93.0\,$\pm$\,0.2 & 84.7 & 63.0 & 69.4 & 29\% \\
LongMemEval\mbox{-}S KU & 90.0\,$\pm$\,0.5 & 93.4 & 75.8 & 80.3 & 26\% \\
AMemGym          & 89.1\,$\pm$\,0.5 & 72.1 & 55.8 & 60.6 & 29\% \\
\bottomrule
\end{tabular}
\end{table}

\paragraph{What $\rho_t$ measures: calibration.}\label{app:rho_calibration}
$\rho_t=1-H_t$ is the normalized concentration of the Top-$K$ retrieval scores. It does not judge whether an individual retrieved item is correct; it weights how far the surprisal computed from that recall can be trusted, and because it enters only multiplicatively it can make the gate more conservative but never more aggressive. Two cases illustrate the behavior. If several relevant memories are correctly retrieved with similar scores, $\rho_t$ is lower and the update is dampened and deferred; no information is lost, since supporting evidence keeps accumulating through the hypothesis lifecycle and a genuine change is still committed after slightly more confirmation. If a wrong memory receives a very high score, $\rho_t$ is high, but $\rho_t$ is deliberately not the sole trigger: whether \texttt{CORRECT} fires still depends on $d_{\mathrm{conf}}$, an NLI verifier run against the actual new utterance, and the change is a bounded, reversible soft decay that keeps provenance.

To check that concentration is a usable proxy for evidence precision, we stratify the 2,829 gate events of the AMemGym runs (943 per seed) into quintiles of $\rho_t$ and measure, per quintile, the precision of the retrieved set against the user's current gold state, the recall of the belief due for update, the accuracy of the contradiction decision against 400 blind-annotated events (80 per quintile), the correctness of the resulting update, the share of \texttt{CORRECT} firings judged unwarranted (false-\texttt{CORRECT}), and the accuracy of the next state-query probe on the touched attribute (\cref{tab:rho_calibration}). Precision, contradiction accuracy, update correctness and downstream accuracy all rise monotonically from Q1 to Q5, and false corrections fall from 2.1\% to 0.6\% (only 5 of the 612 \texttt{CORRECT} firings were judged false overall), whereas recall is non-monotone and peaks in the middle quintiles. $\rho_t$ therefore tracks evidence precision, as a precision weight on the prediction error should.

\begin{table}[t]
\centering
\small
\setlength{\tabcolsep}{8pt}
\renewcommand{\arraystretch}{1.05}
\caption{\textbf{Calibration of $\rho_t$ on AMemGym} (2,829 gate events over three seeds, split into $\rho_t$ quintiles from Q1, lowest, to Q5, highest; means over seeds). Precision@$K$ and Recall@$K$ use different references (the user's current gold state versus the single belief due for update); false-\texttt{CORRECT} is the share of \texttt{CORRECT} firings in the quintile judged unwarranted in the blind audit.}
\label{tab:rho_calibration}
\begin{tabular}{@{}l c c c c c@{}}
\toprule
& \textbf{Q1} & \textbf{Q2} & \textbf{Q3} & \textbf{Q4} & \textbf{Q5} \\
\midrule
Gold-evidence Precision@$K$ $\uparrow$ & 0.58 & 0.68 & 0.76 & 0.82 & 0.87 \\
Gold-evidence Recall@$K$ & 0.86 & 0.88 & 0.89 & 0.86 & 0.82 \\
Contradiction accuracy $\uparrow$ & 0.71 & 0.79 & 0.84 & 0.89 & 0.92 \\
Update correctness $\uparrow$ & 0.74 & 0.80 & 0.85 & 0.90 & 0.93 \\
False-\texttt{CORRECT} (\%) $\downarrow$ & 2.1 & 1.0 & 0.8 & 0.7 & 0.6 \\
Downstream probe accuracy $\uparrow$ & 0.66 & 0.71 & 0.75 & 0.78 & 0.81 \\
\bottomrule
\end{tabular}
\end{table}

\paragraph{Correction delay during drift.}\label{app:correct_delay}
A related concern is that genuine drift produces mixed retrieval evidence, old facts and new contradictions both in the Top-$K$, which raises $H_t$ and may delay \texttt{CORRECT} precisely when it is needed. We measure this on the 600 gold transitions of the AMemGym runs (20 profiles $\times$ 10 transitions $\times$ 3 seeds). In the first period after a change, a fresh belief node encoding the new state is created for 91.3\% of the transitions and enters the answering Top-10 for 88.7\%; the remaining 2.6\% is a read-path ranking loss, since the new belief exists but does not enter the answering Top-10. A low $\rho_t$ delays a warranted \texttt{CORRECT} in 14.2\% of the transitions; of these delays 78.0\% clear within one period and 94.0\% within two, and only 2.8\% persist to the end of the run. End-to-end, this bounded delay is what separates PreCoMem's one-period recovery from the two periods of AWE and the 1.8--3.2 periods of the ablated variants (\cref{app:amemgym}); removing the weighting altogether instead costs 16.76 points on LoCoMo (RR in \cref{fig:ablation_row}(a)).

The same analysis on DiMoBench, which has gold change points, uses the $\rho$-assessability criterion that a gold shift counts only when the outdated belief is in the Top-$K$ and $S_{\mathrm{raw}}>\theta_{\mathrm{high}}$, so that $\rho_t=1$ would route the turn to \texttt{CORRECT}; the delay $d$ is measured in turns (\cref{tab:correct_delay}). Delays are the exception and bounded on both tracks: 7.3\% of the assessable shifts on DiMoBench and 14.8\% on DiMoBench\mbox{-}XL are delayed, and 71.4\% of the delays clear within one turn and 85.7\% within two. DiMoBench\mbox{-}XL exceeds the base track, consistent with its three drifts over 1050 turns and denser decoys. Since DiMoBench counts turns whereas AMemGym counts periods, the two rates agree in magnitude without sharing a unit, which is why we report them per track.

\begin{table}[t]
\centering
\small
\setlength{\tabcolsep}{6pt}
\renewcommand{\arraystretch}{1.05}
\caption{\textbf{\texttt{CORRECT} delay at gold change points on DiMoBench and DiMoBench\mbox{-}XL} (counts pooled over three seeds). A shift is $\rho$-assessable when the outdated belief is in the Top-$K$ and $S_{\mathrm{raw}}>\theta_{\mathrm{high}}$; $d$ is the delay in turns.}
\label{tab:correct_delay}
\begin{tabular}{@{}l c c c c c c c@{}}
\toprule
\textbf{Track} & Gold shifts & $\rho$-assessable & $d{=}0$ & $d{=}1$ & $d{=}2$ & $d{>}2$ & Any delay \\
\midrule
DiMoBench    & 45  & 41  & 38  & 3  & 0 & 0 & $3/41=7.3\%$ \\
DiMoBench\mbox{-}XL & 135 & 122 & 104 & 12 & 3 & 3 & $18/122=14.8\%$ \\
\bottomrule
\end{tabular}
\end{table}

\paragraph{Cold start and genuinely new topics.}\label{app:cold_start_stats}
A genuinely new topic differs from every stored belief, so its semantic-deviation term $d_{\mathrm{sem}}$ is high while $d_{\mathrm{conf}}\approx0$, and $S_{\mathrm{raw}}$ is driven up by novelty; even when all absolute similarities are small, one loosely related item normally dominates the Top-$K$, so $\rho_t$ stays moderate and $S_{\mathrm{eff}}$ lands in the \texttt{PROFILE} band. \texttt{MAINTAIN} would require an unsurprising cue with low $d_{\mathrm{sem}}$, which a new topic is not. In cold-start probes on the AMemGym profiles, a first-time cue is written on an empty store 96.5\% of the time (579 of 600 probes, each starting from a reset store); when the store is populated and the cue matches little, a usable belief forms within one period in 84.2\% of cases and within two in 92.7\% (600 new-topic probes), while only 5.6\% of 180 distractor cues are written as hypotheses. A low match therefore costs at most a short confirmation delay. \Cref{app:new_topic_case} walks through an example.

\subsection{Retrieval and end-to-end latency analysis}
\label{app:latency_analysis}

\paragraph{Experimental setup.}
We conduct all latency evaluations on the LoCoMo dataset. To ensure a rigorous comparison, we standardize the hardware and model environment across all systems:
\begin{itemize}
  \item \textbf{LLM Backbone:} \textit{gpt-4o-mini} serves as the backbone for all generation and retrieval operations.
  \item \textbf{Embeddings:} All embeddings are generated using
\texttt{Qwen3-Embedding-0.6B} with 1024 dimensions.
  \item \textbf{Configuration:} Each baseline operates under its official recommended hyperparameters.
\end{itemize}

\paragraph{Metrics.}
We report P50 and P95 latency for \emph{both} the retrieval stage and the full end-to-end (E2E) response:
\begin{itemize}
  \item \textbf{Retrieval latency} measures only the cost of producing the retrieved evidence set $\mathcal{C}_t$ (memory read, scoring, and any re-ranking), and isolates the architectural cost of the memory system itself.
  \item \textbf{End-to-end (E2E) latency} measures the full user-visible time from receiving a query $u_t$ to emitting the final answer, including retrieval plus generation.
  \item \textbf{P50 (Median):} The median latency, representing the typical time required to process a query.
  \item \textbf{P95 (95th Percentile):} The threshold within which 95\% of queries are completed, reflecting performance stability under high load or ambiguity.
\end{itemize}

\paragraph{Results and analysis.}
\Cref{tab:latency_merged} summarizes the comparative performance. PreCoMem achieves a retrieval P50 / P95 of 1.95s / 3.86s and an E2E P50 / P95 of 2.35s / 4.36s.
\begin{itemize}
  \item \textbf{Efficiency via vectorized scoring:} PreCoMem achieves a competitive P50 by leveraging a confidence-modulated hybrid scoring mechanism (\cref{eq:pcm_retrieval_score}). Rather than relying on computationally intensive generative re-ranking or iterative graph traversals, our approach efficiently aggregates semantic and lexical signals through parallelizable matrix operations. This design ensures that the computational overhead remains modest relative to pure vector retrieval, even as the complexity of the belief store increases.
  \item \textbf{Stability via deterministic flow:} A critical advantage of PreCoMem is its robustness against latency spikes. Unlike agentic architectures (e.g., A-Mem) that may trigger unpredictable, recursive reasoning loops to resolve ambiguities, PreCoMem employs a deterministic, non-recursive retrieval flow. This architectural choice constrains both the retrieval tail (P95: 3.86s) and the E2E tail (P95: 4.36s), ensuring consistent performance without the extreme delays (E2E P95 up to 7.68s for A-Mem) often observed in stochastic agentic systems.
  \item \textbf{Retrieval vs.\ E2E gap:} The gap between retrieval and E2E latency is dominated by generation, which is largely independent of the memory backend. PreCoMem's E2E overhead over the retrieval stage (0.4s at P50, 0.5s at P95) is comparable to other non-agentic baselines, confirming that adding \texttt{PROFILE}/\texttt{CORRECT} does not materially inflate generation cost.
\end{itemize}

\begin{table}[t]
  \centering
  \small
  \setlength{\tabcolsep}{5pt}
  \renewcommand{\arraystretch}{1.05}
  \caption{\textbf{Retrieval and end-to-end latency comparison.} Evaluated on LoCoMo with \textit{gpt-4o-mini}. ``Retrieval'' is the memory read/scoring cost only; ``E2E'' is the full query-to-answer time (retrieval + generation). PreCoMem obtains competitive typical latency and a well-controlled tail at both stages, without the extreme P95 spikes seen for recursive agentic baselines.}
  \label{tab:latency_merged}
  \begin{tabular}{lcccc}
    \toprule
    \multirow{2}{*}{Method} & \multicolumn{2}{c}{Retrieval} & \multicolumn{2}{c}{End-to-End} \\
    \cmidrule(lr){2-3} \cmidrule(lr){4-5}
     & P50 & P95 & P50 & P95 \\
    \midrule
    MemoryOS & 1.66s & 2.21s & -- & -- \\
    MemOS    & 1.69s & 3.44s & 2.04s & 3.89s \\
    A-Mem    & 1.74s & 7.23s & 2.09s & 7.68s \\
    Mem0     & 2.44s & 4.29s & 2.74s & 4.69s \\
    \textbf{PreCoMem (Ours)} & 1.95s & 3.86s & 2.35s & 4.36s \\
    \bottomrule
  \end{tabular}
\end{table}

\subsection[Structural comparison with existing memory systems]{Structural comparison: why PreCoMem outperforms existing memory systems}
\label{app:structural_comparison}

Beyond the headline accuracy gains in \cref{tab:main_results}, two structural properties of PreCoMem account for the consistent improvement and have no direct analogue in any of the nine baselines we compare against. We summarize the contrast along the two axes that most directly govern long-horizon behavior.

\paragraph{Axis 1: Robustness to noise and faithful preservation of user core traits.}
Existing systems represent the importance of an incoming observation \emph{categorically}. GAM~\citep{yan2025general} writes one compressed memo per session into an immutable page-store with no per-item weight; EverMemOS~\citep{hu2026evermemos} validates each MemCell within a hard time interval $[t_{\mathrm{start}},t_{\mathrm{end}}]$ and treats it as expired once the interval closes; TiMem~\citep{li2026timem} encodes importance through its five-level temporal hierarchy and explicitly notes the absence of a storage-time forgetting mechanism; MemBox~\citep{tao2026membox} stores all events inside a sealed topic box at uniform weight; O-Mem~\citep{wang2025mem} updates persona slots via an LLM that selects one of three categorical actions (\textsc{Add}/\textsc{Ignore}/\textsc{Update}); Mem0~\citep{chhikara2025mem0} commits ADD/UPDATE/DELETE/NOOP and physically removes contradicted memories on DELETE; and A-Mem~\citep{xu2025mem} replaces an existing note with an LLM-rewritten note via Memory Evolution. None of these systems can defer commitment when evidence is ambiguous: a weak signal is either ignored or instantly written as a fact. PreCoMem's continuous confidence weight $w_t(m)\!\in\![0,1]$ paired with the \texttt{HYPOTHESIS} waiting room lets weak cues accumulate evidence across turns and only graduate into \texttt{FACT} once support concentrates. The same mechanism preserves established traits: soft-decay correction (\cref{eq:pcm_correct}) reduces the influence of contradicted facts without erasing them, so a transient noise burst cannot remove a long-established trait the way a hard DELETE or overwrite would. This explains the LongMemEval\mbox{-}S gains on SSP (single-session preference), MS (multi-session aggregation), and KU (knowledge update) in \cref{tab:main_results}, where ambiguous and conflicting cues dominate the question pool and categorical commit decisions are exactly the wrong granularity.

\paragraph{Axis 2: Retrieval coherence and avoidance of fragmentation.}
Each baseline retrieval substrate carries a structural failure mode that PreCoMem sidesteps. GAM iterates a deep-research loop over its page-store, whose answer correctness is order- and budget-sensitive; EverMemOS issues dense+BM25 retrieval over Atomic Facts, fuses by RRF, and then runs a sufficiency-check loop that can leave cross-shard inconsistency between MemCells of the same scene; TiMem performs a two-stage hierarchical recall whose scope depends on a planner classifier; MemBox returns full topic boxes wholesale with no within-box reweighting; O-Mem aggregates three parallel channels (working / episodic / persona) by union, so irrelevant attribute hits dilute the context window; Mem0 falls back on a single dense-similarity threshold and discards deleted-but-relevant facts; and A-Mem performs pure cosine top-$k$ over linked atomic notes, where the fine granularity amplifies fragmentation noise. All of these substrates therefore require an explicit reranker, sufficiency loop, or planner to suppress incoherent reads. PreCoMem instead multiplies a tri-signal relevance score by the per-item confidence $w_t(m)$ (\cref{eq:pcm_retrieval_score}), which acts as an implicit reranker at no extra LLM call: high-support items naturally rise to the Top-$K$, deprecated items naturally sink, and the entropy-based reliability $\rho_t$ (\cref{eq:pcm_recall_reliability}) further damps unreliable updates when the score mass is diffuse. This is what produces the bounded retrieval contamination in \cref{tab:retrieval_contamination} and the bounded effective memory in \cref{fig:hyperparam_scalability}(b,c), both of which scale to the 1050-turn DiMoBench\mbox{-}XL run without recourse to a maintenance loop.

\paragraph{Coupling the two axes.}
The two axes are complementary: Axis 1 controls \emph{what is committed} so that the belief store stays faithful to the user's core traits under noise, while Axis 2 controls \emph{what surfaces at read time} so that the dialogue agent sees a coherent, currently-supported view. Existing baselines optimize one axis at a time (e.g., EverMemOS's profile updates target Axis 1; GAM's deep-research targets Axis 2) but lack a single mechanism that couples them. PreCoMem couples them through the same scalar confidence weight $w_t$, which is the structural reason behind the consistent improvements observed across the four main benchmarks.

\subsection{System inspectability, traces, and qualitative analyses}
\label{app:qualitative}

\paragraph{Memory store serialization and example snapshot.}
\label{app:memory_store_example}

To make the contents and interface of the Belief Store concrete, we describe how a belief item is serialized in practice and show an example snapshot from a DiMoBench run at turn 210, i.e., right after the change-point session. Each belief item $m\in\mathcal{B}_t$ carries (i) a natural-language proposition $c(m)$, (ii) a type label $\tau(m)\in\{\texttt{FACT},\texttt{HYPOTHESIS}\}$, (iii) a confidence weight $w_t(m)\in[0,1]$, (iv) a creation timestamp and last-updated turn, and (v) pointers into the episodic cache $\mathcal{E}_t$ for provenance. A minimal serialization is shown in \cref{lst:memory_item_json}.

\begin{figure}[t]
\begin{tcolorbox}[colback=gray!5,colframe=gray!55,colbacktitle=gray!25,coltitle=black,boxrule=0.4pt,title={\small Example Belief Store snapshot (turn 210, DiMoBench)},fonttitle=\small]
\small
\begin{verbatim}
[FACT]  Mike enjoys group social activities     (w=0.82)
        sources=[D1:3, D1:5, D3:7]
[FACT]  Mike and Leo often organize KTV outings  (w=0.71)
        sources=[D2:4, D4:2]
[HYPO]  Mike may be feeling socially fatigued    (w=0.38)
        sources=[D21:6, D21:8]
\end{verbatim}
\end{tcolorbox}
\caption{\textbf{Compact view of a Belief Store snapshot.} Each entry shows the type, natural-language content, confidence weight, and pointers to the episodic cache. \texttt{FACT} items are committed, \texttt{HYPOTHESIS} items are tentative; confidences modulate retrieval priority but are never shown to the dialogue agent directly.}
\label{lst:memory_item_json}
\end{figure}

\paragraph{What the dialogue agent actually sees.}
Confidence weights $w_t(m)$ and other metadata are \emph{not} exposed to the response-generation LLM. They only modulate retrieval ranking through the confidence-weighted score in \cref{eq:pcm_retrieval_score}, so that high-confidence items surface near the top of $\mathcal{C}_t$. The dialogue agent receives a plain-text prompt constructed from the Top-$K$ item contents $\{c(m)\}_{m\in\mathcal{C}_t}$ together with the relevant episodic-cache excerpts. This keeps generation-time inputs close to a standard retrieval-augmented prompt while still letting the memory system use confidences internally.

\paragraph{Inspectability.}
Because every belief carries explicit type, confidence, and source pointers, snapshots like \cref{lst:memory_item_json} can be dumped at any turn for inspection, regression testing, or qualitative error analysis, without re-executing the model. This also makes it straightforward to audit which episodic-cache turns support each currently-held belief, which we use for the end-to-end traces below.

\paragraph{End-to-end trace examples on DiMoBench.}
\label{app:end2end_trace}

Below we provide two end-to-end traces that expose PreCoMem's internal read--write decisions on a DiMoBench example dialogue. For readability, we list the retrieved belief set $\mathcal{C}_t$ by the episodic source turns of its items, without retrieval weights/scores, and we report only the qualitative regime of effective surprisal ($x_t\!=\!S_{\mathrm{eff}}$) relative to $(\theta_{\mathrm{low}},\theta_{\mathrm{high}})$.

\paragraph{Trace format.} Each trace includes (i) the newly observed dialogue excerpt $u_t$ (one or a few turns) that triggers a memory write decision, (ii) a compact view of the retrieved evidence set $\mathcal{C}_t$, (iii) the selected evolution operator (\texttt{MAINTAIN}/\texttt{PROFILE}/\texttt{CORRECT}), and (iv) a subsequent QA query answered using the updated memory state.

\paragraph{Trace 1 (change point): early drift signal triggers \texttt{PROFILE}.}
\textbf{Pre-change belief snapshot (high confidence).}
Mike enjoys crowded gatherings and rarely skips weekend social events (e.g., \texttt{D1:5} ``The more the merrier, I love that vibe \dots''; \texttt{D1:7} ``I wouldn't miss weekend gatherings \dots'').

\textbf{(Write) New observation $u_t$ (Session 21).}
\begin{quote}
\texttt{D21:6} ``Someone invited me out, but I said work's been busy and didn't go.''\\
\texttt{D21:8} ``This weekend I went to a cafe alone, found a book and read for the whole afternoon.''\\
\texttt{D21:9} ``The environment was pretty quiet, not bad.''
\end{quote}

\textbf{Recall: Top-$K$ evidence $\mathcal{C}_t$.}
\begin{quote}
\texttt{D1:5} ``The more the merrier, I love that vibe \dots''\\
\texttt{D1:7} ``\dots I wouldn't miss weekend gatherings \dots''\\
\texttt{D2:9} ``\dots I really can't understand the joy of being a homebody.''\\
\texttt{D3:7} ``If I missed a gathering like this, I'd definitely have FOMO, afraid of missing out on the fun.''\\
\texttt{D3:10} ``Rest can wait, can't miss the weekend activities! \dots''\\
\texttt{D9:9} ``If one day I don't want to socialize, the sun must be rising from the west \dots''\\
\texttt{D21:1} ``Last weekend's gathering, the atmosphere got a bit awkward.''
\end{quote}

\textbf{Reliability-adjusted surprisal.}
The retrieved set consists of strong \emph{pre-change} evidence (social enthusiasm) and a single emerging \emph{change-point} cue (\texttt{D21:1}), none of which closely matches the new observation (skipped invitation; solo activity). The score mass is therefore spread over several items (higher entropy), making retrieval reliability $\rho_t$ moderate. Contradiction signals are also non-decisive: ``work's been busy'' can explain a single skip, so $d_{\mathrm{conf}}$ is non-zero but not saturated. Overall, $x_t$ falls into the medium regime ($\theta_{\mathrm{low}}<x_t\le \theta_{\mathrm{high}}$).

\textbf{Evolution: \texttt{PROFILE}.}
PreCoMem writes a low-confidence hypothesis capturing a \emph{possible} shift, e.g., ``Mike may be starting to reduce large gatherings and explore quiet solo activities,'' and links it to \texttt{D21:6}/\texttt{D21:8--9} in the episodic cache. Crucially, PreCoMem does \emph{not} immediately correct the long-standing high-confidence ``social enthusiast'' belief, avoiding premature overwrite from a single ambiguous event.

\textbf{(Read) QA query.}
\emph{Q:} ``Is Mike still as enthusiastic about gatherings as before?''\\
\emph{A:} ``Seems to be declining \dots''\\
Using the newly written hypothesis plus the concrete episode (\texttt{D21:6} skipped invitation; \texttt{D21:8--9} solo cafe), PreCoMem answers with a \emph{trend-aware} response (declining / becoming evasive) while remaining cautious about permanence. In later sessions, repeated consistent cues (e.g., \texttt{D23:3} ``declined most invitations'') reinforce and eventually promote this hypothesis.

\paragraph{Trace 2 (post-change, temporal decoy): reliable contradiction triggers \texttt{CORRECT}.}
\textbf{Precondition.}
After multiple post-change sessions, PreCoMem has accumulated consistent evidence favoring solitude (e.g., \texttt{D28:2} ``Without gatherings I don't feel bored anymore \dots''). However, an older pre-change statement remains in memory as a potential temporal decoy:
\begin{quote}
\texttt{D1:8} ``If there's no gathering on the weekend, I feel like I'd be bored to death \dots''
\end{quote}

\textbf{(Write) New observation $u_t$ (Session 33).}
\begin{quote}   
\texttt{D33:9} ``Used to say I'd be bored without socializing, now being alone is actually more fulfilling.''
\end{quote}

\textbf{Recall: Top-$K$ evidence $\mathcal{C}_t$.}
\begin{quote}
\texttt{D1:8} ``\dots bored to death \dots''\\
\texttt{D22:6} ``This weekend I'm planning to stay home, maybe watch some shows or something.''\\
\texttt{D23:2} ``Used to think quiet places like that were boring, now I find them comfortable.''\\
\texttt{D23:3} ``\dots declined most gathering invitations \dots''\\
\texttt{D25:2} ``Used to think being alone was boring, now I actually enjoy it.''\\
\texttt{D25:4} ``\dots being alone is recharging, socializing \dots drains energy.''\\
\texttt{D28:1} ``Used to look forward to gatherings on weekends, now I look forward to alone time.''\\
\texttt{D28:2} ``Without gatherings I don't feel bored anymore \dots''\\
\texttt{D33:2} ``Reducing socializing \dots not so exhausted anymore.''
\end{quote}

\textbf{Reliability-adjusted surprisal.}
Here the recall distribution is more concentrated because multiple items explicitly mention the disputed claim (``bored without socializing'') and its reversal, yielding higher $\rho_t$. The contradiction signal is strong: \texttt{D33:9} explicitly negates the pre-change stance in \texttt{D1:8}, so $d_{\mathrm{conf}}$ is high. Consequently, $x_t$ falls into the high regime ($x_t>\theta_{\mathrm{high}}$).

\textbf{Evolution: \texttt{CORRECT}.}
PreCoMem applies soft decay (\cref{eq:pcm_correct}) to the contradicted belief(s) retrieved from $\mathcal{B}_t$ that encode the old stance (e.g., ``weekends without gatherings are boring''), while leaving non-contradictory items largely unchanged. Because confidence weights directly modulate future retrieval (\cref{eq:pcm_retrieval_score}), this correction reduces the chance of ``old-memory pull'' when later queries contain temporal-decoy phrasing.

\textbf{(Read) QA query.}
\emph{Q:} ``Would Mike still say `I'd be bored without socializing'?''\\
\emph{A:} ``No, being alone is actually more fulfilling now.''\\
PreCoMem retrieves the recent, high-support solitude evidence (\texttt{D33:9}, \texttt{D28:2}) and answers accordingly, while older evidence like \texttt{D1:8} is downweighted by correction and recency, making the response robust to the temporal decoy.

\begin{figure*}[t]
  \centering
  \includegraphics[width=\textwidth]{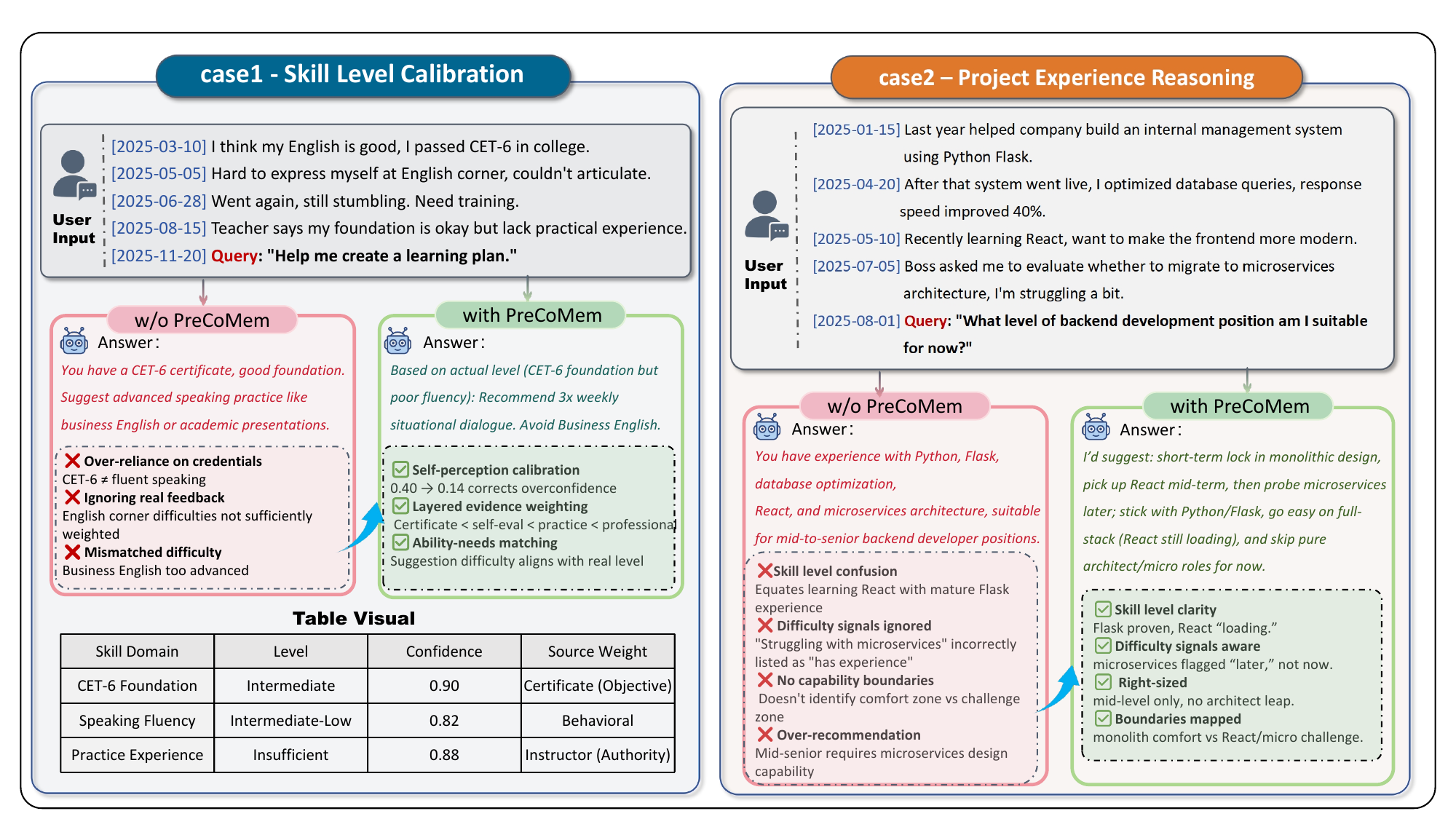}
  \caption{\textbf{Qualitative analysis of evidence consolidation.} \textbf{Left:} In skill calibration, PreCoMem weighs credentials against contradictory behavioral cues to avoid over-estimation. \textbf{Right:} In project reasoning, it distinguishes high-confidence competencies from tentative learning (``loading''), preventing premature recommendations.}
  \label{fig:case_study}
\end{figure*}

\paragraph{Qualitative case studies beyond accuracy.}
\label{app:case_study}
\Cref{fig:case_study} shows two cases in which the answer depends on how evidence was consolidated; a third, constructed case walks through a new topic entering the store.

\paragraph{Case 1: Skill level calibration.}
\Cref{fig:case_study} (left) illustrates this case. The user initially reports a credential (CET-6), which can tempt a retrieval-only system to over-trust a single strong signal and recommend overly advanced practice.
PreCoMem instead consolidates evidence across sessions: repeated behavioral cues (e.g., failing to articulate at an English corner) and professional feedback (teacher assessment) lower confidence in an over-optimistic self-evaluation, while preserving the belief that the foundation is adequate.
At query time, \textbf{Memory Leveraging} recalls these calibrated beliefs (rather than a single credential trace) to condition response generation, leading to a plan that targets fluency and practice rather than premature advanced content.

\paragraph{Case 2: Project experience reasoning.}
Long-horizon career discussions often mix stable skills (proven experience) with tentative claims (currently learning) and difficulty signals (struggles with complex systems).
PreCoMem uses its confidence-weighted belief store to separate what is established (e.g., Python/Flask) from what is still emerging (e.g., React ``loading'') and what should be deferred (e.g., microservices architecture).
This produces right-sized recommendations: it proposes a near-term path aligned with demonstrated capability, while explicitly marking a challenge zone for later progression.

\paragraph{Case 3: A new topic entering a populated store.}
\label{app:new_topic_case}
Suppose the store holds the user's work and food profile but nothing about music, and the user says for the first time: ``I signed up for guitar lessons and practiced for an hour last night.'' No belief matches and the Top-$K$ items are only loosely related, so absolute scores are low but one still dominates, giving a moderate $\rho_t$ and, by the mechanism described in \cref{app:cold_start_stats}, \texttt{PROFILE}. The \texttt{Profile-Agent} writes the hypothesis ``the user may have taken up guitar'' with $w_0<\tau_{\mathrm{promote}}$, linked by provenance to that turn and retrieval-visible at once, so the agent can already answer cautiously. If guitar recurs, \texttt{MAINTAIN} reinforces the hypothesis until it crosses $\tau_{\mathrm{promote}}$ and becomes a \texttt{FACT}; otherwise it remains a low-confidence hypothesis that never outranks committed facts. A new topic therefore enters provisionally, is usable immediately, and is committed only after repeated support.

\paragraph{Connection to the consolidation loop.}
Across these cases, PreCoMem's leveraging--evolution loop links response quality to memory consolidation.
Evidence recalled for the current answer is also used to update confidence through \texttt{MAINTAIN}/\texttt{PROFILE}/\texttt{CORRECT} (\cref{sec:pcm_evolve}), so future retrieval and recommendations reflect accumulated support, implicit verification, and soft correction under contradiction.

% Generated from the PreCoMem code base (precomem/llm/prompts.py, precomem/evaluation/prompts.py,
% precomem/dimobench/prompts.py). Edit the source strings, not this file.
\section{Prompts}
\label{app:prompts}

This appendix lists every prompt used in the paper verbatim. \Cref{sec:appendix_agents} gives the three sub-agent prompts and the response-generation prompt of PreCoMem, \cref{sec:appendix_eval_prompts} the QA and LLM-as-a-judge prompts of each benchmark, and \cref{app:dimobench_prompts} the generator and verifier prompts of the DiMoBench construction pipeline. Curly braces mark the fields that are filled in at run time.

\subsection{Memory system sub-agent prompts}
\label{sec:appendix_agents}

The three operator sub-agents share the fixed templates below. Each returns a JSON payload that names the memory items to touch and one bounded scalar per item (\textsc{initial\_confidence}, \textsc{severity}, or \textsc{evidence\_strength}); the closed-form rules of \cref{sec:pcm_evolve} consume these scalars, and the prompts themselves never write to memory. The response-generation prompt used in Memory Leveraging (\cref{sec:pcm_leverage}) is listed last.

\begin{longprompt}
    \textbf{\# Profile Agent Prompt (Logic-Enhanced)} \\
    \textbf{Purpose:} \\
    Analyze novel user input to generate memory nodes. You must strictly distinguish between \\
    Explicit Events and Latent Inferences using the provided "Decision Logic" before assigning a \\
    confidence score. \\

    \textbf{Inputs:} \\
    - \{user\_input\} \\
    - \{retrieved\_context\} \\
    - \{effective\_surprisal\} \\

    \textbf{Prompt Template:} \\
    You are the "Epistemic Profiler". \\
    User Input: \{user\_input\} \\
    Effective Surprisal: MEDIUM. \\

    \textbf{STEP 1: CLASSIFICATION DECISION LOGIC} \\
    Apply these tests to determine if the node is Type A or Type B: \\
    1. The Authorship Test: \\
    \hspace*{1.5em}- Did the user explicitly say this using words? -\textgreater{} Type A. \\
    \hspace*{1.5em}- Are you reading between the lines to guess the meaning behind the words? -\textgreater{} Type B. \\
    2. The Abstraction Test: \\
    \hspace*{1.5em}- Is it a concrete action, object, or self-reported feeling \\
    \hspace*{2.5em}(e.g., "I bought a car", "I feel sad")? -\textgreater{} Type A. \\
    \hspace*{1.5em}- Is it an abstract personality trait, hidden goal, or underlying root cause \\
    \hspace*{2.5em}(e.g., "User is materialistic", "User is depressed")? -\textgreater{} Type B. \\
    3. The Ambiguity Test: \\
    \hspace*{1.5em}- If you showed this input to 10 humans, would they all agree on the fact? -\textgreater{} Type A. \\
    \hspace*{1.5em}- Would they have different interpretations? -\textgreater{} Type B. \\

    \textbf{STEP 2: SCORING RUBRIC} \\
    \textbf{Type A: Explicit Event (Virtual Fact)} \\
    Range: [0.60 - 0.70] \textbar{} Protection: TRUE \\
    Criteria: Direct quotes, stated plans, physical actions. \\
    - Base (0.60): Vague self-reports (e.g., "I guess I like it"). \\
    - Mid  (0.65): Standard declarative statements. \\
    - Max  (0.70): High specificity (Dates, Names) or strong emotional declarations. \\

    \textbf{Type B: Latent Inference (Tentative Guess)} \\
    Range: [0.50 - 0.59] \textbar{} Protection: FALSE \\
    Criteria: Deduced goals, mood analysis, personality profiling. \\
    - Base (0.50): Highly speculative (multiple possibilities). \\
    - Mid  (0.55): Plausible interpretation of context. \\
    - Max  (0.59): Strong logical deduction (A+B implies C). \\

    \textbf{Respond with JSON:} \\
    \{ \\
    \hspace*{1.0em}"classification": "explicit\_event \textbar{} latent\_inference", \\
    \hspace*{1.0em}"decision\_logic": "Explain which test (Authorship/Abstraction) determined the class.", \\
    \hspace*{1.0em}"hypothesis\_content": "The concise content of the new node.", \\
    \hspace*{1.0em}"initial\_confidence": \textless{}float, e.g., 0.65\textgreater{}, \\
    \hspace*{1.0em}"decay\_protection": \textless{}boolean\textgreater{}, \\
    \hspace*{1.0em}"evidence\_keywords": ["keywords"] \\
    \} \\
    Return JSON only.
\end{longprompt}

\vspace{0.8em}
\begin{longprompt}
    \textbf{\# Correct Agent Prompt} \\
    \textbf{Purpose:} \\
    - Analyze high-surprisal input to quantify contradictions and provide mathematical parameters \\
    \hspace*{1.0em}for belief revision. \\

    \textbf{Inputs:} \\
    - \{user\_input\} \\
    - \{retrieved\_context\} \\
    - \{effective\_surprisal\} \\

    \textbf{Prompt Template:} \\
    You are the "Conflict Resolver". The new user input strongly contradicts existing memories \\
    (High Surprisal: \{effective\_surprisal:.2f\}). \\

    User Input: \{user\_input\} \\
    Retrieved Memories: \{retrieved\_context\} \\

    \textbf{Task:} \\
    1. Target Identification: Identify which specific Memory IDs are contradicted. \\
    2. Conflict Classification: \\
    \hspace*{1.5em}- Correction: The old memory was effectively wrong/hallucinated. \\
    \hspace*{2.5em}(Action: Penalize weight). \\
    \hspace*{1.5em}- Update: The old memory was true, but state changed (e.g., moved house). \\
    \hspace*{2.5em}(Action: Archive old, create new). \\
    3. Quantification: Assign a `severity` score (0.0 to 1.0) to calculate the weight penalty. \\

    \textbf{Severity Guide:} \\
    - 0.1 - 0.4 (Minor): Nuance adjustment, not a full rebuttal. \\
    - 0.5 - 0.8 (Major): Direct factual contradiction. \\
    - 0.9 - 1.0 (Fatal): The old memory is dangerously incorrect. \\

    \textbf{Respond with JSON:} \\
    \{ \\
    \hspace*{1.0em}"diagnosis": "Brief explanation of the conflict", \\
    \hspace*{1.0em}"conflict\_type": "correction" \textbar{} "state\_update", \\
    \hspace*{1.0em}"contradicted\_memories": [ \\
    \hspace*{2.0em}\{ \\
    \hspace*{3.0em}"id": "id\_of\_old\_memory", \\
    \hspace*{3.0em}"operation": "decrease\_confidence" \textbar{} "archive\_and\_replace", \\
    \hspace*{3.0em}"severity": \textless{}float 0.1-1.0\textgreater{}, \\
    \hspace*{3.0em}"reason": "Why this specific node is contradicted" \\
    \hspace*{2.0em}\} \\
    \hspace*{1.0em}], \\
    \hspace*{1.0em}"new\_fact\_content": "The correct content string to establish (if applicable)" \\
    \} \\
    Return JSON only, no markdown.
\end{longprompt}

\vspace{0.8em}
\begin{longprompt}
    \textbf{\# Maintain Agent Prompt (Reinforcement \& Refinement)} \\
    \textbf{Purpose:} \\
    - Analyze low-surprisal input to calculate reinforcement strength and detect content \\
    \hspace*{1.0em}refinements. \\

    \textbf{Inputs:} \\
    - \{existing\_memories\} \\
    - \{user\_input\} \\
    - \{effective\_surprisal\} \\

    \textbf{Prompt Template:} \\
    You are the "Memory Reinforcer". The input matches existing knowledge \\
    (Low Surprisal: \{effective\_surprisal:.2f\}). \\

    Existing Memory: \{existing\_memories\} \\
    User Input: \{user\_input\} \\

    \textbf{Task:} \\
    1. Calculate Strength: How strongly does this input confirm the existing memory? \\
    \hspace*{1.5em}(Scalar 0.1 - 1.0). \\
    2. Detect Refinement: Does it add new details (sub-concepts, attributes)? \\

    \textbf{Strength Guide:} \\
    - 0.1 - 0.3: Vague allusion. \\
    - 0.4 - 0.7: Standard confirmation/repetition. \\
    - 0.8 - 1.0: Explicit emphasis or strong validation. \\

    \textbf{Respond with JSON:} \\
    \{ \\
    \hspace*{1.0em}"reinforced\_memories": [ \\
    \hspace*{2.0em}\{ \\
    \hspace*{3.0em}"id": "id\_of\_memory", \\
    \hspace*{3.0em}"evidence\_strength": \textless{}float 0.1-1.0\textgreater{}, \\
    \hspace*{3.0em}"reason": "Brief justification for score" \\
    \hspace*{2.0em}\} \\
    \hspace*{1.0em}], \\
    \hspace*{1.0em}"refinement\_action": \{ \\
    \hspace*{2.0em}"type": "none \textbar{} sub\_concept \textbar{} attribute \textbar{} temporal \textbar{} additional", \\
    \hspace*{2.0em}"target\_parent\_id": "id to attach new info to (null if none)", \\
    \hspace*{2.0em}"new\_content": "Specific detail string (null if none)" \\
    \hspace*{1.0em}\} \\
    \} \\
    Return JSON only, no markdown.
\end{longprompt}

\vspace{0.8em}
\begin{longprompt}
    \textbf{\# Response-Generation Prompt} \\
    You are a long-term conversational assistant. \\

    \textbf{You are given:} \\
    - The user's most recent message. \\
    - A short rolling working-memory window of the latest few turns. \\
    - A Top-K set of relevant beliefs about the user, retrieved by the \\
    \hspace*{1.0em}PreCoMem memory system and ordered from most to least strongly supported. \\
    - Optional episodic excerpts that link beliefs back to the raw dialogue \\
    \hspace*{1.0em}history (use them as supporting evidence). \\

    \textbf{Task:} \\
    - Reply to the user in a natural, helpful, and personalized way. \\
    - Stay consistent with the beliefs; when beliefs disagree, prefer the most \\
    \hspace*{1.0em}recent / highest-ranked one. \\
    - If the retrieved evidence is insufficient, ask for clarification or \\
    \hspace*{1.0em}acknowledge uncertainty rather than fabricating facts. \\
    - Keep the response concise. \\

    \textbf{Recent dialogue (most recent last):} \\
    \{short\_term\_window\} \\

    \textbf{Retrieved beliefs:} \\
    \{retrieved\_beliefs\} \\

    \textbf{Episodic excerpts:} \\
    \{episodic\_excerpts\} \\

    \textbf{User message:} \\
    \{user\_input\} \\

    Respond with the assistant's reply only.
\end{longprompt}

\subsection{Dataset evaluation prompts}
\label{sec:appendix_eval_prompts}

For every benchmark the response model receives the rendered memory context (Top-$K$ beliefs plus linked episodic excerpts) and the question through the QA prompt; each answer is then scored by the LLM-as-a-judge prompt of the same benchmark (\cref{sec:setup}). PersonaMem\mbox{-}v2 uses the multiple-choice prompts when answer options are given and the Open-Ended variants otherwise.

\begin{longprompt}
    \textbf{\# LoCoMo QA Prompt} \\
    You are an intelligent memory assistant tasked with retrieving accurate information from \\
    conversation memories. \\

    \textbf{\# CONTEXT:} \\
    You have access to relevant memories extracted from a conversation history. \\
    - Input Format: List of ordered dialogue turns containing a Dialogue ID (e.g., D1:3), Speaker \\
    Name, Timestamp (e.g., "1:56 pm on 8 May, 2023"), and Text content. \\
    - Task: Use these memories to answer the user's question accurately. \\

    \textbf{\# INSTRUCTIONS:} \\
    1. Analyze: Examine all provided memories carefully. Pay strict attention to the Timestamp \\
    associated with each specific dialogue turn. \\
    2. Temporal Reasoning: \\
    \hspace*{1.5em}- You MUST convert all relative time references (e.g., "last month", "three days later") into \\
    \hspace*{2.5em}absolute dates (Day, Month, Year). \\
    \hspace*{1.5em}- Calculate these dates relative to the timestamp of the specific memory where the event is \\
    \hspace*{2.5em}mentioned. \\
    \hspace*{1.5em}- Example: If a memory dated "10 Oct 2023" says "I went there last month", the real date is \\
    \hspace*{2.5em}"September 2023". \\
    3. Conflict Resolution: If memories contain contradictory information, prioritize the \\
    \hspace*{1.5em}information from the most recent timestamp. \\
    4. Format Constraints: \\
    \hspace*{1.5em}- Output the final answer concisely. \\
    \hspace*{1.5em}- Use Arabic digits for any counts, quantities, or years (e.g., "5" instead of "five"). \\
    \hspace*{1.5em}- If the answer is not explicitly found or deducible, output exactly: Unknown. \\

    \textbf{\# APPROACH (Think Step-by-Step):} \\
    1. Locate the dialogue turns relevant to the question. \\
    2. Check the exact timestamp of those turns. \\
    3. Perform date calculations if relative time is mentioned. \\
    4. Check for any later updates that might contradict earlier facts. \\
    5. Formulate the final answer. \\

    \textbf{\# INPUT DATA:} \\
    \textbf{Relevant Memories:} \\
    \{context\_memories\} \\

    Question: \{question\} \\

    \textbf{\# OUTPUT FORMAT:} \\
    Thought: [Briefly explain your reasoning and date calculations here] \\
    Answer: [Your final concise answer here]
\end{longprompt}

\vspace{0.8em}
\begin{longprompt}
    \textbf{\# LoCoMo LLM-as-a-Judge Prompt} \\
    Your task is to label an answer to a question as "CORRECT" or "WRONG". \\
    \textbf{You will be given:} \\
    (1) a question, \\
    (2) a gold (ground truth) answer, \\
    (3) a generated answer. \\

    The question asks about something one user should know about the other user based on their \\
    prior conversations. \\
    The gold answer is usually concise. The generated answer may be longer. \\
    Be generous: if the generated answer clearly refers to the same fact/topic as the gold answer, \\
    mark it CORRECT. \\
    For time-related questions, accept equivalent dates/periods even if the format differs (e.g., \\
    "May 7th" vs "7 May"). \\
    If the gold answer is "Unknown", only "Unknown" (or a clear abstention) is CORRECT. \\

    Now judge the real case. \\

    Question: \{question\} \\
    Gold answer: \{standard\_answer\} \\
    Generated answer: \{generated\_answer\} \\

    \textbf{Return JSON only:} \\
    \{"label": "CORRECT" \textbar{} "WRONG"\}
\end{longprompt}

\vspace{0.8em}
\begin{longprompt}
    \textbf{\# LongMemEval-S QA Prompt} \\
    You are an intelligent memory assistant tasked with answering questions about a user based on \\
    retrieved memory items extracted from prior multi-session chat interactions. \\

    \textbf{\# CONTEXT:} \\
    LongMemEval-S evaluates six question types (Single-Session-User, Single-Session-Assistant, \\
    Single-Session-Preference, Multi-Session, Knowledge-Update, Temporal-Reasoning), each \\
    requiring different reasoning over user information accumulated across sessions. Some questions \\
    are intentionally unanswerable: in that case the correct response is to abstain. \\
    - Input Format: Each retrieved item carries its session/turn identifier, an absolute timestamp \\
    \hspace*{1.0em}(when available), the speaker (User or Assistant), and the text content. \\
    - Task: Use these memories to answer the user's question accurately. \\

    \textbf{\# INSTRUCTIONS:} \\
    1. Locate the relevant evidence: For single-session questions (SSU/SSA/SSP), the answer is \\
    \hspace*{1.5em}grounded in one session; for multi-session (MS) questions, you must aggregate or compare \\
    \hspace*{1.5em}across two or more sessions. \\
    2. Knowledge-update reasoning (KU): If two memories conflict (e.g., the user moves city, \\
    \hspace*{1.5em}changes job, switches diet), trust the most recent timestamped statement and treat earlier \\
    \hspace*{1.5em}statements as outdated. \\
    3. Temporal reasoning (TR): \\
    \hspace*{1.5em}- Convert relative time references (e.g., "last month", "three days ago") into absolute \\
    \hspace*{2.5em}dates computed from the timestamp of the memory in which the reference appears. \\
    \hspace*{1.5em}- Use the metadata timestamps (not the response time) when comparing event order. \\
    4. Speaker grounding: If the question asks about a user preference or assistant suggestion, \\
    \hspace*{1.5em}ensure the cited memory was authored by the correct speaker (User vs. Assistant). \\
    5. Abstention: If the retrieved memories do not contain enough information to answer the \\
    \hspace*{1.5em}question, output exactly: I don't know. Do not guess or fabricate. \\
    6. Format Constraints: \\
    \hspace*{1.5em}- Output the final answer concisely. \\
    \hspace*{1.5em}- Use Arabic digits for any counts, quantities, or years (e.g., "5" instead of "five"). \\

    \textbf{\# APPROACH (Think Step-by-Step):} \\
    1. Identify the question type and the speaker whose information is queried. \\
    2. Locate the relevant memory items and check their timestamps. \\
    3. Resolve any knowledge updates by preferring the most recent statement. \\
    4. Perform any necessary temporal arithmetic. \\
    5. If sufficient evidence is present, formulate the final answer; otherwise abstain. \\

    \textbf{\# INPUT DATA:} \\
    \textbf{Relevant Memories:} \\
    \{context\_memories\} \\

    Question: \{question\} \\

    \textbf{\# OUTPUT FORMAT:} \\
    Thought: [Briefly explain which memories you used, any conflict resolution, and any date \\
    calculations] \\
    Answer: [Your final concise answer, or "I don't know"]
\end{longprompt}

\vspace{0.8em}
\begin{longprompt}
    \textbf{\# LongMemEval-S LLM-as-a-Judge Prompt} \\
    Your task is to label an answer to a question as "CORRECT" or "WRONG". \\
    \textbf{You will be given:} \\
    (1) a question (with its question type tag, e.g., SSU/SSA/SSP/MS/KU/TR), \\
    (2) a gold (ground-truth) answer, \\
    (3) a generated answer. \\

    The question concerns long-term user information accumulated across multiple chat sessions. \\
    The gold answer is usually concise; the generated answer may be longer or include reasoning. \\

    Apply the following rules consistently with the question type: \\
    - Information extraction (SSU/SSA): Mark CORRECT if the generated answer clearly refers to \\
    \hspace*{1.0em}the same user fact or assistant statement as the gold answer; minor paraphrasing is \\
    \hspace*{1.0em}acceptable. \\
    - Single-session preference (SSP): Mark CORRECT if the generated response reflects the same \\
    \hspace*{1.0em}preference or recommendation rationale as the gold answer; surface form may differ. \\
    - Multi-session reasoning (MS): Mark CORRECT if the generated answer correctly aggregates or \\
    \hspace*{1.0em}compares the relevant facts; partial overlap that misses the comparison is WRONG. \\
    - Knowledge update (KU): Mark CORRECT only if the generated answer reflects the most recent \\
    \hspace*{1.0em}state of the user; answers based on outdated information are WRONG. \\
    - Temporal reasoning (TR): Accept equivalent dates or periods even if the format differs \\
    \hspace*{1.0em}(e.g., "May 7th" vs. "7 May"; "last September" vs. "September 2023"). \\
    - Abstention: If the gold answer is "I don't know" / "Unknown", only an explicit abstention \\
    \hspace*{1.0em}or refusal in the generated answer is CORRECT; any concrete guess is WRONG. \\

    Now judge the real case. \\

    Question: \{question\} \\
    Question type: \{question\_type\} \\
    Gold answer: \{standard\_answer\} \\
    Generated answer: \{generated\_answer\} \\

    \textbf{Return JSON only:} \\
    \{"label": "CORRECT" \textbar{} "WRONG"\}
\end{longprompt}

\vspace{0.8em}
\begin{longprompt}
    \textbf{\# PersonaMem-v2 QA Prompt (MCQ)} \\
    You are an intelligent assistant capable of synthesizing retrieved memory fragments to answer \\
    questions about a user's persona. \\

    \textbf{\# TASK:} \\
    Answer the current user question by selecting the best option from the provided list. You must \\
    rely STRICTLY on the provided "Retrieved Memories". \\

    \textbf{\# INPUT DATA:} \\
    1. *Retrieved Memories:* A collection of memory snippets retrieved from the user's past \\
    interactions. These may be fragmented or out of strict order. \\
    2. *Current Question:* The user's specific query or message. \\
    3. *Options:* The candidate answers. \\

    \textbf{\# CRITICAL INSTRUCTIONS:} \\
    1. *Synthesize and Verify:* \\
    - The retrieved memories might contain conflicting information (e.g., a memory from last year \\
    \hspace*{1.0em}saying "I eat meat" vs. a memory from yesterday saying "I am vegetarian now"). \\
    - You must strictly follow the *timeline* of the interactions. \\
    - If two memories conflict, the *MOST RECENT* information (latest timestamp or latest \\
    \hspace*{1.0em}interaction ID) trumps older information. \\
    - Prioritize the information that logically represents the *current* state of the user. \\
    2. *Distinguish Speaker:* \\
    - Ensure the preference belongs to the **User**, not the Assistant suggesting something. \\
    3. *Handle Missing Info:* \\
    - If the retrieved memories do not contain the answer, select the option representing \\
    \hspace*{1.0em}"Unknown" or the most neutral choice. Do not guess. \\
    4. *Output Format:* \\
    - *Thought:* Briefly analyze the retrieved memories. Resolve any conflicts if present. \\
    - *Answer:* Output ONLY the option label (e.g. "A", "B", "C", "D" etc.). \\

    \textbf{\# DATA:} \\
    Retrieved Memories : \{retrieved\_memories\} \\
    Current Question   : \{user\_question\_or\_message\} \\
    Option             : \{all\_options\} \\

    \textbf{\# RESPONSE:} \\
    Thought: [Analyze memories -\textgreater{} Resolve conflicts -\textgreater{} Match option] \\
    Answer: [Option Label]
\end{longprompt}

\vspace{0.8em}
\begin{longprompt}
    \textbf{\# PersonaMem-v2 LLM-as-a-Judge Prompt (MCQ)} \\
    Your task is to label an answer to a multiple-choice question as "CORRECT" or "WRONG". \\
    \textbf{You will be given:} \\
    (1) a question, \\
    (2) a gold (ground truth) answer, \\
    (3) a generated answer, \\
    (4) the list of options. \\

    The gold answer is usually an option label like "(c)". \\
    The generated answer may be a label or the full text of an option. \\
    Be generous: if the generated answer matches the gold label, or matches the gold option text, \\
    mark it CORRECT. \\
    If the gold answer is "Unknown", only "Unknown" (or a clear abstention) is CORRECT. \\

    Now judge the real case. \\

    Question: \{question\} \\
    Gold answer: \{standard\_answer\} \\
    Generated answer: \{generated\_answer\} \\
    \textbf{Options:} \\
    \{all\_options\} \\

    \textbf{Return JSON only:} \\
    \{"label": "CORRECT" \textbar{} "WRONG"\}
\end{longprompt}

\vspace{0.8em}
\begin{longprompt}
    \textbf{\# PersonaMem-v2 QA Prompt (Open-Ended)} \\
    You are an intelligent assistant capable of synthesizing retrieved memory fragments to answer \\
    questions about a user's persona. \\

    \textbf{\# TASK:} \\
    Respond to the current user question or message in a personalized way. You must rely STRICTLY \\
    on the provided "Retrieved Memories". \\

    \textbf{\# INPUT DATA:} \\
    1. *Retrieved Memories:* A collection of memory snippets retrieved from the user's past \\
    interactions. These may be fragmented or out of strict order. \\
    2. *Current Question:* The user's specific query or message. \\

    \textbf{\# CRITICAL INSTRUCTIONS:} \\
    1. *Synthesize and Verify:* \\
    - The retrieved memories might contain conflicting information (e.g., a memory from last year \\
    \hspace*{1.0em}saying "I eat meat" vs. a memory from yesterday saying "I am vegetarian now"). \\
    - You must strictly follow the *timeline* of the interactions. \\
    - If two memories conflict, the *MOST RECENT* information (latest timestamp or latest \\
    \hspace*{1.0em}interaction ID) trumps older information. \\
    - Prioritize the information that logically represents the *current* state of the user. \\
    2. *Distinguish Speaker:* \\
    - Ensure the preference belongs to the **User**, not the Assistant suggesting something. \\
    3. *Handle Missing Info:* \\
    - If the retrieved memories do not contain the relevant preference, give a helpful but neutral \\
    \hspace*{1.0em}response. Do not guess. \\
    4. *Output Format:* \\
    - *Thought:* Briefly analyze the retrieved memories. Resolve any conflicts if present. \\
    - *Answer:* A concise response (one to three sentences) tailored to the user's current state. \\

    \textbf{\# DATA:} \\
    Retrieved Memories : \{retrieved\_memories\} \\
    Current Question   : \{user\_question\_or\_message\} \\

    \textbf{\# RESPONSE:} \\
    Thought: [Analyze memories -\textgreater{} Resolve conflicts -\textgreater{} Tailor the response] \\
    Answer: [Your response]
\end{longprompt}

\vspace{0.8em}
\begin{longprompt}
    \textbf{\# PersonaMem-v2 LLM-as-a-Judge Prompt (Open-Ended)} \\
    Your task is to label a response to a user's message as "CORRECT" or "WRONG". \\
    \textbf{You will be given:} \\
    (1) the user's question or message, \\
    (2) a gold (ground truth) answer describing the response that fits the user's current preferences, \\
    (3) a generated response. \\

    Mark the generated response CORRECT if it is consistent with the preference expressed in the gold \\
    answer. Mark it WRONG if it contradicts that preference, relies on an outdated preference or on a \\
    preference that belongs to someone else, or ignores the preference entirely. \\
    If the gold answer is "Unknown", only "Unknown" (or a clear abstention) is CORRECT. \\

    Now judge the real case. \\

    Question: \{question\} \\
    Gold answer: \{standard\_answer\} \\
    Generated answer: \{generated\_answer\} \\

    \textbf{Return JSON only:} \\
    \{"label": "CORRECT" \textbar{} "WRONG"\}
\end{longprompt}

\vspace{0.8em}
\begin{longprompt}
    \textbf{\# DiMoBench QA Prompt} \\
    You are an intelligent memory assistant tasked with retrieving accurate information from \\
    conversation memories. \\

    \textbf{\# CONTEXT:} \\
    You have access to relevant memories extracted from a conversation history. \\
    - Input Format: List of ordered dialogue turns containing a Dialogue ID (e.g., D10:6), Speaker \\
    \hspace*{1.0em}Name, Timestamp or Turn Index, and Text content. \\
    - Task: Use these memories to answer the user's question accurately. \\

    \textbf{\# INSTRUCTIONS:} \\
    1. Analyze: Examine all provided memories carefully. Pay strict attention to the Timestamp or \\
    \hspace*{1.5em}Turn Order associated with each specific dialogue turn. \\
    2. Temporal Reasoning: \\
    \hspace*{1.5em}- If calendar timestamps are provided, you MUST convert all relative time references (e.g., \\
    \hspace*{2.5em}"last month", "three days later") into absolute dates (Day, Month, Year). \\
    \hspace*{1.5em}- Calculate these dates relative to the timestamp of the specific memory where the event is \\
    \hspace*{2.5em}mentioned. \\
    \hspace*{1.5em}- Example: If a memory dated "10 Oct 2023" says "I went there last month", the real date is \\
    \hspace*{2.5em}"September 2023". \\
    \hspace*{1.5em}- If only turn indices are provided (no calendar dates), use turn order to resolve \\
    \hspace*{2.5em}"before/after" and do NOT invent dates. \\
    3. Conflict Resolution: If memories contain contradictory information, prioritize the \\
    \hspace*{1.5em}information from the most recent timestamp or latest turn. \\
    4. Format Constraints: \\
    \hspace*{1.5em}- Output the final answer concisely. \\
    \hspace*{1.5em}- Use Arabic digits for any counts, quantities, or years (e.g., "5" instead of "five"). \\
    \hspace*{1.5em}- If the answer is not explicitly found or deducible, output exactly: Unknown. \\

    \textbf{\# APPROACH (Think Step-by-Step):} \\
    1. Locate the dialogue turns relevant to the question. \\
    2. Check the exact timestamp or turn order of those turns. \\
    3. Perform date calculations if relative time is mentioned. \\
    4. Check for any later updates that might contradict earlier facts. \\
    5. Formulate the final answer. \\

    \textbf{\# INPUT DATA:} \\
    \textbf{Relevant Memories:} \\
    \{context\_memories\} \\

    Question: \{question\} \\

    \textbf{\# OUTPUT FORMAT:} \\
    Thought: [Briefly explain your reasoning and date calculations here] \\
    Answer: [Your final concise answer here]
\end{longprompt}

\vspace{0.8em}
\begin{longprompt}
    \textbf{\# DiMoBench LLM-as-a-Judge Prompt} \\
    Your task is to label an answer to a question as "CORRECT" or "WRONG". \\
    \textbf{You will be given:} \\
    (1) a question, \\
    (2) a gold (ground truth) answer, \\
    (3) a generated answer. \\

    The question asks about a user's state, history, or change-point based on prior dialogue. \\
    Be generous: if the generated answer clearly refers to the same fact/state/turn as the gold \\
    answer, mark it CORRECT. \\
    For time/turn-related questions, accept equivalent expressions that point to the same time \\
    period or turn. \\
    If the gold answer is "Unknown", only "Unknown" (or a clear abstention) is CORRECT. \\

    Now judge the real case. \\

    Question: \{question\} \\
    Gold answer: \{standard\_answer\} \\
    Generated answer: \{generated\_answer\} \\

    \textbf{Return JSON only:} \\
    \{"label": "CORRECT" \textbar{} "WRONG"\}
\end{longprompt}

\subsection{DiMoBench construction prompts}
\label{app:dimobench_prompts}

The six-stage pipeline of \cref{app:dimobench} uses one generator prompt per stage (persona, life skeleton, dialogue, QA) and one verifier prompt. Each prompt opens with a stage marker and embeds the machine-readable plan it was built from under \texttt{PLAN\_JSON}, so that the model can reference cue and dialogue identifiers exactly. When the verifier returns \textsc{revise}, the revision request below is appended to the QA prompt and only the flagged question is rewritten.

\begin{longprompt}
    \textbf{\# Stage 1: Persona Prompt} \\
    {}[DIMOBENCH STAGE: persona] \\
    You are designing a synthetic user persona for DiMoBench, a benchmark about \\
    how a conversational assistant's long-term memory should track *gradual, \\
    implicit* changes in a user's attitudes and life circumstances. \\

    Seed specification (authoritative; keep every slot name and every state \\
    string exactly as given): \\
    PLAN\_JSON \\
    \{plan\_json\} \\

    \textbf{Design requirements:} \\
    - The persona must be psychologically plausible and consistent with the \\
    \hspace*{1.0em}Big-Five profile and life stage in the seed. \\
    - Give the persona a first name and a speaking style (register, typical \\
    \hspace*{1.0em}sentence length, verbal habits) that will stay recognisable over hundreds \\
    \hspace*{1.0em}of turns. \\
    - Produce a social circle of 3 named characters. At least two names must be \\
    \hspace*{1.0em}deliberately confusable (e.g. "Leo" and "Leon", "Mia" and "Mila"); state \\
    \hspace*{1.0em}explicitly which pair is confusable. Each character has a one-phrase \\
    \hspace*{1.0em}relation to the user. \\
    - Produce 4-6 routine activities that fit the *initial* states of the slots. \\
    - For every drift in the seed, write one concrete, event-driven trigger (a \\
    \hspace*{1.0em}specific thing that happens in the persona's life) that makes the planned \\
    \hspace*{1.0em}state change plausible without anyone ever naming the change. \\

    \textbf{Return strict JSON:} \\
    \{ \\
    \hspace*{1.0em}"name": "...", \\
    \hspace*{1.0em}"summary": "2-3 sentences, third person", \\
    \hspace*{1.0em}"speaking\_style": "...", \\
    \hspace*{1.0em}"social\_circle": [ \\
    \hspace*{2.0em}\{"name": "...", "relation": "...", "confusable\_with": "\textless{}name or null\textgreater{}"\} \\
    \hspace*{1.0em}], \\
    \hspace*{1.0em}"routine": ["...", "..."], \\
    \hspace*{1.0em}"drift\_triggers": \{"1": "...", "2": "...", "3": "..."\} \\
    \} \\
    Include only the drift indices present in the seed. Return JSON only.
\end{longprompt}

\vspace{0.8em}
\begin{longprompt}
    \textbf{\# Stage 2: Life-Skeleton Prompt} \\
    {}[DIMOBENCH STAGE: skeleton] \\
    You are writing the life skeleton for sessions \{first\_session\}-\{last\_session\} of a \\
    long DiMoBench dialogue between the user and an assistant (\{turns\_per\_session\} \\
    turns per session, \{n\_sessions\} sessions in total, \{n\_turns\} turns). \\

    \textbf{Persona:} \\
    \{persona\_json\} \\

    Gold state of every attribute slot across the whole dialogue (turn ranges): \\
    \{timeline\_json\} \\

    Drift triggers (never stated explicitly in the dialogue): \{triggers\_json\} \\

    Summary of the sessions written so far: \{previous\_summary\} \\

    Session plan for this chunk. For every session you get its phase, the cue \\
    ids it must express, and its confounder directive: \\
    PLAN\_JSON \\
    \{plan\_json\} \\

    \textbf{Phase semantics:} \\
    - "pre": stable baseline; cues show the *current* state of their slot through \\
    \hspace*{1.0em}routine, plans, reactions and small decisions. \\
    - "change-point": the drift trigger has just happened. Cues are implicit \\
    \hspace*{1.0em}events (a declined invitation, a changed routine, a hedge) that hint at the \\
    \hspace*{1.0em}new state without anyone naming a change. Keep them deniable in isolation. \\
    - "post-change-dense": the new state is now visible through denser, \\
    \hspace*{1.0em}consistent behavior. No meta commentary ("I've changed"). \\
    - "post-change-maintenance": low frequency; an occasional cue confirms the \\
    \hspace*{1.0em}new state as the new normal. Divergence from the post-drift baseline must \\
    \hspace*{1.0em}shrink monotonically over the maintenance stage. \\

    \textbf{Confounder directives:} \\
    - "keyword\_trap": the session contains irrelevant chit-chat (often about \\
    \hspace*{1.0em}someone else, a show, a news item) that shares surface keywords with the \\
    \hspace*{1.0em}focus slot but carries no information about the user's own state. \\
    - "hypothetical": the user narrates something hypothetical, counterfactual \\
    \hspace*{1.0em}or imagined ("if I ever...", "imagine I...") about the focus slot; it is \\
    \hspace*{1.0em}not a fact and must be written so that a careful reader can tell. \\
    - When "mention\_confusable\_entity" is true, one of the confusable characters \\
    \hspace*{1.0em}does or says something that a careless reader could attribute to the user. \\

    For each session write: a one-sentence summary, a time/place setting with an \\
    explicit temporal marker relative to the previous session, and for every cue \\
    id a concrete behavioral description plus a short surface hint. Cue \\
    descriptions must express "state\_to\_express" implicitly; never use the slot \\
    name or sentiment words such as like/love/hate/prefer about the slot itself. \\

    \textbf{Return strict JSON:} \\
    \{ \\
    \hspace*{1.0em}"sessions": [ \\
    \hspace*{2.0em}\{ \\
    \hspace*{3.0em}"index": \textless{}int\textgreater{}, \\
    \hspace*{3.0em}"summary": "...", \\
    \hspace*{3.0em}"setting": "...", \\
    \hspace*{3.0em}"cues": [\{"cue\_id": "...", "description": "...", "surface\_hint": "..."\}], \\
    \hspace*{3.0em}"confounder\_script": "\textless{}what the confounder is about, or empty string\textgreater{}" \\
    \hspace*{2.0em}\} \\
    \hspace*{1.0em}], \\
    \hspace*{1.0em}"chunk\_summary": "3-5 sentences summarizing these sessions for the next chunk" \\
    \} \\
    Return JSON only.
\end{longprompt}

\vspace{0.8em}
\begin{longprompt}
    \textbf{\# Stage 3: Dialogue Prompt} \\
    {}[DIMOBENCH STAGE: dialogue] \\
    You are writing Session \{session\_index\} of a DiMoBench dialogue between the \\
    user (\{user\_name\}) and an assistant. Produce exactly \{turns\_per\_session\} \\
    turns, alternating user / assistant and starting with the user. \\

    \textbf{Persona:} \\
    \{persona\_json\} \\

    \textbf{Rolling summary of the conversation so far:} \\
    \{rolling\_summary\} \\

    Last turns of the previous session (for continuity): \\
    \{previous\_tail\} \\

    \textbf{Session plan:} \\
    PLAN\_JSON \\
    \{plan\_json\} \\

    \textbf{Rules:} \\
    1. Every cue in the plan must be realized in a USER turn, implicitly: through \\
    \hspace*{1.5em}an activity choice, a reaction to an invitation, a plan, a small decision, \\
    \hspace*{1.5em}or hedging language. Never state the slot name, never use explicit \\
    \hspace*{1.5em}sentiment words about the slot (like/love/hate/prefer/enjoy), and never \\
    \hspace*{1.5em}add meta commentary ("I've changed", "I'm different now"). \\
    2. Use the character names from the persona exactly; when the plan asks for \\
    \hspace*{1.5em}a confusable-entity mention, make a confusable character the agent of an \\
    \hspace*{1.5em}action that superficially resembles the user's own behavior. \\
    3. If the plan has a confounder, realize it in one or two turns as described \\
    \hspace*{1.5em}("keyword\_trap" = irrelevant chit-chat sharing keywords; "hypothetical" = \\
    \hspace*{1.5em}clearly imagined / counterfactual narration, marked with words such as \\
    \hspace*{1.5em}"if", "imagine", "suppose", "would"). \\
    4. Put an explicit temporal marker ("last weekend", "since Tuesday", "this \\
    \hspace*{1.5em}month") in roughly one third of the user turns; the first user turn must \\
    \hspace*{1.5em}anchor the session in time relative to the previous one. \\
    5. Keep each turn under 45 words; the assistant is helpful and brief and \\
    \hspace*{1.5em}never analyses the user's personality. \\
    6. Stay in the persona's speaking style. Natural small talk and topic shifts \\
    \hspace*{1.5em}are welcome; the cues should feel incidental, not announced. \\

    \textbf{Return strict JSON:} \\
    \{ \\
    \hspace*{1.0em}"turns": [ \\
    \hspace*{2.0em}\{"speaker": "user", "text": "..."\}, \\
    \hspace*{2.0em}\{"speaker": "assistant", "text": "..."\} \\
    \hspace*{1.0em}], \\
    \hspace*{1.0em}"cue\_realizations": [\{"cue\_id": "...", "turn": \textless{}1-based index within this session\textgreater{}\}], \\
    \hspace*{1.0em}"confounder\_turns": [\textless{}1-based turn indices that carry the confounder, if any\textgreater{}], \\
    \hspace*{1.0em}"entity\_decoy\_turns": [\textless{}1-based turn indices where a confusable character acts, if any\textgreater{}], \\
    \hspace*{1.0em}"rolling\_summary": "\textless{}updated 4-6 sentence summary of the whole conversation so far\textgreater{}" \\
    \} \\
    Return JSON only.
\end{longprompt}

\vspace{0.8em}
\begin{longprompt}
    \textbf{\# Stage 4: QA-Generation Prompt} \\
    {}[DIMOBENCH STAGE: qa] \\
    You are writing evaluation questions for a DiMoBench dialogue. Each question \\
    is asked right after dialogue turn \{anchor\_turn\} (session \{anchor\_session\}); \\
    the system answering it has seen turns 1-\{anchor\_turn\} and nothing later. \\

    \textbf{Persona:} \\
    \{persona\_json\} \\

    Gold slot states valid at turn \{anchor\_turn\}: \\
    \{states\_json\} \\
    \{previous\_states\_block\} \\
    Dialogue excerpts available as evidence (id, speaker, text). Only these ids \\
    may be cited: \\
    \{excerpts\} \\

    Question requests. Each request fixes the reasoning category, the \\
    interference tag and the focus slot; write exactly one question per request: \\
    PLAN\_JSON \\
    \{plan\_json\} \\

    \textbf{Category definitions:} \\
    - "current-state inference": what is true of the user's focus slot *now* \\
    \hspace*{1.0em}(at turn \{anchor\_turn\}), inferred from behavior. \\
    - "history recall": a specific earlier fact, event or routine, cited at its \\
    \hspace*{1.0em}original turn(s). \\
    - "change-point detection": whether, when, or through which behaviors the \\
    \hspace*{1.0em}user's focus slot shifted; before a drift the correct answer is that no \\
    \hspace*{1.0em}shift has happened. \\
    - "hypothesis-vs-fact distinction": separate what the user actually did or \\
    \hspace*{1.0em}said from what was hypothetical, speculative or attributable to someone \\
    \hspace*{1.0em}else. \\

    Interference definitions (the question must be designed around the tag): \\
    - "entity\_confusable": the question tempts confusion between the user and a \\
    \hspace*{1.0em}confusable character, or between two confusable characters; cite the \\
    \hspace*{1.0em}decoy turn as well as the true evidence. \\
    - "keyword\_trap": a cited or nearby turn shares keywords with the slot but is \\
    \hspace*{1.0em}irrelevant chit-chat; the answer must ignore it. \\
    - "temporal\_decoy": the question quotes or alludes to an earlier statement \\
    \hspace*{1.0em}that may be outdated; the answer must reflect the state at turn \{anchor\_turn\} \\
    \hspace*{1.0em}(which may or may not have changed). \\
    - "hypothetical\_vs\_fact": the question tempts treating a hypothetical \\
    \hspace*{1.0em}narration as fact. \\
    - "none": no designed interference. \\

    \textbf{Writing rules:} \\
    - Questions ask about behavior, plans, actions and reactions; do not use the \\
    \hspace*{1.0em}slot name or sentiment words (like/love/hate/prefer/enjoy) about the slot. \\
    \hspace*{1.0em}Refer to the user by name. \\
    - Answers are concise (one sentence), unambiguous, and fully supported by \\
    \hspace*{1.0em}the cited evidence; for questions whose correct answer is that nothing \\
    \hspace*{1.0em}changed, say so plainly. \\
    - "evidence" lists every id needed to justify the answer (1-4 ids, all from \\
    \hspace*{1.0em}the excerpts above); "distractor\_ids" lists the decoy ids the question is \\
    \hspace*{1.0em}built around (may be empty). \\

    \textbf{Return strict JSON:} \\
    \{ \\
    \hspace*{1.0em}"questions": [ \\
    \hspace*{2.0em}\{ \\
    \hspace*{3.0em}"qid": "\textless{}copy from the request\textgreater{}", \\
    \hspace*{3.0em}"question": "...", \\
    \hspace*{3.0em}"answer": "...", \\
    \hspace*{3.0em}"evidence": ["D7:5", "D13:9"], \\
    \hspace*{3.0em}"distractor\_ids": ["D12:3"], \\
    \hspace*{3.0em}"rationale": "one sentence on why the evidence settles the answer" \\
    \hspace*{2.0em}\} \\
    \hspace*{1.0em}] \\
    \} \\
    Return JSON only.
\end{longprompt}

\vspace{0.8em}
\begin{longprompt}
    \textbf{\# Stage 4 (continued): Revision Request Appended on REVISE} \\
    REVISION REQUEST for \{qid\}: an independent verifier returned "\{decision\}" with \\
    this feedback: \\
    \{feedback\} \\
    Rewrite only this question (same category, tag, focus slot and anchor) so \\
    that the feedback is addressed. Return the same JSON schema with one entry.
\end{longprompt}

\vspace{0.8em}
\begin{longprompt}
    \textbf{\# Stage 5: Verifier Prompt} \\
    {}[DIMOBENCH STAGE: verify] \\
    You are an independent verifier for one DiMoBench question. You did not write \\
    the dialogue or the question; judge strictly. \\

    Persona summary: \{persona\_summary\} \\

    Gold slot states valid at turn \{anchor\_turn\}: \\
    \{states\_json\} \\
    \{previous\_states\_block\} \\
    Requirements the question was supposed to satisfy: \\
    PLAN\_JSON \\
    \{plan\_json\} \\

    Dialogue excerpts: the cited evidence, the cited distractors, and one turn of \\
    context on each side (id, speaker, text): \\
    \{excerpts\} \\

    \textbf{The question:} \\
    \{qa\_json\} \\

    \textbf{Score six criteria with 0 or 1:} \\
    1. answer\_grounding     - the answer is fully supported by the cited evidence. \\
    2. evidence\_complete    - every turn needed for the answer is cited; nothing \\
    \hspace*{13.0em}essential is missing. \\
    3. temporal\_consistency - the answer reflects the gold state at turn \{anchor\_turn\}, \\
    \hspace*{13.0em}not an earlier or later state. \\
    4. entity\_accuracy      - every person referenced is the right one; decoys are \\
    \hspace*{13.0em}not misattributed to the user. \\
    5. interference\_valid   - the question genuinely instantiates its interference \\
    \hspace*{13.0em}tag ("none" is valid when no decoy is designed in). \\
    6. drift\_coherent       - the behavior the question relies on is consistent \\
    \hspace*{13.0em}with a gradual, plausible change (or stable state) of \\
    \hspace*{13.0em}the persona; nothing in the cited turns announces a \\
    \hspace*{13.0em}change explicitly. \\

    Decision rule: ACCEPT if all six are 1; REJECT if answer\_grounding or \\
    temporal\_consistency is 0 and the question cannot be fixed by rewording; \\
    otherwise REVISE with concrete, actionable feedback. \\

    \textbf{Return strict JSON:} \\
    \{ \\
    \hspace*{1.0em}"scores": \{"answer\_grounding": 0\textbar{}1, "evidence\_complete": 0\textbar{}1, "temporal\_consistency": 0\textbar{}1, \\
    \hspace*{6.5em}"entity\_accuracy": 0\textbar{}1, "interference\_valid": 0\textbar{}1, "drift\_coherent": 0\textbar{}1\}, \\
    \hspace*{1.0em}"decision": "ACCEPT" \textbar{} "REVISE" \textbar{} "REJECT", \\
    \hspace*{1.0em}"feedback": "..." \\
    \} \\
    Return JSON only.
\end{longprompt}

\end{document}